\documentclass{article}

\usepackage{PRIMEarxiv}

\usepackage[utf8]{inputenc} 
\usepackage[T1]{fontenc}    
\usepackage{hyperref}       
\usepackage{url}            
\usepackage{booktabs}       
\usepackage{amsfonts}       
\usepackage{nicefrac}       
\usepackage{microtype}      
\usepackage{lipsum}
\usepackage{fancyhdr}       
\usepackage{graphicx}       
\graphicspath{{media/}}     

\usepackage{amsmath}
\usepackage{algorithm}
\usepackage{array}
\usepackage[caption=false,font=normalsize,labelfont=sf,textfont=sf]{subfig}
\usepackage{textcomp}
\usepackage{stfloats}
\usepackage{verbatim}
\usepackage[square,numbers]{natbib}
\usepackage{multirow}
\usepackage{svg}
\usepackage{makecell}
\usepackage{placeins}
\usepackage{algorithm}
\usepackage{algpseudocode}
\usepackage{setspace}

\newcolumntype{K}[1]{>{\centering\arraybackslash}p{#1}}

\title{Investigating Bayesian optimization for expensive-to-evaluate black box functions: Application in fluid dynamics}

\author{
  Mike Diessner \\
  School of Computing \\
  Newcastle University \\
  Newcastle upon Tyne, UK\\
  \texttt{m.diessner2@newcastle.ac.uk} \\
  \And
  Joseph O'Connor \\
  Department of Aeronautics \\
  Imperial College London \\
  London, UK\\
  \And
  Andrew Wynn \\
  Department of Aeronautics \\
  Imperial College London \\
  London, UK\\
  \And
  Sylvain Laizet \\
  Department of Aeronautics \\
  Imperial College London \\
  London, UK\\
  \And
  Yu Guan \\
  School of Computing \\
  Newcastle University \\
  Newcastle upon Tyne, UK\\
  \And
  Kevin Wilson \\
  School of Mathematics, Statistics and Physics \\
  Newcastle University \\
  Newcastle upon Tyne, UK\\
  \texttt{kevin.wilson@newcastle.ac.uk} \\
  \And
  Richard D. Whalley \\
  School of Engineering \\
  Newcastle University \\
  Newcastle upon Tyne, UK \\
  \texttt{richard.whalley@newcastle.ac.uk} \\
}

\begin{document}
\maketitle

\begin{abstract}
Bayesian optimization provides an effective method to optimize expensive-to-evaluate black box functions. It has been widely applied to problems in many fields, including notably in computer science, e.g. in machine learning to optimize hyperparameters of neural networks, and in engineering, e.g. in fluid dynamics to optimize control strategies that maximize drag reduction. This paper empirically studies and compares the performance and the robustness of common Bayesian optimization algorithms on a range of synthetic test functions to provide general guidance on the design of Bayesian optimization algorithms for specific problems. It investigates the choice of acquisition function, the effect of different numbers of training samples, the exact and Monte Carlo based calculation of acquisition functions, and both single-point and multi-point optimization. The test functions considered cover a wide selection of challenges and therefore serve as an ideal test bed to understand the performance of Bayesian optimization to specific challenges, and in general. To illustrate how these findings can be used to inform a Bayesian optimization setup tailored to a specific problem, two simulations in the area of computational fluid dynamics are optimized, giving evidence that suitable solutions can be found in a small number of evaluations of the objective function for complex, real problems. The results of our investigation can similarly be applied to other areas, such as machine learning and physical experiments, where objective functions are expensive to evaluate and their mathematical expressions are unknown.
\end{abstract}

\keywords{Bayesian optimization \and Gaussian processes \and Computer simulations \and Black box functions \and Fluid dynamics \and Turbulent drag reduction}

\section{Introduction}
\label{sec1}

 Expensive black box functions are functions characterized by high evaluation costs and unknown or non-existent mathematical expressions. These functions cannot be optimized with conventional methods as they often require gradient information and large numbers of function evaluations. Bayesian optimization, however, is a specialized optimization strategy developed for this specific scenario. It can be used to locate global optima of objective functions that are expensive to evaluate and whose mathematical expressions are unknown \citep{Shahriari2016, gramacy2020surrogates}. 
 
 Recently, it has been applied successfully in the data-intensive field of computational fluid dynamics (CFD). Most notably, \citet{Talnikar2014} developed a Bayesian optimization framework for the parallel optimization of large eddy simulations. They used this framework (i) on a one-dimensional problem to determine the wave speed of a traveling wave that maximizes the skin-friction drag reduction in a turbulent channel flow and (ii) on a four-dimensional problem to find an efficient design for the trailing edge of a turbine blade that minimizes the turbulent heat transfer and pressure loss. For the former, Bayesian optimization was able to locate a wave speed to generate a skin-friction drag reduction of 60\%, while for the latter within 35 objective function evaluations a design was found that reduced the heat transfer by 17\% and the pressure loss by 21\%. \citet{Mahfoze2019} utilized Bayesian optimization on a four-dimensional problem to locate optimal low-amplitude wall-normal blowing strategies to reduce the skin-friction drag of a turbulent boundary-layer with a net power saving of up to 5\%, within 20 optimization evaluations. \citet{Morita2022} considered three CFD problems: the first two problems concerned the shape optimization of a cavity and a channel flow. The third problem optimized the hyperparameters of a spoiler-ice model. Lastly, \citet{Nabae2021} maximized the skin-friction drag reduction in a turbulent channel flow by optimizing the velocity amplitude and the phase speed of a traveling wave-like wall deformation. They achieved a maximum drag reduction of 60.5\%.
 
 Apart from these experiments, Bayesian optimization has also been used to tune hyperparameters of statistical models. For example, \citet{wu2019hyperparameter} optimized the hyperparameters of a selection of models, such as random forests, convolutional neural networks, recurrent neural networks, and multi-grained cascade forests. In these low-dimensional cases (two or three parameters) the model performance increased and the time to find the solutions was reduced compared to manual search for each model. 
 
While these examples show that Bayesian optimization performs well on a wide selection of problems, such as computer simulations and hyperparameter tuning of machine learning models, there is no extensive study on the many different types of Bayesian optimization algorithms in the literature. In particular, there is a research gap regarding the performance and the robustness of Bayesian optimization when applied to distinct challenges. This paper aims to address this gap by considering a wide range of algorithms and a wide range of problems with increasing levels of complexity that are, consequently, increasingly difficult to solve. The comparison and the thorough analysis of these algorithms can inform the design of Bayesian optimization algorithms and allows them to be tailored to the unique problem at hand. To show how the findings can guide the setup of Bayesian optimization, two novel simulations in the area of computational fluid dynamics are optimized, demonstrating that Bayesian optimization can find suitable solutions to complex problems in a small number of function evaluations. The results of these experiments show that Bayesian optimization is indeed capable of finding promising solutions for expensive-to-evaluate black box functions. In these specific cases, solutions were found that (a) maximize the drag reduction over a flat plate and (b) achieve drag reduction and net energy savings simultaneously. For these experiments, the Newcastle University Bayesian Optimization (NUBO) framework, currently being developed to optimize fluid flow problems, is utilized.

The paper is structured as follows. In Section~\ref{sec2}, the Bayesian optimization algorithm, in particular the underlying Gaussian process (GP) and the different classes of acquisition functions, are reviewed. Section~\ref{sec3.1} discusses the synthetic benchmark functions and their advantages over other types of common benchmarking methods. In Section~\ref{sec3.2}, four different sets of simulations are presented. These experiments study~\ref{sec3.2.1} multiple analytical single-point acquisition functions,~\ref{sec3.2.2} varying numbers of initial training points,~\ref{sec3.2.3} acquisition functions utilizing Monte Carlo sampling and~\ref{sec3.2.4} various multi-point acquisition functions. Finally, Section~\ref{sec3.3} discusses the main findings and relates them back to the problems presented in the introduction and in general. These findings are then used to inform the algorithm used to optimize the setup of two computational fluid dynamics experiments in Section~\ref{sec4}. Lastly, a brief conclusion is drawn in Section~\ref{sec5}.

\section{Bayesian optimisation}
\label{sec2}

Bayesian optimization aims to find the global optimum of an expensive-to-evaluate objective function, whose mathematical expression is unknown or does not exist, in a minimum number of evaluations. It first fits a surrogate model, most commonly a GP, to some initial training data. The GP reflects the current belief about the objective function and is used to compute heuristics, called acquisition functions. When optimized, these functions suggest the candidate point that should be evaluated next from the objective function. After the response for the new candidate point is computed from the objective function, it is added to the training data and the process is repeated until a satisfying solution is found or a predefined evaluation budget is exhausted \citep{Shahriari2016}. The following sections provide an overview of the two main components of Bayesian optimization, the GP and the acquisition functions.

\subsection{Gaussian process}
\label{sec2.1}

Consider an objective function $\textbf{f} : \mathcal{X} \rightarrow \mathbb{R}$, where $\mathcal{X} \subset \mathbb{R}^d$ is a d-dimensional design space, that allows the evaluation of one d-dimensional point $\textbf{x}_i$ or multiple d-dimensional points $\textbf{X}$ within the design space $\mathcal{X}$ yielding one or multiple observations $\textbf{f}(\textbf{x}_i) = y_i$ and $\textbf{f}(\textbf{X}) = \textbf{y}$ respectively. A popular choice for the surrogate model to represent such an objective function is a GP. A GP is a stochastic process for which any finite set of points can be represented by a multivariate Normal distribution. This so called prior is defined by a mean function $\mu_0 : \mathcal{X} \rightarrow \mathbb{R}$ and a positive definite covariance function or kernel $k : \mathcal{X} \times \mathcal{X} \rightarrow \mathbb{R}$. These induce a mean vector $\textbf{m}_i := \mu_0(\textbf{x}_i)$ and a covariance matrix $\textbf{K}_{i, j} := k (\textbf{x}_i, \textbf{x}_j)$ that define the prior distribution of the GP as

\begin{gather}
    \textbf{f} \hspace{1mm} | \hspace{1mm} \textbf{X} \sim \mathcal{N}(\textbf{m}, \textbf{K})
\end{gather}where $\textbf{f} \hspace{1mm} | \hspace{1mm} \textbf{X}$ are the unknown responses from the objective function given a collection of points $\textbf{x}_{1:n}$. Given a set of observed design points $\mathcal{D}_n = \{(\textbf{x}_i, y_i)\}^n_{i=1}$ and a new point also called a candidate point $\textbf{x}$, the posterior predictive distribution of $\textbf{x}$ can be computed:

\begin{gather}
    Y \hspace{1mm} | \hspace{1mm} \mathcal{D}_n \sim \mathcal{N}(\mu_n(\textbf{x}), \sigma_n^2(\textbf{x}))
\end{gather}
\noindent with the posterior mean and variance
\begin{gather}
    \mu_n(\textbf{x}) = \mu_0(\textbf{x}) + \textbf{k}(\textbf{x})^T (\textbf{K} + \sigma^2\textbf{I})^{-1} (\textbf{y} - \textbf{m})\\
    \sigma^2_n(\textbf{x}) = k(\textbf{x}, \textbf{x}) - \textbf{k}(\textbf{x})^T(\textbf{K} + \sigma^2\textbf{I})^{-1} \textbf{k}(\textbf{x})
\end{gather}
where $\textbf{K} = k(\textbf{x}_{1:n}, \textbf{x}_{1:n})$ is a square matrix and $\textbf{k}(\textbf{x}) = k(\textbf{x}, \textbf{x}_{1:n})$ is a vector containing the covariances between the new point $\textbf{x}$ and all design points $\textbf{x}_{1:n}$ \citep{Shahriari2016, Snoek2012, gramacy2020surrogates}.

For the experiments presented in this paper, the zero mean function is implemented as it fulfils the theoretical properties required to compute the variable hyperparameter $\beta$ for the Gaussian Process Upper Confidence Bound (GP-UCB) algorithm as derived in \citet{Srinivas2009} (see Section II. B.). While there are many covariance kernels used in the literature, this paper implements the Mat\'{e}rn kernel with $\nu = 5/2$ with an individual length scale parameter for each input dimension as proposed by \citet{Snoek2012} for practical optimization problems, as other kernels such as the squared-exponential kernel also known as the radial basis function kernel can be too smooth for applied problems such as physical experiments. This kernel requires an additional parameter, the output scale $\tau^2$, to allow the kernel to be scaled above a certain threshold by multiplying the output scale with the posterior variance $\tau^2 \sigma_n^2(\textbf{x})$ \citep{gramacy2020surrogates}. A nugget term $v$ is added to the diagonal of the covariance matrix $\textbf{K}_{i, j} := k (\textbf{x}_i, \textbf{x}_j) + vI$ to take possible errors, measurement and other, into consideration where $I$ is the identity matrix. Nuggets have been shown to improve the computational stability and the performance of the algorithm such as the coverage and robustness of the results when using sparse data \citep{Andrianakis2012}. Additionally, \citet{Gramacy2010} give a detailed reasoning why nuggets should be added to deterministic problems that exceeds solely technical advantages. For example, while there might not be a measurement error in a deterministic simulation, the simulation itself is biased as it only approximates reality. In such cases, it makes sense to include a model to take possibles biases into account. The hyperparameters of the GP, in this case the nugget, the output scale and the length scales of the Mat\'{e}rn kernel, can be estimated directly from the data, for example using maximum likelihood estimation (MLE) or maximum a posteriori (MAP) estimation \citep{Shahriari2016, Rasmussen2006}. Algorithm 1 in Section 2 of the supplementary material in the appendix shows how the hyperparameters can be estimated via MLE by maximizing the log marginal likelihood of the GP.

\subsection{Acquisition functions}
\label{sec2.2}

Acquisition functions utilize the posterior distribution of the GP, that is the surrogate model representing the objective function, to propose the next point to sample from the objective function. \citet{Shahriari2016} group acquisition functions into four categories: improvement-based policies, optimistic policies, information-based policies and portfolios. This section presents some popular representatives of each group, which we shall focus on in this paper. Pseudocode for all acquisition functions can be found in Section 2 of the supplementary material in the appendix .

Improvement-based policies such as Probability of Improvement (PI) and Expected Improvement (EI) propose points that are better than a specified target, for example the best point evaluated so far, with a high probability. Given the best point so far $\textbf{x}^*$ and its response value $y^*=f(\textbf{x}^*)$, PI (\ref{eq:PI}) and EI (\ref{eq:EI}) can be computed as:

\begin{gather}
    \alpha_{PI}(\textbf{x}; \mathcal{D}_n)  = \Phi\left(\frac{z}{\sigma_n(\textbf{x})}\right) \label{eq:PI}\\
    \alpha_{EI}(\textbf{x}; \mathcal{D}_n) = z \Phi\left(\frac{z}{\sigma_n(\textbf{x})}\right) + \sigma_n(\textbf{x}) \phi \left(\frac{z}{\sigma_n(\textbf{x})}\right) \label{eq:EI}
\end{gather}
respectively, where $z = \mu_n(\textbf{x}) - y^*$ and $\Phi$ and $\phi$ are the CDF and the PDF of the standard Normal distribution.

The Upper Confidence Bound (UCB) acquisition function is an optimistic policy. It is optimistic with regards to the uncertainty (variance) of the GP, that is UCB assumes the uncertainty to be true. In contrast to the improvement-based methods, UCB (\ref{eq:UCB}) has a tuneable hyperparameter $\beta_n$ that balances the exploration-exploitation trade-off. This trade-off describes the decision the Bayesian optimization algorithm has to make at each iteration. The algorithm could either explore and select new points in an area with high uncertainty or it could exploit and select points in areas where the GP makes a high prediction \citep{Shahriari2016}. Thus, chosing how to set the trade-off parameter $\beta_n$ is an important decision. Common strategies for setting $\beta_n$ include fixing it to a particular value or varying it, as in the GP-UCB algorithm \citep{Srinivas2009}.

\begin{gather}
    \alpha_{UCB}(\textbf{x}; \mathcal{D}_n) = \mu_n(\textbf{x}) + \sqrt{\beta_n} \sigma_n(\textbf{x}) \label{eq:UCB}
\end{gather}
Portfolios consider multiple acquisition functions at each step and choose the best-performing one. The Hedge algorithm outlined by \citet{Brochu2010} is used in this research article. It requires the computation and optimization of all acquisition functions in the portfolio at each Bayesian optimization iteration and the assessment of their proposed new point compared to the posterior of the updated GP in retrospect. This means that portfolios are computationally costly. However, because the performance of different acquisition functions varies for each iteration, i.e. for some iterations, one acquisition function would be optimal, while another function would be preferred in a different iteration, the Hedge algorithm should in theory select the best acquisition function at each iteration, yielding a better solution than when just considering one individual acquisition function.

In addition to these analytical acquisition functions, Monte Carlo sampling can be implemented. While this is an approximation of the analytical functions, Monte Carlo sampling does not require the oftentimes strenuous explicit computations involved with the analytical functions, especially when considering multi-point approaches. Information-based policies, in particular those using entropy, aim to find a candidate point \textbf{x} that reduces the entropy of the posterior distribution $p(\textbf{x}|\mathcal{D}_n)$. While Entropy Search (ES) and Predictive Entropy Search (PES) are computationally expensive, \citet{Wang2017} and \citet{Takeno2020} introduce Max-value Entropy Search (MES), a method that uses information about simple to compute maximal response values instead of costly to compute entropies.

Furthermore, the analytical acquisition functions can be used with Monte Carlo sampling by reparameterizing equations (\ref{eq:PI}) - (\ref{eq:UCB}) \citep{Wilson2017}. Then, they can be computed by sampling from the posterior distribution of the GP instead of computing the acquisition functions directly:

\begin{gather}
    \alpha_{MC \hspace{1mm} PI}(\textbf{x}; \mathcal{D}_n) = \sigma \left( \frac{\max(\mu_n(\textbf{x}) + \textbf{Lz}) - y^*}{\tau} \right) \label{eq:MCPI} \\
    \alpha_{MC \hspace{1mm} EI}(\textbf{x}; \mathcal{D}_n) = \max\left(0, \hspace{1mm} \max\left(\mu_n(\textbf{x}) + \textbf{Lz}\right) - y^*\right) \label{eq:MCEI} \\
    \alpha_{MC \hspace{1mm} UCB}(\textbf{x}; \mathcal{D}_n) = \max\left(\mu_n(\textbf{x}) + \sqrt{\beta \pi / 2} |\textbf{Lz}|\right) \label{eq:MCUBC}
\end{gather}
where $\textbf{z}$ is a vector containing samples from a Standard Normal distribution $\textbf{z} \sim \mathcal{N}(0, \mathcal{\textbf{I}})$ and $\textbf{L}$ is the lower triangular matrix of the Cholesky decomposition of the covariance matrix $\textbf{K} = \textbf{LL}^T$. The softmax function $\sigma$ with the so-called temperature parameter $\tau$ enables the reparameterization of PI by approximating necessary gradients that are otherwise almost entirely equal to 0. As $\tau \rightarrow 0$, this approximation turns precise as described in Section 2 of \citet{Wilson2017}.

It is straightforward to extend the reparameterizations of the analytical acquisition functions to multi-point methods that propose more than one point at each iteration of the Bayesian optimization algorithm. In this case, samples are taken from a joint distribution (including training points and new candidate points) that is optimized with respect to the new candidate points. This optimization can be performed jointly, i.e. optimizing the full batch of points simultaneously, or in a sequential manner where a new joint distribution is computed for each point in the batch including the previously computed batch points, but only optimizing it with respect to the newest candidate point \citep{botorch}.
While analytical functions cannot be extended naturally to the multi-point case, there exist some frameworks that allow the computation of batches. The Constant Liar framework uses EI in combination with a proxy target, usually the minimal or maximal response value encountered so far. After one iteration of optimizing EI, the lie is taken as the true response for the newly computed candidate point and the process is repeated until all points in the batch are computed. At this point, the true responses of the candidate points are evaluated via the objective function and the next batch is computed \citep{Ginsbourger}. Similarly, the Gaussian Process Batched Upper Confidence Bound (GP-BUCB) approach leverages the fact that the variance of the multivariate Normal distribution can be updated based solely on the inputs of a new candidate point. However, to update the posterior mean, the response value is required. GP-BUCB updates the posterior variance after each new candidate point, while the posterior mean stays the same for the full batch. After all points of a batch are computed, the response values are gathered from the objective function and the next batch is computed \citep{Desautels2014}.

\section{Synthetic test functions}
\label{sec3}

\subsection{Functions}
\label{sec3.1}

While there are various methods of benchmarking the performance of optimization algorithms, such as sampling objective functions from Gaussian processes or optimizing parameters of models \citep{Snoek2012, Brochu2010, Wang2017}, this paper focuses on using synthetic test functions. Synthetic test functions have the main advantages that their global optima and their underlying shape are known. Thus, when testing the algorithms outlined in section~\ref{sec2.2} on these functions, the distance from their results to the true optimum can be evaluated. For other benchmarking methods, this might not be possible as, for example, in the case of optimizing the hyperparameters of a model, the true optimum, i.e. the solution optimizing the performance of the model, is rarely known. Hence, results can only be discussed relative to other methods, ignorant of the knowledge of how close any method actually is to the true optimum.  Furthermore, knowing the shape of the test function is advantageous as it provides information on how challenging the test function is. For example, knowing that a test function is smooth and has a single optimum indicates that it is less complex and thus less challenging than a test function that possesses multiple local optima, in each of which the algorithm potentially could get stuck.

\begin{figure}
    \begin{center}
        \includegraphics[width=10cm]{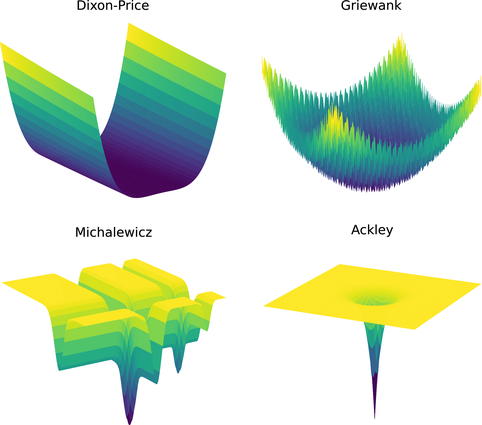}
    \end{center}
    \caption{Different challenges and levels of complexity of the simulations represented by the shapes of four test functions.}\label{fig:Figure 1}
\end{figure}

\begin{table}[h]
    \centering
    \caption{Overview of the seven synthetic test functions.}
    \begin{tabular}{ c | c c c }
        Test function & Number of input dimensions D & Shape & Number of optima \\[0.2cm]
        \hline \hline \\
        Sphere & 10 & Bowl-shaped & 1\\[0.15cm]
        Dixon-Price & 10 & Valley-shaped & 1\\[0.15cm]
        Griewank & 8 & Oscillatory & 1\\[0.15cm]
        Hartmann & 6 & Multi-modal & 6\\[0.15cm]
        Noisy Hartmann & 6 & Noisy & 6\\[0.15cm]
        Michalewicz & 5 & Steep edges & 120\\[0.15cm]
        Ackley & 6 & Mostly flat & 1\\
    \end{tabular}
    \label{table:Table 1}
\end{table}

This paper assesses the Bayesian optimization algorithms on eight different synthetic test functions spanning a wide range of challenges. Table \ref{table:Table 1} presents the test functions and gives details on the number of input dimensions, the shape and the number of optima to give an idea about the complexity of the individual functions, while Figure \ref{fig:Figure 1} shows some of the functions that can be presented in three-dimensional space and indicates the increasing complexity of the functions. Detailed mathematical definitions for all test functions can be found in Section 3 of the supplementary material in the appendix. The number of input dimensions for the test functions was chosen to be equal to or slightly greater than the number in the simulations recently published in the fluid dynamics community, which is typically three to eight \citep{Mahfoze2019, Morita2022, Talnikar2014, Nabae2021}. The 10D Sphere and 10D Dixon-Price functions are less challenging problems with a high degree of smoothness and only one optimum. They are therefore considered for a higher number of input dimensions. The 8D Griewank function adds a layer of complexity by introducing oscillatory properties. The 6D Hartmann function increases the level of difficulty further through its multi-modality. It has six optima with only one global optimum. This function is also considered in two noisy variants to simulate typical measurement uncertainty encountered during experiments in fluid dynamics. In particular, the standard deviation of the added Gaussian noise is calculated so that it represents the measurement errors of state-of-the-art Micro-Electro-Mechanical-Systems (MEMS) sensors which directly measure time-resolved skin-friction drag in turbulent air flows (e.g. the flow over an aircraft); that is, 1.4 to 2.4\% in an experimental setting \citep{Nima2022}. Factoring in the range of the 6D Hartmann function, the corresponding standard deviations for the Gaussian noise, taken as a 99\% confidence interval, are therefore 0.0155 and 0.0266 respectively. The most complex test functions are the 5D Michalewicz and the 6D Ackley functions (this paper considers a modified Ackley function with $a=20.0$, $b=0.5$ and $c=0.0$). While the former has 120 optima and steep ridges, the latter is mostly flat with a single global minimum in the centre of the space. The gradient of the function close to the optimum and the large flat areas represent a great level of difficulty, as illustrated in the bottom row of Figure \ref{fig:Figure 1}.

\subsection{Results}
\label{sec3.2}

This section focuses on the results for four of the eight test functions that were considered in this paper. The results for the other four test functions, as well as more extensive tables, can be found in the supplementary material in the appendix. The 8D Griewank function, the 6D Hartmann function in variations without noise and with the high noise level and the 6D Ackley function are examined here, as they represent increasing levels of complexity and come with unique challenges as illustrated in Section~\ref{sec3.1}. Mirroring real world applications, a budget for the number of evaluations was imposed on the test functions: 200 evaluations for the Griewank and Hartmann functions and 500 for the more complex Ackley function. If not otherwise stated, the number of  initial training points is equivalent to five points per input dimension of the given test function, i.e. 40 training points for the Griewank function and 30 training points for the Hartmann and Ackley functions. For each of the methods in the following sections, 50 different optimization runs were computed to investigate how robust and sensitive the methods are to varying initial training points. Each run was initialized with a different set of initial training points sampled from a Maximin Latin Hypercube Design \citep{Husslage2011}. However, the points were identical for all methods for a specific test function. All experiments were run on container instances on the cloud with the same specifications (two CPU cores and 8GB of memory) to make runs comparable.

Overall, four different sets of experiments are considered. First, analytical single-point acquisition functions are compared. Second, the effects of a varying number of initial training points are investigated. Third, analytical methods are compared with Monte Carlo methods and, lastly, multi-point or batched methods are compared to the single-point results. As a baseline of performance, space filling designs exhausting the full evaluation budgets were sampled from a Maximin Latin Hypercube.

\subsubsection{Analytical single-point acquisition functions}
\label{sec3.2.1}

Section~\ref{sec2.2} describes four different groups of acquisition functions: improvement-based, optimistic, portfolios, and entropy-based. In this section, representative functions from the first three groups are tested on the synthetic test functions outlined above. The entropy-based approach is considered in Section~\ref{sec3.2.3}. The focus lies on analytical single-point acquisition functions as they are widely used and thus present a natural starting point. Overall, seven different methods are considered: PI, EI, UCB with a variable and fixed $\beta$ (5.0 and 1.0) and a Hedge portfolio that combines the PI, EI and variable UCB acquisition functions. For a detailed discussion of these functions see Section~\ref{sec2.2}.

\begin{figure}
    \begin{minipage}[t]{0.02\textwidth}
        \vspace{-50mm}
        (A)
    \end{minipage}
    \begin{minipage}[b]{0.48\textwidth}
        \centering
        \includegraphics[width=0.9\linewidth]{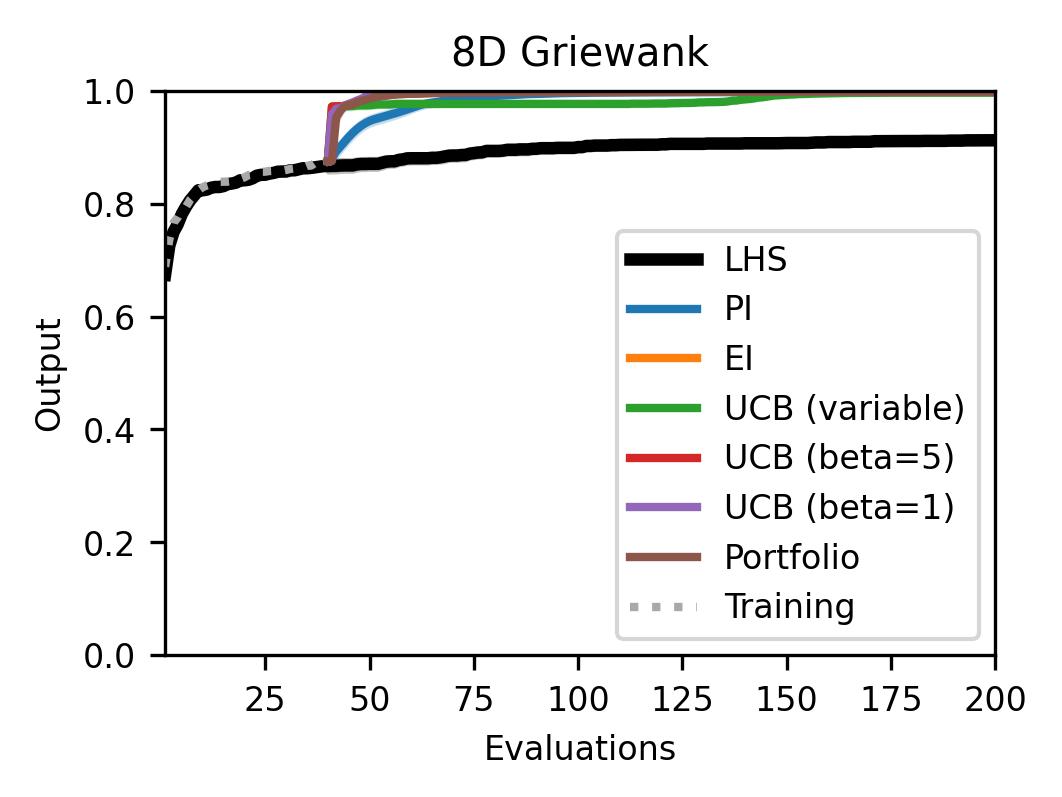}
        \label{fig:Subfigure 2a}
    \end{minipage}
    \begin{minipage}[t]{0.02\textwidth}
        \vspace{-50mm}
        (B)
    \end{minipage}
    \begin{minipage}[b]{0.48\textwidth}
        \centering
        \includegraphics[width=0.9\linewidth]{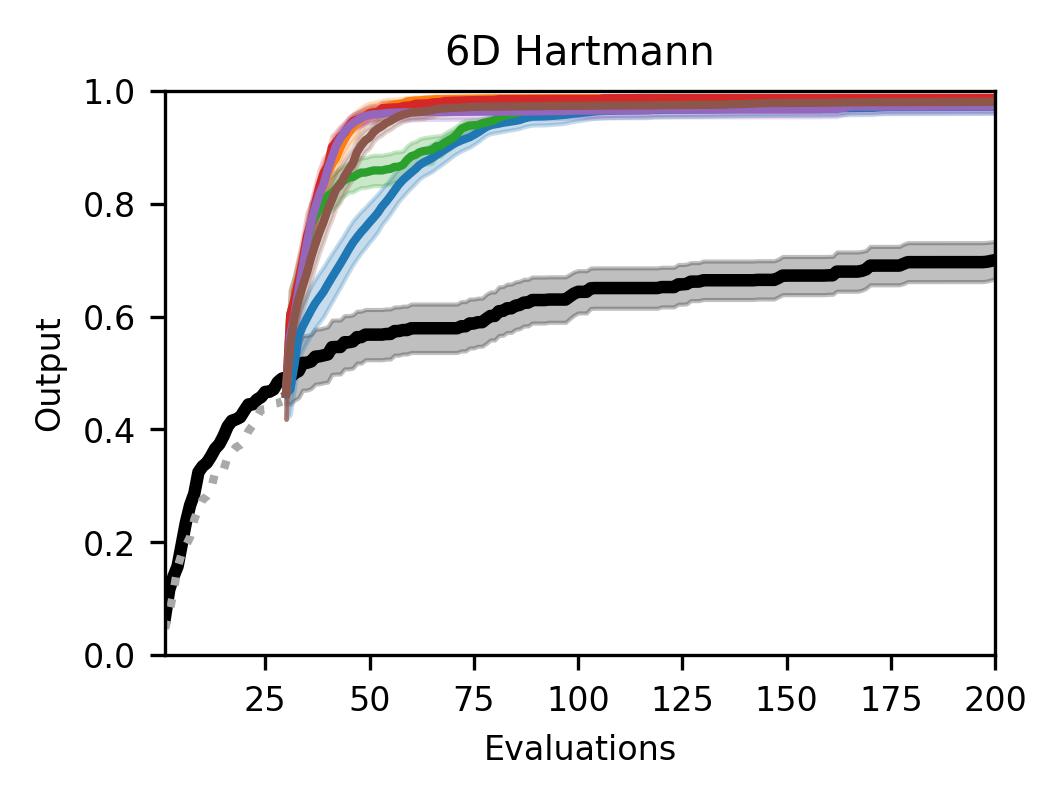}
        \label{fig:Subfigure 2b}
    \end{minipage}  
   
    \begin{minipage}[t]{0.02\textwidth}
        \vspace{-50mm}
        (C)
    \end{minipage}
    \begin{minipage}[b]{0.48\textwidth}
        \centering
        \includegraphics[width=0.9\linewidth]{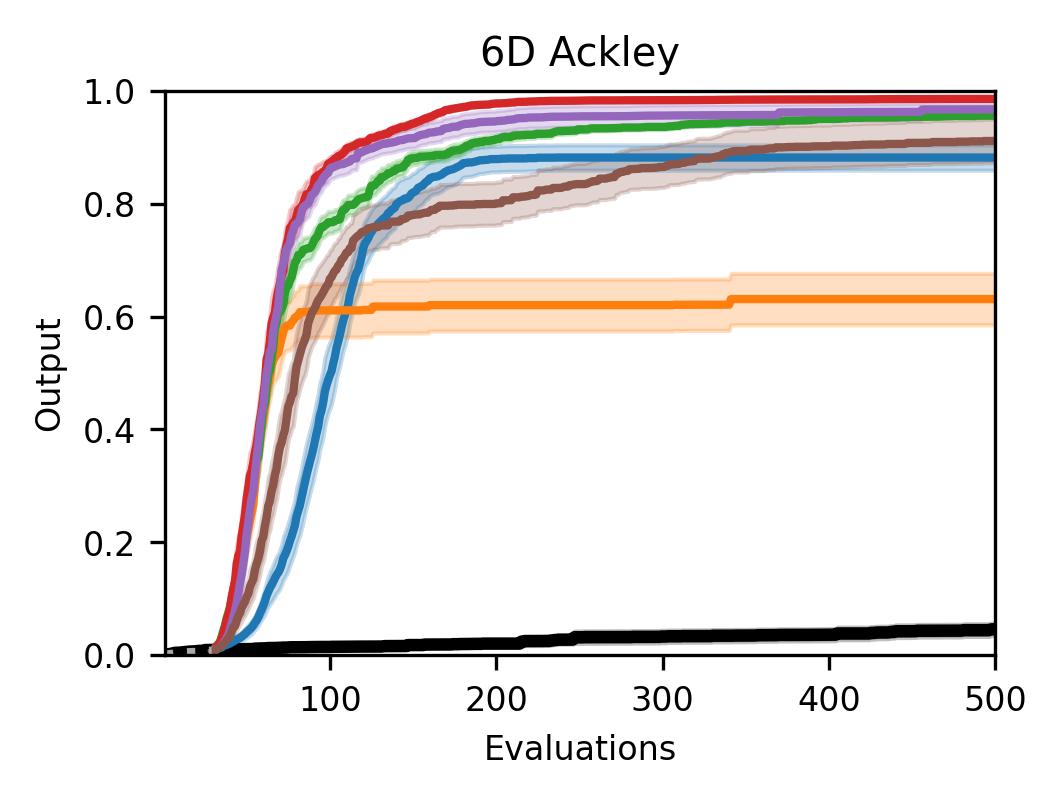}
        \label{fig:Subfigure 2c}
    \end{minipage}
    \begin{minipage}[t]{0.02\textwidth}
        \vspace{-50mm}
        (D)
    \end{minipage}
    \begin{minipage}[b]{0.48\textwidth}
        \centering
        \includegraphics[width=0.9\linewidth]{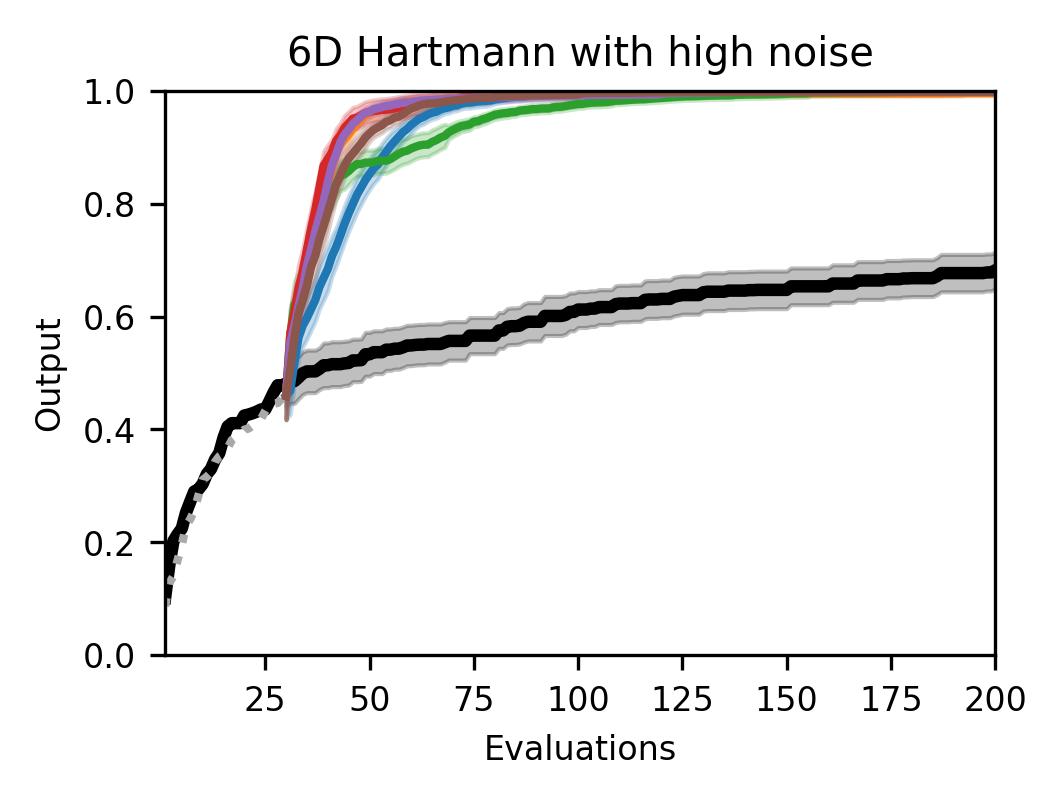}
        \label{fig:Subfigure 2d}
    \end{minipage} 

    \caption{Performance plots for analytical single-point acquisition functions with five initial starting points per input dimension. Solid lines represent the mean over the 50 runs while the shaded area represents the 95\% confidence intervals.}
    \label{fig:Figure 2}
\end{figure}

Figure \ref{fig:Figure 2} presents the best solutions, defined by the output value closest to the global optimum, found so far at each evaluation. The outputs are normalized to the unit range where 1.0 represents the global optimum. Most methods perform very well on the Griewank function, all reaching 1.00, and both variations of the Hartmann function with and without added noise (all reaching 1.00 and $>$0.97 respectively). The acquisition functions all find the optimum or a solution fairly close to the optimum within the allocated evaluation budget. However, PI and variable UCB typically take more evaluations to find a solution close to the optimum. All acquisition functions perform noticeably better than the Latin Hypercube benchmark and also exhibit much less variation over the 50 runs, as indicated by the 95\% confidence intervals. There is little difference between the Hartmann function with and without added noise, indeed the results are almost identical. The Ackley function on the other hand is more challenging. While the UCB methods still perform very well (all $>$ 0.96) the performance of the portfolio and PI decrease slightly (0.91 and 0.88 respectively) and EI performs considerably worse (0.63) with much more variability between runs.

\begin{table}
    \centering
    \caption{Averaged area under the curve with standard error for analytical single-point acquisition functions with five initial training points per input dimension.}
    \begin{tabular}{ c | c c c c }
         Method & Griewank & Hartmann & Noisy Hartmann & Ackley \\[0.2cm]
         \hline \hline \\
        PI & \makecell{0.99 \\ ($\pm$ 0.00)} & \makecell{0.91 \\ ($\pm$ 0.04)} & \makecell{0.95 \\ ($\pm$ 0.02)} & \makecell{0.76 \\ ($\pm$ 0.08)} \\[0.5cm]
        EI & \makecell{1.00 \\ ($\pm$ 0.00)} & \makecell{0.97 \\ ($\pm$ 0.02)} & \makecell{0.97 \\ ($\pm$ 0.02)} & \makecell{0.59 \\ ($\pm$ 0.15)} \\[0.5cm]
        UCB (variable) & \makecell{0.98 \\ ($\pm$ 0.01)} & \makecell{0.94 \\ ($\pm$ 0.03)} & \makecell{0.95 \\ ($\pm$ 0.03)} & \makecell{0.85 \\ ($\pm$ 0.01)} \\[0.5cm]
        UCB ($\beta$=5) & \makecell{1.00 \\ ($\pm$ 0.00)} & \makecell{0.97 \\ ($\pm$ 0.02)} & \makecell{0.98 \\ ($\pm$ 0.02)} & \makecell{0.90 \\ ($\pm$ 0.02)} \\[0.5cm]
        UCB ($\beta$=1) & \makecell{1.00 \\ ($\pm$ 0.00)} & \makecell{0.95 \\ ($\pm$ 0.04)} & \makecell{0.97 \\ ($\pm$ 0.02)} & \makecell{0.88 \\ ($\pm$ 0.05)} \\[0.5cm]
        Hedge & \makecell{1.00 \\ ($\pm$ 0.00)} & \makecell{0.95 \\ ($\pm$ 0.03)} & \makecell{0.97 \\ ($\pm$ 0.02)} & \makecell{0.77 \\ ($\pm$ 0.11)}
    \end{tabular}
    \label{table:Table 2}
\end{table}

Similar conclusions can be drawn from Table \ref{table:Table 2}, which presents the area under the curve (AUC) of each method. The AUC indicates how quickly the individual methods find solutions near the optimum. A perfect score (1.0) would indicate that the algorithm finds the optimum perfectly at the first iteration. The lower the score (a) the further the algorithm is away from the optimum and (b) the more evaluations are required for the algorithm to find promising solutions. While all methods score at least 0.98 for the Griewank function with a standard error of at most 0.01, the scores worsen slightly for the Hartmann function and significantly for the Ackley function. In particular, the AUC of EI for the Ackley function is very poor (0.59) and exhibits a high degree of variability between the 50 runs (standard error: 0.15). At the opposite end of the performance spectrum lie the optimistic methods, i.e. the UCB with variable and fixed $\beta$, that perform very well on all test functions with low standard errors. These results suggest that while the choice of acquisition function is less relevant for simple and moderately complex objective functions such as the Griewank or Hartmann functions, it is instrumental for solving challenging problems such as the Ackley function, especially if they are characterized by large flat areas. Here the optimistic acquisition functions are advantageous and should be preferred over the Hedge portfolio and improvement-based approaches. It should be noted that this may only be true for the specific portfolio defined previously. Implementing a different collection of acquisition functions could yield different results.

\subsubsection{Varying number of initial training points}
\label{sec3.2.2}

The training points used to initialize the Bayesian optimization algorithm directly effect the surrogate model, i.e. the Gaussian process, that in turn is a representation of the objective function. A larger number of training points yields a Gaussian process that will typically represent the objective function more closely as it incorporates more points and thus more information. However, the more evaluations of the total budget are allocated for these initial training points, the fewer points can be evaluated as part of the Bayesian optimization algorithm. This trade-off indicates that the number of training points (and by extension their selection) is an important choice in Bayesian optimization and should be considered thoroughly. This section explores this trade-off by taking the same experimental setup as the previous section but varying the number of initial training points. Overall, setups with one, five and ten initial points per input dimension of the objective function are considered.

\begin{figure}
    \begin{minipage}[t]{0.02\textwidth}
        \vspace{-50mm}
        (A)
    \end{minipage}
    \begin{minipage}[b]{0.48\textwidth}
        \centering
        \includegraphics[width=0.9\linewidth]{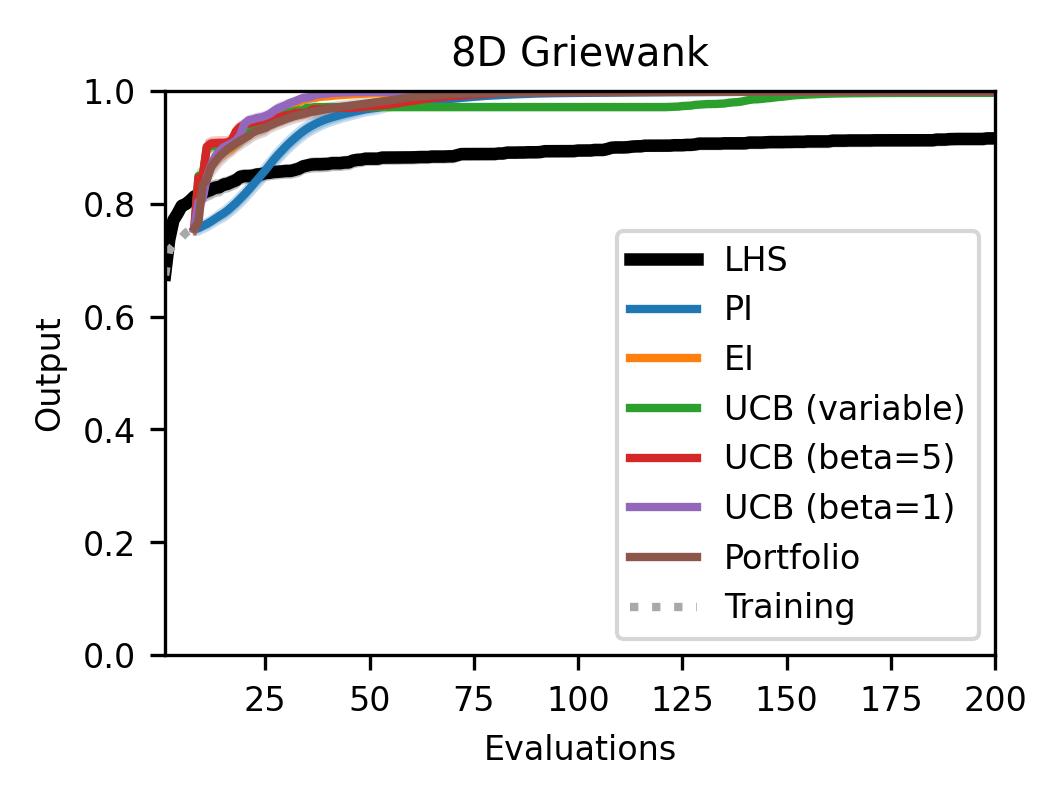}
        \label{fig:Subfigure 3a}
    \end{minipage}
    \begin{minipage}[t]{0.02\textwidth}
        \vspace{-50mm}
        (B)
    \end{minipage}
    \begin{minipage}[b]{0.48\textwidth}
        \centering
        \includegraphics[width=0.9\linewidth]{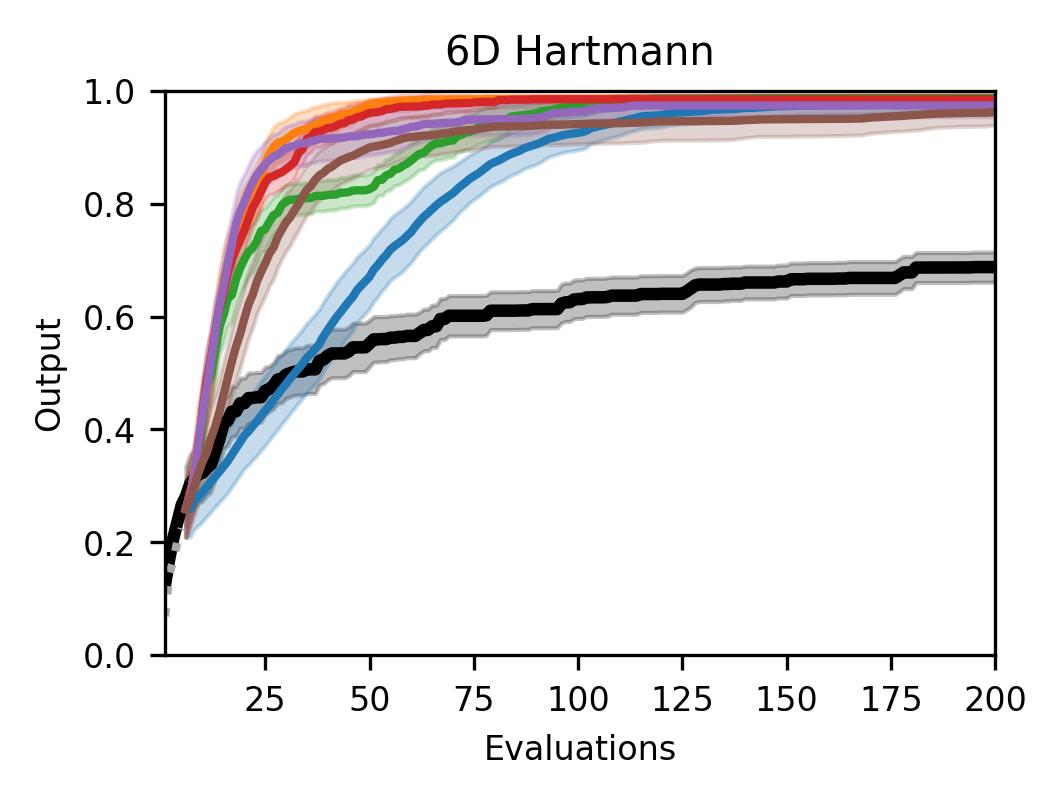}
        \label{fig:Subfigure 3b}
    \end{minipage}  
   
    \begin{minipage}[t]{0.02\textwidth}
        \vspace{-50mm}
        (C)
    \end{minipage}
    \begin{minipage}[b]{0.48\textwidth}
        \centering
        \includegraphics[width=0.9\linewidth]{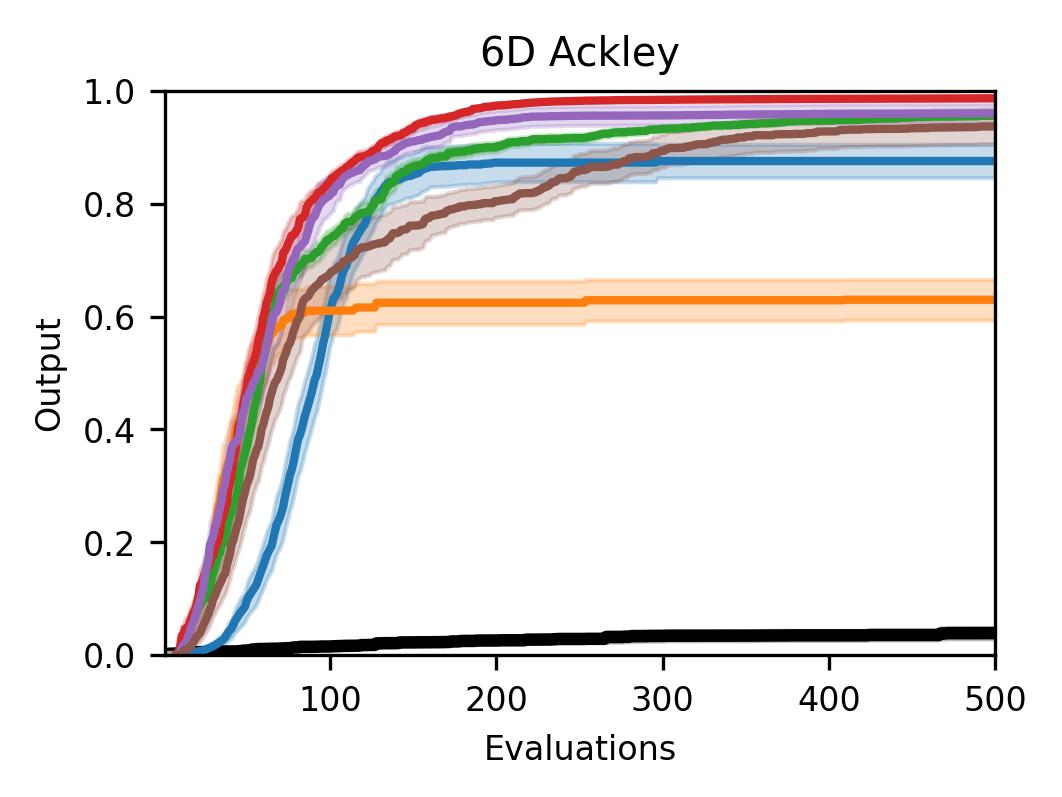}
        \label{fig:Subfigure 3c}
    \end{minipage}
    \begin{minipage}[t]{0.02\textwidth}
        \vspace{-50mm}
        (D)
    \end{minipage}
    \begin{minipage}[b]{0.48\textwidth}
        \centering
        \includegraphics[width=0.9\linewidth]{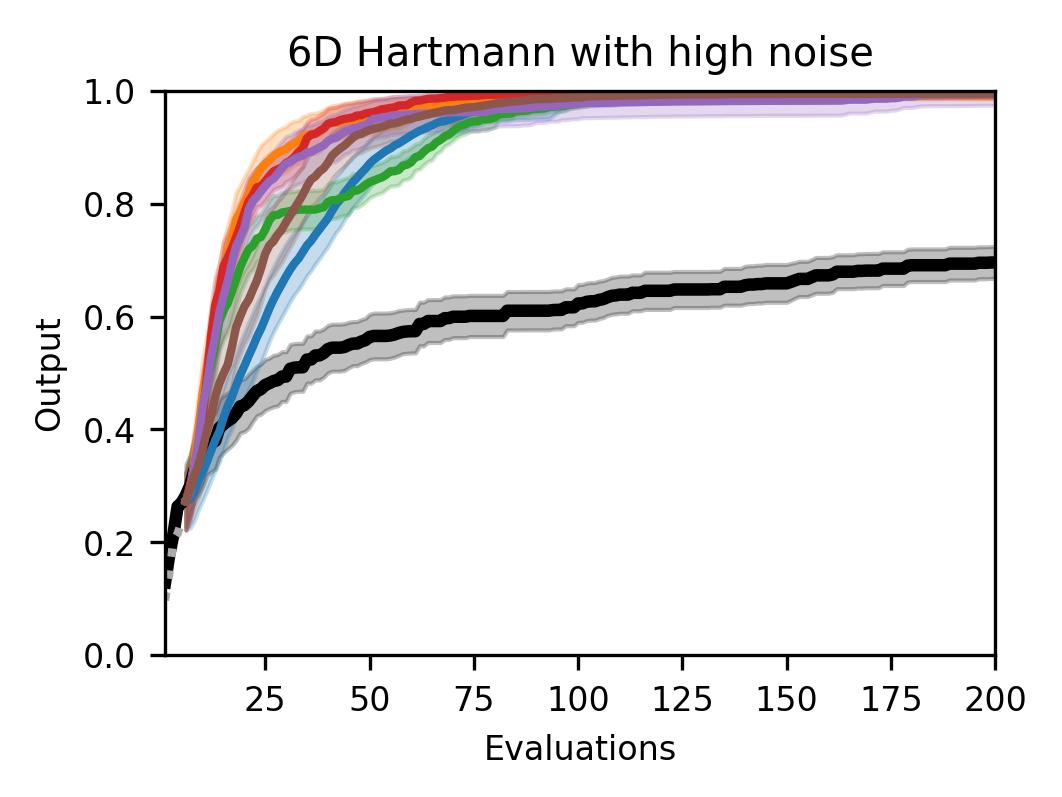}
        \label{fig:Subfigure 3d}
    \end{minipage} 

    \caption{Performance plots for analytical single-point acquisition functions with one initial starting point per input dimension. Solid lines represent the mean over the 50 runs while the shaded area represents the 95\% confidence intervals.}
    \label{fig:Figure 3}
\end{figure}

\begin{table}
    \centering
    \caption{Averaged area under the curve with standard error for analytical single-point acquisition functions with one initial training point per input dimension.}
    \begin{tabular}{ c | c c c c }
         Method & Griewank & Hartmann & Noisy Hartmann & Ackley \\[0.2cm]
         \hline \hline \\
        PI & \makecell{0.97 \\ ($\pm$ 0.01)} & \makecell{0.82 \\ ($\pm$ 0.09)} & \makecell{0.90 \\ ($\pm$ 0.06)} & \makecell{0.73 \\ ($\pm$ 0.11)} \\[0.5cm]
        EI & \makecell{0.99 \\ ($\pm$ 0.00)} & \makecell{0.95 \\ ($\pm$ 0.03)} & \makecell{0.95 \\ ($\pm$ 0.03)} & \makecell{0.58 \\ ($\pm$ 0.13)} \\[0.5cm]
        UCB (variable) & \makecell{0.97 \\ ($\pm$ 0.01)} & \makecell{0.91 \\ ($\pm$ 0.04)} & \makecell{0.91 \\ ($\pm$ 0.03)} & \makecell{0.82 \\ ($\pm$ 0.01)} \\[0.5cm]
        UCB ($\beta$=5) & \makecell{0.98 \\ ($\pm$ 0.00)} & \makecell{0.94 \\ ($\pm$ 0.03)} & \makecell{0.95 \\ ($\pm$ 0.03)} & \makecell{0.87 \\ ($\pm$ 0.02)} \\[0.5cm]
        UCB ($\beta$=1) & \makecell{0.99 \\ ($\pm$ 0.00)} & \makecell{0.93 \\ ($\pm$ 0.08)} & \makecell{0.93 \\ ($\pm$ 0.09)} & \makecell{0.85 \\ ($\pm$ 0.06)} \\[0.5cm]
        Hedge & \makecell{0.98 \\ ($\pm$ 0.01)} & \makecell{0.89 \\ ($\pm$ 0.11)} & \makecell{0.92 \\ ($\pm$ 0.05)} & \makecell{0.76 \\ ($\pm$ 0.09)} 
    \end{tabular}
    \label{table:Table 3}
\end{table}

\begin{figure}
    \begin{minipage}[t]{0.02\textwidth}
        \vspace{-50mm}
        (A)
    \end{minipage}
    \begin{minipage}[b]{0.48\textwidth}
        \centering
        \includegraphics[width=0.9\linewidth]{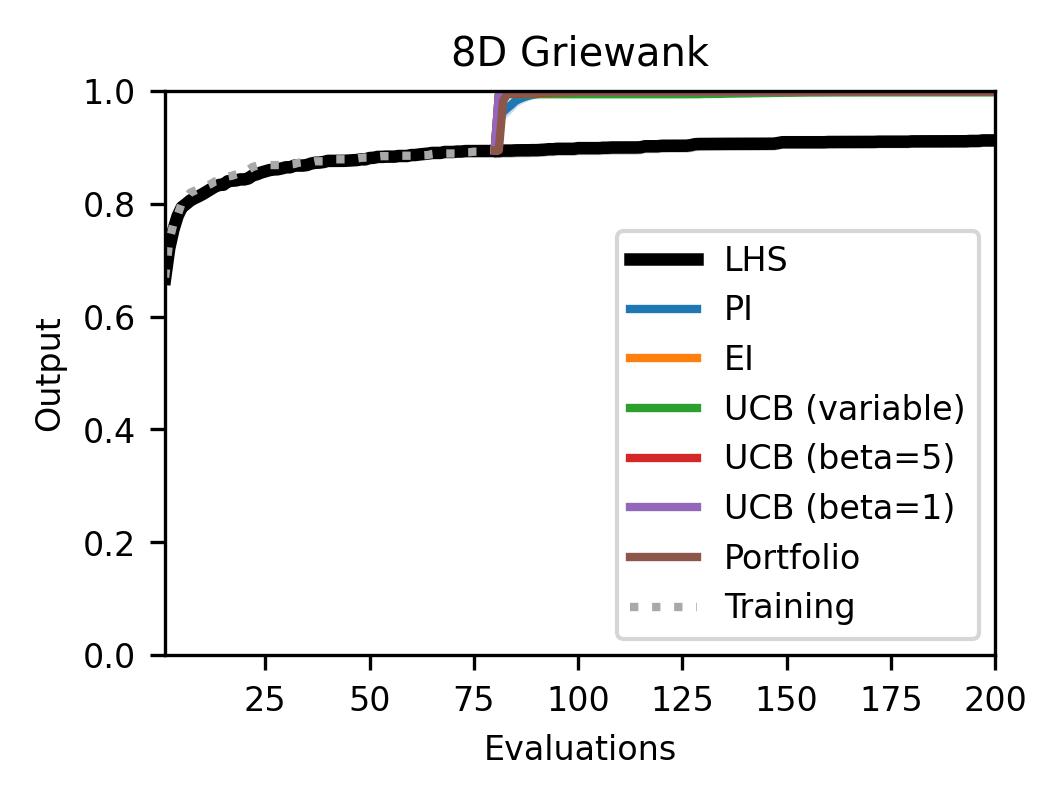}
        \label{fig:Subfigure 4a}
    \end{minipage}
    \begin{minipage}[t]{0.02\textwidth}
        \vspace{-50mm}
        (B)
    \end{minipage}
    \begin{minipage}[b]{0.48\textwidth}
        \centering
        \includegraphics[width=0.9\linewidth]{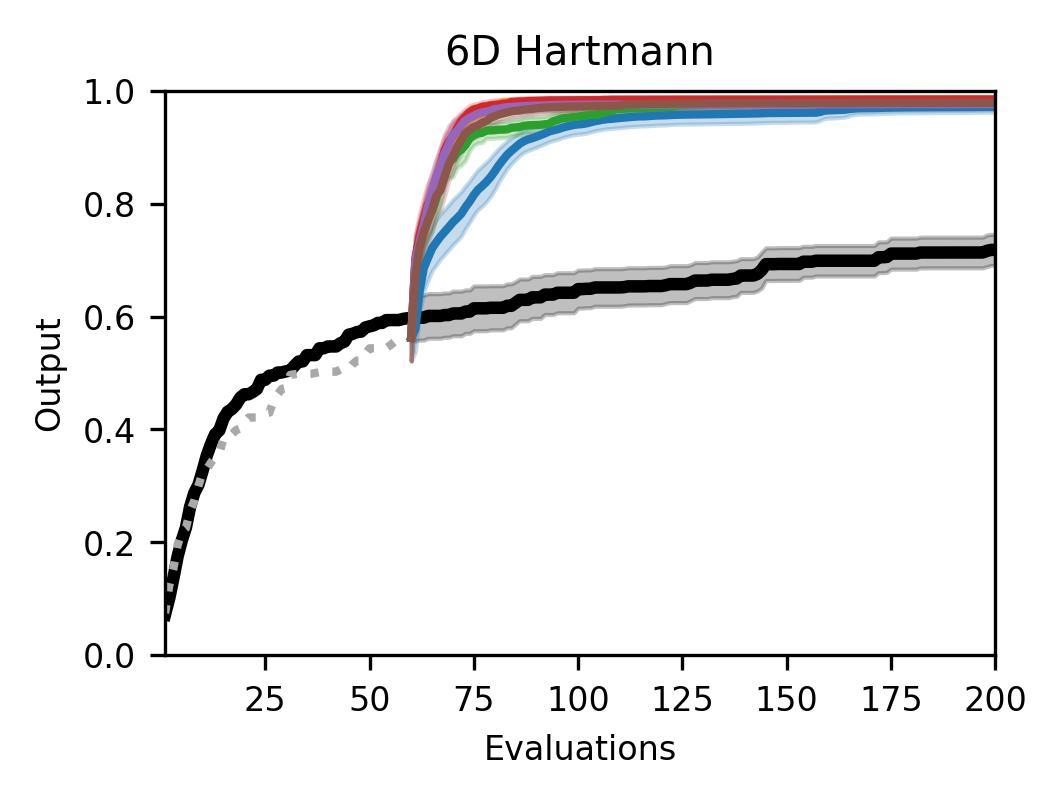}
        \label{fig:Subfigure 4b}
    \end{minipage}  
   
    \begin{minipage}[t]{0.02\textwidth}
        \vspace{-50mm}
        (C)
    \end{minipage}
    \begin{minipage}[b]{0.48\textwidth}
        \centering
        \includegraphics[width=0.9\linewidth]{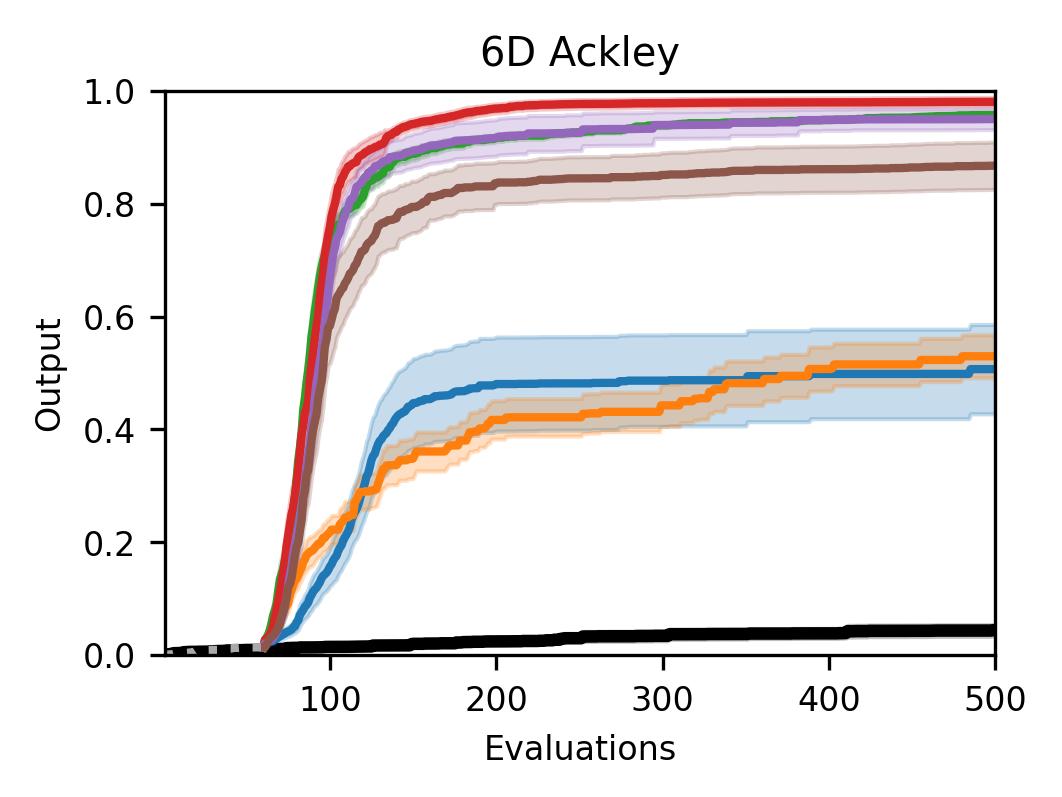}
        \label{fig:Subfigure 4c}
    \end{minipage}
    \begin{minipage}[t]{0.02\textwidth}
        \vspace{-50mm}
        (D)
    \end{minipage}
    \begin{minipage}[b]{0.48\textwidth}
        \centering
        \includegraphics[width=0.9\linewidth]{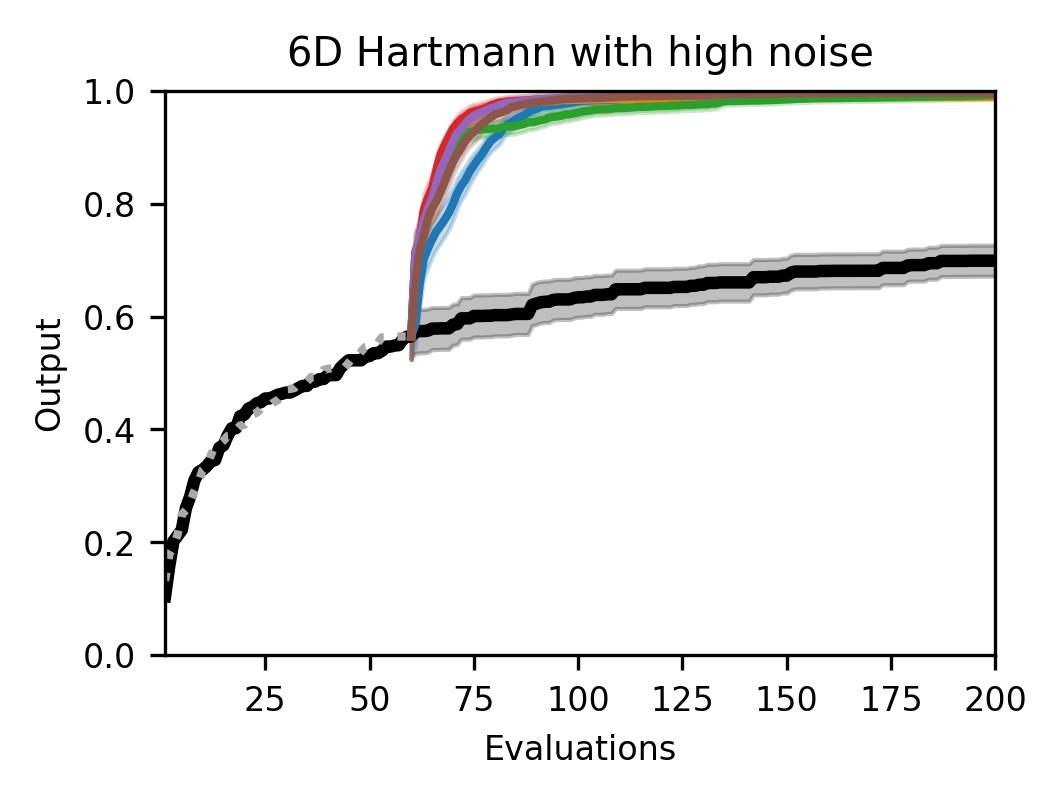}
        \label{fig:Subfigure 4d}
    \end{minipage} 

    \caption{Performance plots for analytical single-point acquisition functions with ten initial starting points per input dimension. Solid lines represent the mean over the 50 runs while the shaded area represents the 95\% confidence intervals.}
    \label{fig:Figure 4}
\end{figure}

Figures \ref{fig:Figure 3} and \ref{fig:Figure 4} depict the performance plots for the case with one and ten training points per dimension respectively. If the number of points is reduced to one point per dimension, the individual methods find solutions that are virtually identical to the results with five training points per dimension (see supplementary material in the appendix). EI still performs much worse than other methods for the Ackley function. Furthermore, most methods seem to find their best solution in a comparable or just slightly higher number of evaluations than before, as the AUC values in Table \ref{table:Table 3} show. This is expected, as using fewer initial training points means that the Bayesian optimization algorithm has less information at the earlier iterations than when using more initial training points. However, the difference in mean AUC is small and the results suggest that Bayesian optimization makes up for this lack of information quickly. PI and the Hedge portfolio have the highest decrease in performance on average for the Hartmann function with the mean AUC decreasing by 0.09 and 0.06 respectively. The variability between runs of the Hedge algorithm also rises, as indicated by a AUC standard error that is 0.08 higher. PI and the portfolio also perform worse for the Noisy Hartmann function, where the mean AUC decreased by 0.05 for both methods. Intuitively, this make sense as the surrogate model includes less information than before and thus it takes more evaluations to find a good solution. In early iterations, the individual methods deviate from one-another more than when five training points per dimension are used.

\begin{table}
    \centering
    \caption{Averaged area under the curve with standard error for analytical single-point acquisition functions with ten initial training points per input dimension.}
    \begin{tabular}{ c | c c c c }
         Method & Griewank & Hartmann & Noisy Hartmann & Ackley \\[0.2cm]
         \hline \hline \\
        PI & \makecell{1.00 \\ ($\pm$ 0.00)} & \makecell{0.93 \\ ($\pm$ 0.05)} & \makecell{0.96 \\ ($\pm$ 0.02)} & \makecell{0.43 \\ ($\pm$ 0.25)} \\[0.5cm]
        EI & \makecell{1.00 \\ ($\pm$ 0.00)} & \makecell{0.97 \\ ($\pm$ 0.02)} & \makecell{0.97 \\ ($\pm$ 0.02)} & \makecell{0.41 \\ ($\pm$ 0.07)} \\[0.5cm]
        UCB (variable) & \makecell{1.00 \\ ($\pm$ 0.00)} & \makecell{0.95 \\ ($\pm$ 0.03)} & \makecell{0.96 \\ ($\pm$ 0.03)} & \makecell{0.87 \\ ($\pm$ 0.01)} \\[0.5cm]
        UCB ($\beta$=5) & \makecell{1.00 \\ ($\pm$ 0.00)} & \makecell{0.97 \\ ($\pm$ 0.02)} & \makecell{0.98 \\ ($\pm$ 0.02)} & \makecell{0.91 \\ ($\pm$ 0.03)} \\[0.5cm]
        UCB ($\beta$=1) & \makecell{1.00 \\ ($\pm$ 0.00)} & \makecell{0.96 \\ ($\pm$ 0.03)} & \makecell{0.98 \\ ($\pm$ 0.03)} & \makecell{0.86 \\ ($\pm$ 0.08)} \\[0.5cm]
        Hedge & \makecell{1.00 \\ ($\pm$ 0.00)} & \makecell{0.96 \\ ($\pm$ 0.03)} & \makecell{0.97 \\ ($\pm$ 0.02)} & \makecell{0.78 \\ ($\pm$ 0.13)}
    \end{tabular}
    \label{table:Table 4}
\end{table}

If the number of starting points is increased to ten points per dimension, there is essentially no change to the case with just five starting points per dimension for the Griewank and the two Hartmann functions. For the Ackley function, however, the performance in terms of both the best solution found and the AUC worsen (Table \ref{table:Table 4}) for EI and, most significantly, for PI. The average best solution for the latter decreases by 0.37 to only 0.51 while EI decreases by 0.10 to 0.53. Both policies struggle with the large area of the test function that gives the same response value and is hence flat (see Section~\ref{sec3.1}). The optimistic policies and the portfolio perform much better and no real change is noticeable to results of Section~\ref{sec3.2.1}.

These results suggest that choosing a larger number of training points to initialize the Bayesian optimization algorithm cannot necessarily be equated with better performance and solutions. This is particularly true considering that there was no improved performance when increasing from five to ten points per input dimension. On the other hand, reducing the number of training points did not yield results that were much worse. Overall, a similar picture as in the previous section emerges: While all methods perform well on simpler problems, optimistic policies achieve the best results on the more challenging problems independent of the number of starting points. For the test functions we considered, five training points per dimension appeared to be sufficient, with no discernible improvement when moving to ten training points, and a small loss of performance when reducing to one training point per dimension.

\subsubsection{Monte Carlo single-point acquisition functions}
\label{sec3.2.3}

The previous experiments considered analytical acquisition functions. This section goes one step further and assesses the Monte Carlo approach outlined in Section~\ref{sec2.2}. As not all acquisition functions can be rewritten to suit such an approach, the experiments are restricted to PI, EI and UCB with variable and fixed $\beta$. Additionally, MES is introduced as a new method.

\begin{figure}
    \begin{minipage}[t]{0.02\textwidth}
        \vspace{-50mm}
        (A)
    \end{minipage}
    \begin{minipage}[b]{0.48\textwidth}
        \centering
        \includegraphics[width=0.9\linewidth]{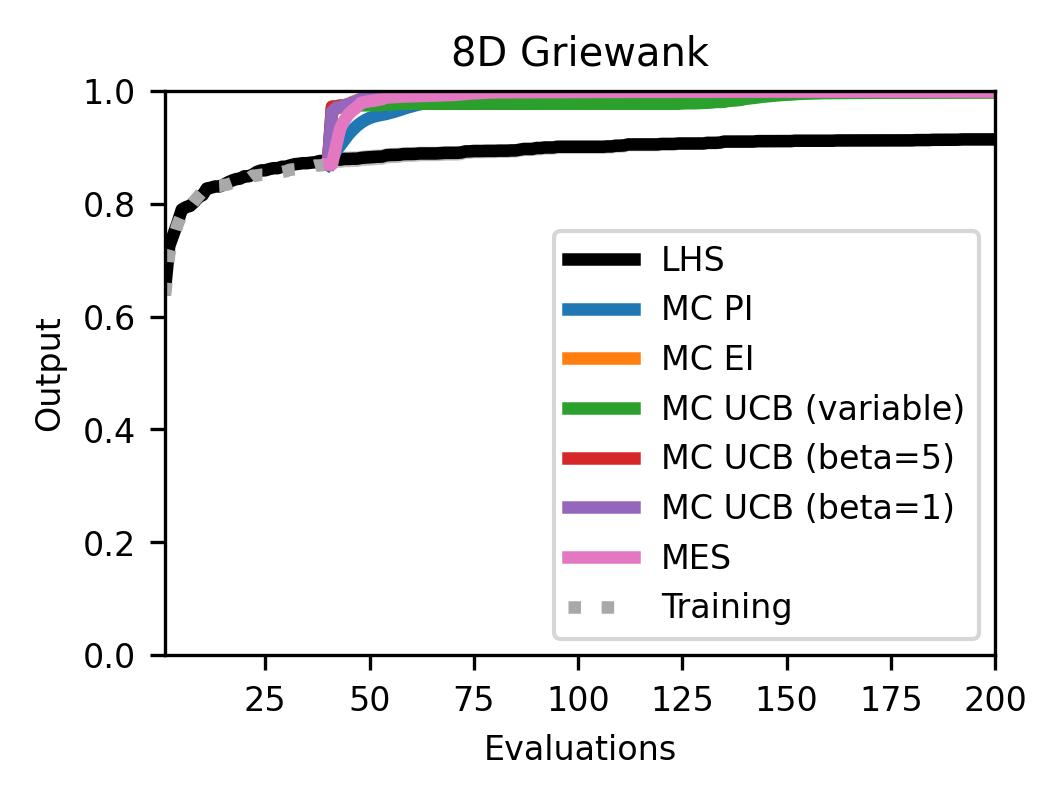}
        \label{fig:Subfigure 5a}
    \end{minipage}
    \begin{minipage}[t]{0.02\textwidth}
        \vspace{-50mm}
        (B)
    \end{minipage}
    \begin{minipage}[b]{0.48\textwidth}
        \centering
        \includegraphics[width=0.9\linewidth]{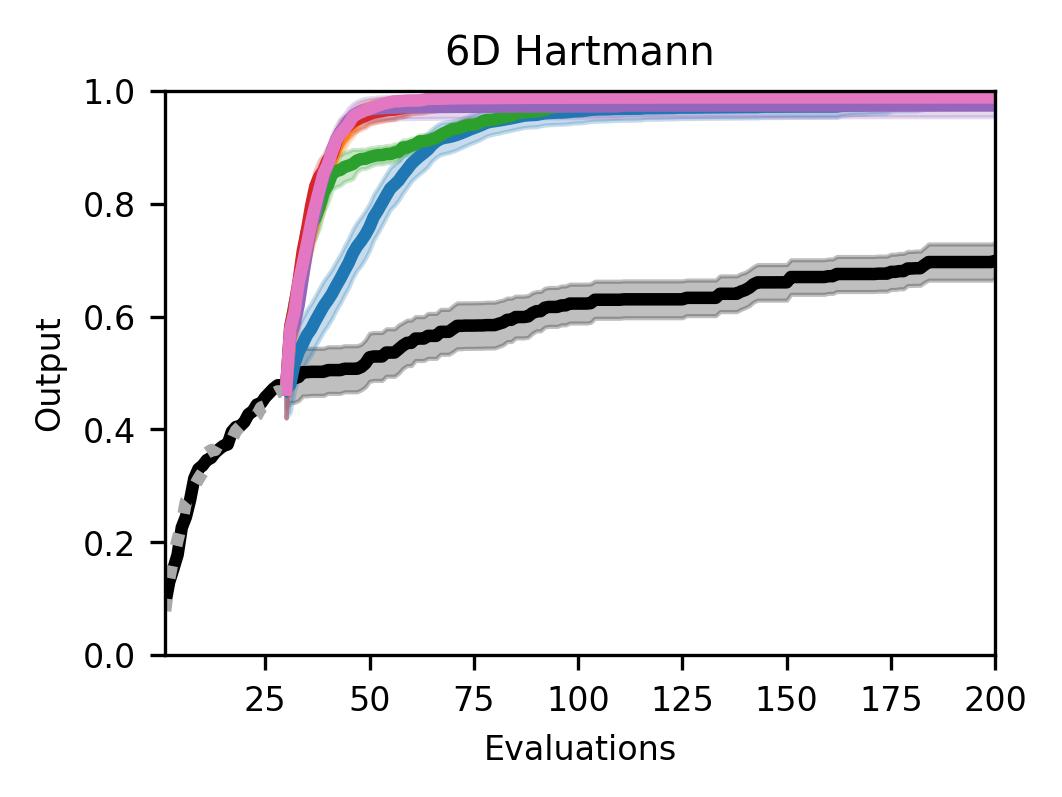}
        \label{fig:Subfigure 5b}
    \end{minipage}  
   
    \begin{minipage}[t]{0.02\textwidth}
        \vspace{-50mm}
        (C)
    \end{minipage}
    \begin{minipage}[b]{0.48\textwidth}
        \centering
        \includegraphics[width=0.9\linewidth]{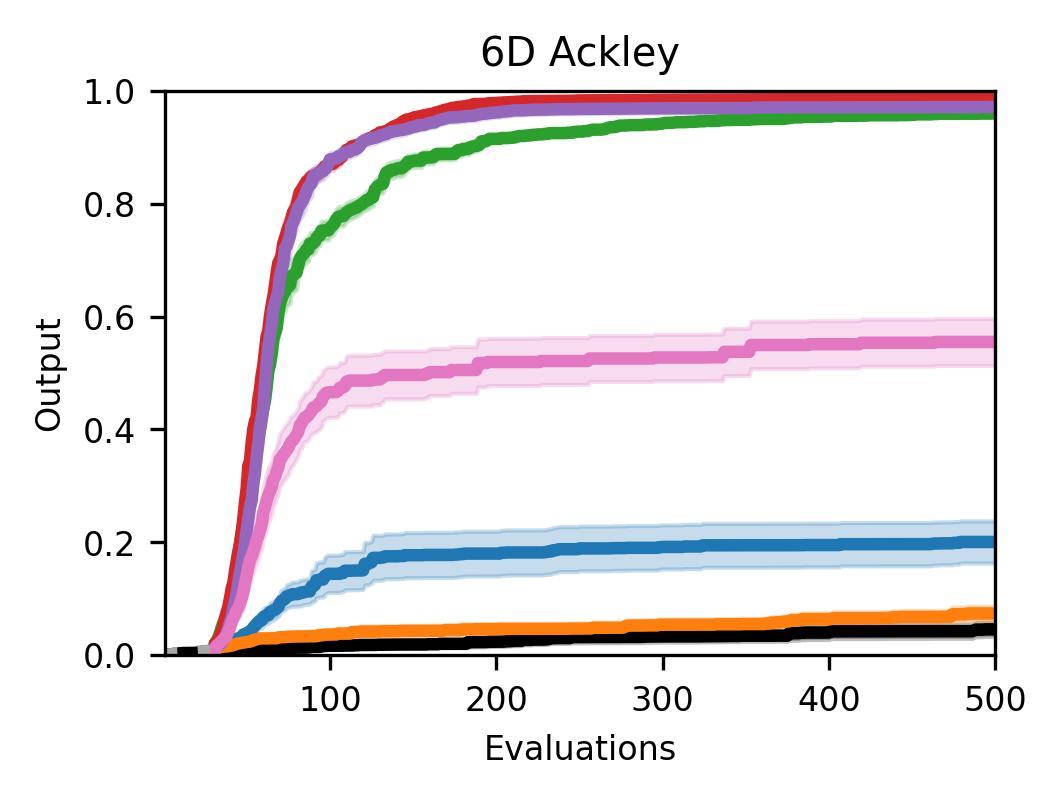}
        \label{fig:Subfigure 5c}
    \end{minipage}
        \begin{minipage}[t]{0.02\textwidth}
        \vspace{-50mm}
        (D)
    \end{minipage}
    \begin{minipage}[b]{0.48\textwidth}
        \centering
        \includegraphics[width=0.9\linewidth]{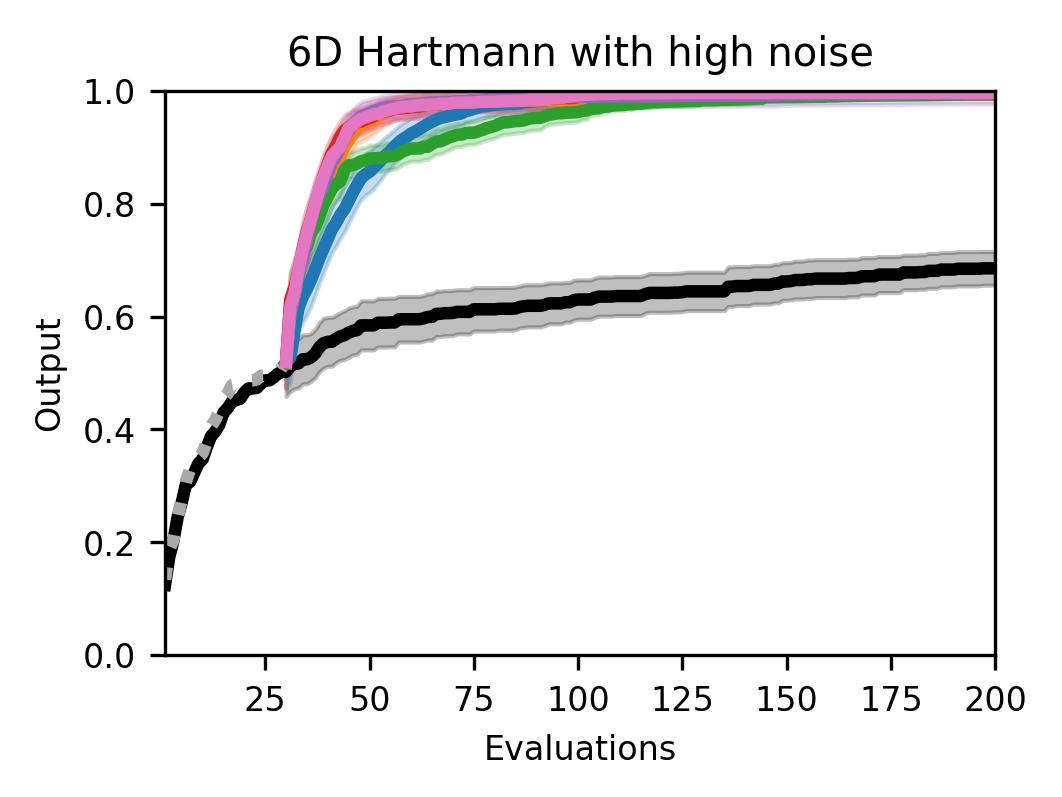}
        \label{fig:Subfigure 5d}
    \end{minipage} 

    \caption{Performance plots for Monte Carlo single-point acquisition functions  with five initial starting points per input dimension. Solid lines represent the mean over the 50 runs while the shaded areas represent the 95\% confidence intervals.}
    \label{fig:Figure 5}
\end{figure}

\begin{table}
    \centering
    \caption{Averaged area under the curve with standard error for Monte Carlo single-point acquisition functions with five initial training points per input dimension.}
    \begin{tabular}{ c | c c c c }
         Method & Griewank & Hartmann & Noisy Hartmann & Ackley \\[0.2cm]
         \hline \hline \\
        MC PI & \makecell{0.99 \\ ($\pm$ 0.00)} & \makecell{0.92 \\ ($\pm$ 0.06)} & \makecell{0.95 \\ ($\pm$ 0.04)} & \makecell{0.17 \\ ($\pm$ 0.12)} \\[0.5cm]
        MC EI & \makecell{1.00 \\ ($\pm$ 0.00)} & \makecell{0.97 \\ ($\pm$ 0.02)} & \makecell{0.97 \\ ($\pm$ 0.04)} & \makecell{0.05 \\ ($\pm$ 0.03)} \\[0.5cm]
        MC UCB (variable) & \makecell{0.98 \\ ($\pm$ 0.01)} & \makecell{0.95 \\ ($\pm$ 0.02)} & \makecell{0.95 \\ ($\pm$ 0.04)} & \makecell{0.86 \\ ($\pm$ 0.01)} \\[0.5cm]
        MC UCB ($\beta$=5) & \makecell{1.00 \\ ($\pm$ 0.00)} & \makecell{0.97 \\ ($\pm$ 0.02)} & \makecell{0.98 \\ ($\pm$ 0.03)} & \makecell{0.91 \\ ($\pm$ 0.01)} \\[0.5cm]
        MC UCB ($\beta$=1) & \makecell{1.00 \\ ($\pm$ 0.00)} & \makecell{0.96 \\ ($\pm$ 0.07)} & \makecell{0.97 \\ ($\pm$ 0.07)} & \makecell{0.89 \\ ($\pm$ 0.03)} \\[0.5cm]
        MES & \makecell{0.99 \\ ($\pm$ 0.00)} & \makecell{0.97 \\ ($\pm$ 0.02)} & \makecell{0.97 \\ ($\pm$ 0.03)} & \makecell{0.49 \\ ($\pm$ 0.12)} 
    \end{tabular}
    \label{table:Table 5}
\end{table}

For the Griewank and both Hartmann functions all results are almost identical to the analytical case except for the variability between runs which increased slightly for some of the methods. However, Figure \ref{fig:Figure 5} clearly shows that the performance for PI and EI tested on the Ackley function decreased significantly. The average best solutions decreased by 0.68 and 0.56 and the average AUC (Table \ref{table:Table 5}) decreased by 0.59 and 0.54 respectively. Table \ref{table:Table 5} shows that MES performs well on the Griewank and Hartmann functions, all reaching an AUC of above 0.97 with low standard errors. When it comes to the more complex Ackley function, however, MES performs much worse. With a mean AUC of 0.49 its performance ranks below the optimistic acquisition functions but still above the improvement-based methods. Earlier sections showed that improvement-based policies (in particular EI) perform poorly on the Ackley function when using analytical acquisition functions, but the results from this comparison show that their performance suffers even more severely when using the Monte Carlo approach. One reason for this could be that the Monte Carlo variants are essentially an approximation of the analytical acquisition functions as discussed in Section~\ref{sec2.2}. However, there seems to be little to no change when using optimistic policies. This suggests that, similar to previous sections, optimistic policies should be preferred when optimizing complex and challenging objective functions with large flat areas using Monte Carlo acquisition functions.

\subsubsection{Multi-point acquisition functions}
\label{sec3.2.4}

The previous sections focused on single-point approaches, where each iteration of the Bayesian optimization algorithm yields one new point that is sampled from the objective function before the next iteration is started. While this makes sense for objective functions that can be evaluated quickly or do not allow parallel evaluations, it might slow down the optimization process needlessly when objective functions take a long time to evaluate and allow parallel evaluations. Thus, this section implements multi-point acquisition functions that propose a batch of points at each iteration, which are then evaluated simultaneously before the next iteration.

Section~\ref{sec2.2} outlined how Monte Carlo approaches using the reparameterization trick can be extended to compute batches naturally. MES does not use reparameterization and is thus not naturally extendable to the multi-point setting. Hence, we considered the same acquisition functions as in the previous section but this time for a batch size of five points. Each of the acquisition functions is optimized with two different methods, sequentially and jointly. The latter optimizes the acquisition function based on a joint distribution that includes training points and new points and optimizes all new points in a single step. The former recomputes the joint distribution each time a new point is found and only optimizes with respect to the newest point. For example, for a batch of five points this process is repeated five times \citep{botorch}. This approach, also known as greedy optimization, might be preferable and can yield better results \citep{Wilson2018}. While analytical functions cannot be extended to the multi-point case as easily as the Monte Carlo evaluations, there are some frameworks that allow the computation of batches. This section considers the Constant Liar approach with a lie equivalent to the minimal (CL min) and maximal (CL max) value so far \citep{Ginsbourger} and the GP-BUCB algorithm, an extension of the UCB function \citep{Desautels2014}. For more details see Section~\ref{sec2.2}. 

\begin{figure}
    \begin{minipage}[t]{0.02\textwidth}
        \vspace{-50mm}
        (A)
    \end{minipage}
    \begin{minipage}[b]{0.48\textwidth}
        \centering
        \includegraphics[width=0.9\linewidth]{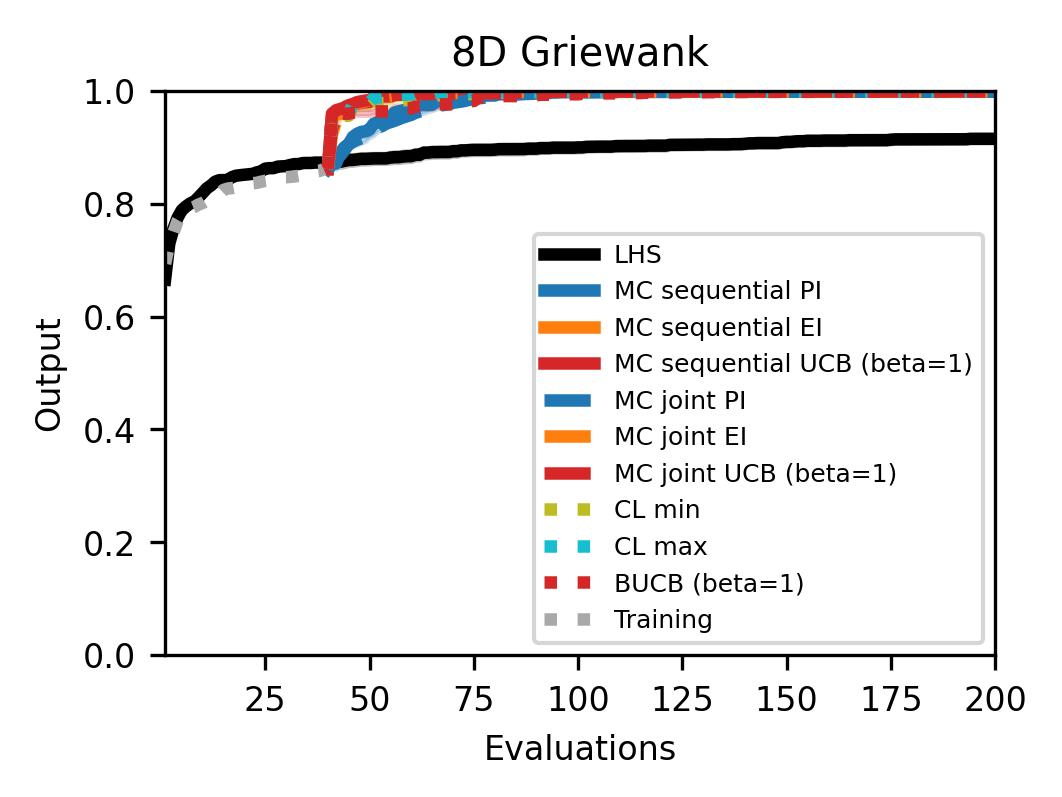}
        \label{fig:Subfigure 6a}
    \end{minipage}
    \begin{minipage}[t]{0.02\textwidth}
        \vspace{-50mm}
        (B)
    \end{minipage}
    \begin{minipage}[b]{0.48\textwidth}
        \centering
        \includegraphics[width=0.9\linewidth]{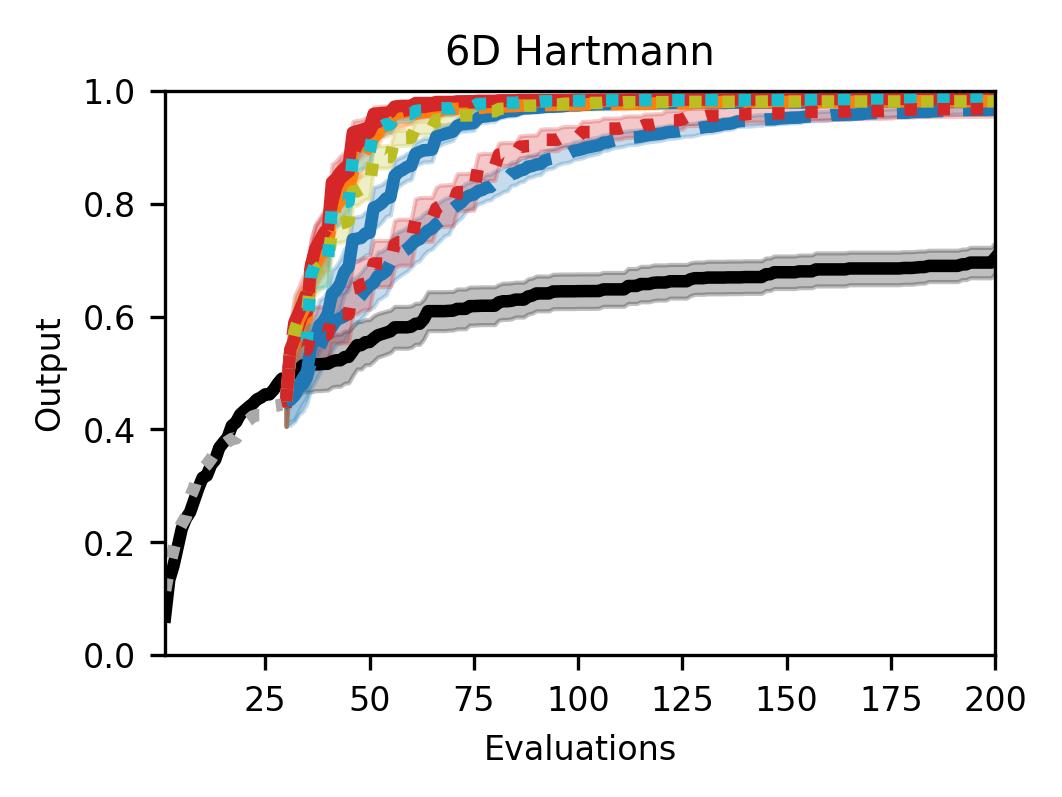}
        \label{fig:Subfigure 6b}
    \end{minipage}  
   
    \begin{minipage}[t]{0.02\textwidth}
        \vspace{-50mm}
        (C)
    \end{minipage}
    \begin{minipage}[b]{0.48\textwidth}
        \centering
        \includegraphics[width=0.9\linewidth]{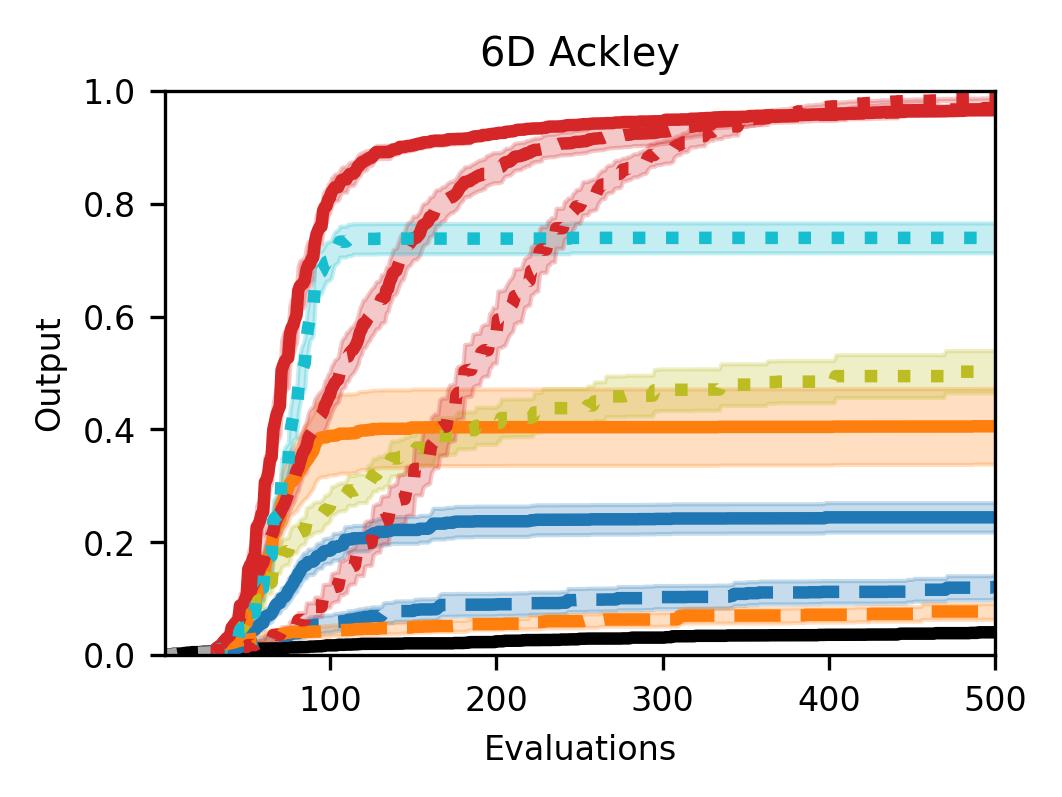}
        \label{fig:Subfigure 6c}
    \end{minipage}
    \begin{minipage}[t]{0.02\textwidth}
        \vspace{-50mm}
        (D)
    \end{minipage}
    \begin{minipage}[b]{0.48\textwidth}
        \centering
        \includegraphics[width=0.9\linewidth]{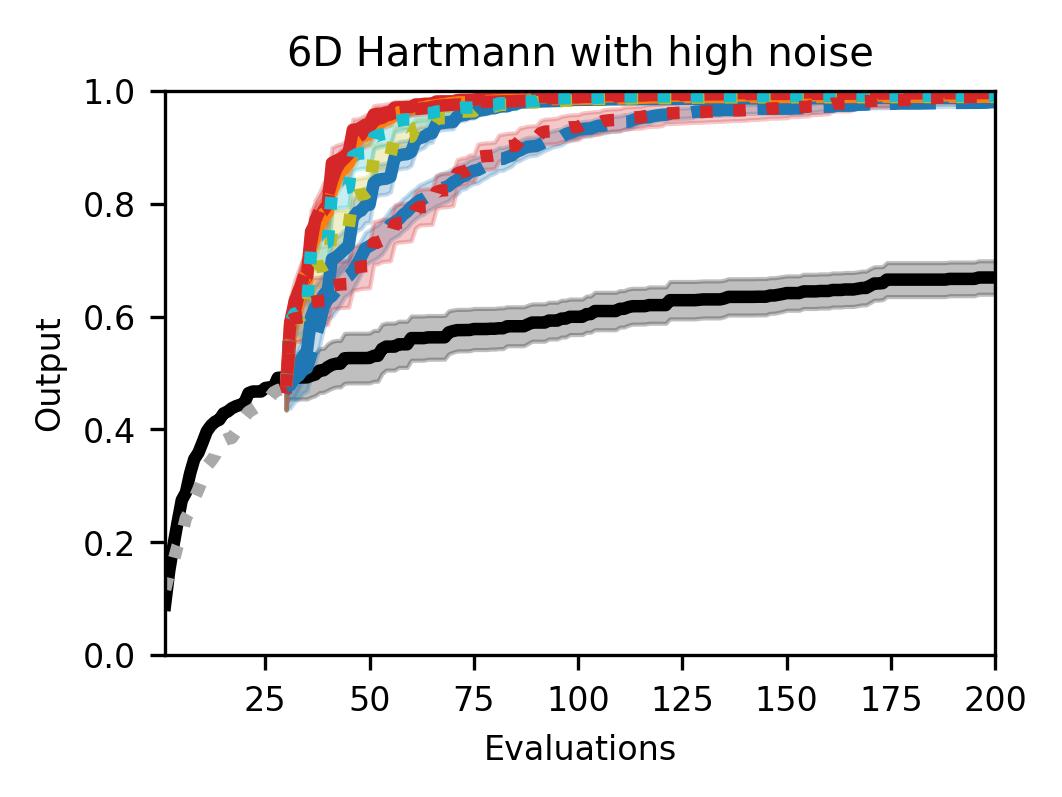}
        \label{fig:Subfigure 6d}
    \end{minipage} 

    \caption{Performance plots for multi-point acquisition functions with five initial training points per input dimension. Solid lines represent the mean over the 50 runs while shaded areas represent the 95\% confidence intervals. PI in blue, EI in orange, UCB in red.}
    \label{fig:Figure 6}
\end{figure}

Figure \ref{fig:Figure 6} shows the results of some multi-point acquisition functions. They essentially find identical solutions, on average, to the single-point approach for the Griewank function and Hartmann function with and without noise. Only selected methods are shown in the plot. Particularly, two variations of the UCB approach are not included as they are almost identical to the UCB with $\beta$=1 (see supplementary material in the appendix for results of all methods). While there are no changes to the AUC for the Griewank function, there are some differences for the Hartmann functions that suggest that methods require a different number of evaluations to find the best value, as Table \ref{table:Table 6} shows. Although the joint Monte Carlo approach for PI and UCB (variable and $\beta$=5) have a lower mean AUC, the decrease is less severe than for the GP-BUCB methods that all decrease by 0.08 to 0.10. The other methods perform comparably to the single-point case and the variability between the runs of the sequential and joint Monte Carlo UCB with $\beta$=1 even decreases. For the noisy Hartmann test function, the results are very similar to the Hartmann function without noise. In general, over all experiments there was no real drop in performance that can be declared as a result of adding noise up to 2.4\% of the overall objective function range.

\begin{table}
    \centering
    \caption{Averaged area under the curve with standard errors for multi-point acquisition function with five initial training points per input dimension.}
    \begin{tabular}{ c c | c c c c }
         Type & Method & Griewank & Hartmann & Noisy Hartmann & Ackley \\[0.2cm]
         \hline \hline 
         & & & & & \\
        \parbox{0mm}{\multirow{12}{10pt}{\rotatebox{90}{Sequential Monte Carlo}}} & PI & \makecell{0.99 \\ ($\pm$ 0.00)} & \makecell{0.92 \\ ($\pm$ 0.03)} & \makecell{0.94 \\ ($\pm$ 0.02)} & \makecell{0.22 \\ ($\pm$ 0.08)} \\[0.5cm]
        & EI & \makecell{1.00 \\ ($\pm$ 0.00)} & \makecell{0.95 \\ ($\pm$ 0.03)} & \makecell{0.97 \\ ($\pm$ 0.02)} & \makecell{0.37 \\ ($\pm$ 0.22)} \\[0.5cm]
        & \makecell{UCB \\ (variable)} & \makecell{0.98 \\ ($\pm$ 0.01)} & \makecell{0.94 \\ ($\pm$ 0.03)} & \makecell{0.94 \\ ($\pm$ 0.03)} & \makecell{0.84 \\ ($\pm$ 0.01)} \\[0.5cm]
        & \makecell{UCB \\ ($\beta$=5)} & \makecell{0.99 \\ ($\pm$ 0.00)} & \makecell{0.95 \\ ($\pm$ 0.02)} & \makecell{0.96 \\ ($\pm$ 0.02)} & \makecell{0.88 \\ ($\pm$ 0.02)} \\[0.5cm]
        & \makecell{UCB \\ ($\beta$=1)} & \makecell{1.00 \\ ($\pm$ 0.00)} & \makecell{0.96 \\ ($\pm$ 0.02)} & \makecell{0.97 \\ ($\pm$ 0.02)} & \makecell{0.86 \\ ($\pm$ 0.02)} \\[0.5cm]
         \hline
        & & & & & \\[0.15cm]
         \parbox{0mm}{\multirow{12}{*}{\rotatebox[origin=c]{90}{Joint Monte Carlo}}} & PI & \makecell{0.99 \\ ($\pm$ 0.00)} & \makecell{0.86 \\ ($\pm$ 0.05)} & \makecell{0.89 \\ ($\pm$ 0.04)} & \makecell{0.09 \\ ($\pm$ 0.06)} \\[0.5cm]
         & EI & \makecell{1.00 \\ ($\pm$ 0.00)} & \makecell{0.95 \\ ($\pm$ 0.03)} & \makecell{0.96 \\ ($\pm$ 0.02)} & \makecell{0.06 \\ ($\pm$ 0.03)} \\[0.5cm]
         & \makecell{UCB \\ (variable)} & \makecell{0.97 \\ ($\pm$ 0.02)} & \makecell{0.89 \\ ($\pm$ 0.04)} & \makecell{0.90 \\ ($\pm$ 0.04)} & \makecell{0.81 \\ ($\pm$ 0.02)} \\[0.5cm]
         & \makecell{UCB \\ ($\beta$=5)} & \makecell{0.99 \\ ($\pm$ 0.00)} & \makecell{0.94 \\ ($\pm$ 0.03)} & \makecell{0.95 \\ ($\pm$ 0.03)} & \makecell{0.87 \\ ($\pm$ 0.02)} \\[0.5cm]
         & \makecell{UCB \\ ($\beta$=1)} & \makecell{1.00 \\ ($\pm$ 0.00)} & \makecell{0.96 \\ ($\pm$ 0.02)} & \makecell{0.97 \\ ($\pm$ 0.02)} & \makecell{0.78 \\ ($\pm$ 0.06)} \\[0.5cm]
         \hline
         & & & & & \\[0.15cm]
         \parbox{0mm}{\multirow{12}{*}{\rotatebox[origin=c]{90}{Analytical}}} & CL min & \makecell{0.99 \\ ($\pm$ 0.00)} & \makecell{0.94 \\ ($\pm$ 0.02)} & \makecell{0.95 \\ ($\pm$ 0.02)} & \makecell{0.40 \\ ($\pm$ 0.08)} \\[0.5cm]
         & CL max & \makecell{1.00 \\ ($\pm$ 0.00)} & \makecell{0.95 \\ ($\pm$ 0.03)} & \makecell{0.96 \\ ($\pm$ 0.03)} & \makecell{0.67 \\ ($\pm$ 0.09)} \\[0.5cm]
         & \makecell{BUCB \\ (variable)} & \makecell{0.98 \\ ($\pm$ 0.02)} & \makecell{0.85 \\ ($\pm$ 0.05)} & \makecell{0.85 \\ ($\pm$ 0.05)} & \makecell{0.65 \\ ($\pm$ 0.03)} \\[0.5cm]
         & \makecell{BUCB \\ ($\beta$=5)} & \makecell{0.98 \\ ($\pm$ 0.00)} & \makecell{0.88 \\ ($\pm$ 0.05)} & \makecell{0.90 \\ ($\pm$ 0.05)} & \makecell{0.66 \\ ($\pm$ 0.05)} \\[0.5cm]
         & \makecell{BUCB \\ ($\beta$=1)} & \makecell{0.99 \\ ($\pm$ 0.00)} & \makecell{0.88 \\ ($\pm$ 0.06)} & \makecell{0.90 \\ ($\pm$ 0.06)} & \makecell{0.65 \\ ($\pm$ 0.05)}
    \end{tabular}
    \label{table:Table 6}
\end{table}

Most methods find similar best solutions to the Ackley test function as their single-point counterparts. However, there are some changes in the improvement-based policies: the sequential Monte Carlo EI and PI, and CL max, find solutions that are better than before (by 0.33, 0.04 and 0.11 respectively). At the opposite end, CL min and joint Monte Carlo PI provide inferior solutions in the batched case (0.13 and 0.08 respectively). Looking at how quickly the individual methods find good values on average, i.e. the AUC, it is clear that the analytical multi-point and joint Monte Carlo methods perform worse than the single-point implementations for the Ackley function: the AUC of all analytical multi-point methods worsen by 0.19 -- 0.24, except for CL max that improved by 0.08. Similarly, all joint methods provide poorer performance than in the sequential single-point optimization where UCB with $\beta$=1 sees the largest drop of 0.11. EI is the exception and stays about the same. The biggest improvement is provided by the sequential Monte Carlo EI with an increase of 0.32. However, this improvement is mainly caused by the very poor performance of the single-point Monte Carlo EI acquisition function. Furthermore, this approach comes with a higher variability, as the standard error of the AUC rises by 0.19.

In terms of the best solutions found, multi-point methods present a good alternative to single-point acquisition functions. They generally find best solutions comparable to the single-point approach, but for a slightly larger number of objective function evaluations. However, while it requires more evaluations, the multi-point approach would still be faster when computing batches in parallel. Multi-point acquisition functions will, therefore, be most beneficial for expensive to evaluate objective functions that can be evaluated in parallel. With some exceptions, this benefit requires a higher number of evaluations until solutions comparable to single-point methods are found. Overall, the optimistic methods, combined with sequential optimization, appear to be favorable over the rest, as the red lines in Figure \ref{fig:Figure 6} clearly show.

\subsection{Discussion}
\label{sec3.3}

Six main conclusions can be drawn from the simulation results presented in detail above. This section discusses these findings and relates them back to the applied problems that motivated this work, i.e. optimizing experiments in engineering, particularly in fluid dynamics, and tuning hyperparameters of neural networks and other statistical models.

The first findings concern the choice of acquisition functions related to the complexity of the problem. The results show that this choice is less important when optimizing simpler objective functions. Improvement-based, optimistic, and information-based acquisition functions, as well as portfolios, performed well on a wide range of synthetic test functions with up to ten input dimensions (see supplementary material in the appendix for more examples to reinforce this result.). However, for more complex functions, such as the Ackley function, the optimistic methods performed significantly better than the rest and thus should be favored. For the Bayesian optimization algorithm, this means that all acquisition functions considered in this paper can yield good results. But the choice of acquisition function must be considered more carefully with increasing complexity of the objective function. This indicates the importance of expert knowledge when applying Bayesian optimization to a specific problem, such as a drag reduction problem in fluid dynamics. Basing the choice of acquisition function on specific knowledge about the suspected complexity of the objective function could potentially increase the performance of the Bayesian optimization algorithm significantly. In cases where no expert knowledge or other information about the objective function is accessible, results suggest that the optimistic methods are a good starting point.

Secondly, the results suggest that the number of initial training points is not critical in achieving good performance from the algorithm. In fact, there is only a little difference when increasing the number of starting points from one point per input dimension to ten points per dimension. While the performance of the acquisition functions differs initially for the former case, they still find comparable results to the runs with five or ten points per input dimension over all iterations. This means that Bayesian optimization explores the space efficiently even when provided with only a few training points. Deciding on the number of starting points is an important decision in problems where evaluating a point is expensive. For example, when evaluating a set of hyperparameters for a complex model such as a deep neural network, the model has to be trained anew for each set of hyperparameters, racking up costs in time and computing resources. In most cases, the problem boils down to dividing a predefined budget into two parts: evaluations to initialize the algorithm and evaluations for points proposed by the Bayesian optimization algorithm. The decision of how many points to allocate for the training data is important as using too many could mean that the Bayesian optimization algorithm does not have sufficient evaluations available to find a good solution, i.e. the budget is exhausted before a promising solution is found. The simulations in this paper suggest that this decision might not be as complex as it initially seems, as only a few training points are necessary for the algorithm to achieve good results. Taking this approach of using only a small number of evaluations for training points saves more evaluations for the optimization itself, thus increasing the chances of finding a good solution.

The third finding regards the information-based acquisition functions. Through the range of different simulations, PI and EI failed to find good results for the Ackley function. The reason for this is the large area of the parameter space where all response values are identical and thus the response surface is flat. As discussed in Section~\ref{sec2.2}, improvement-based acquisition functions propose a point that is most likely to improve upon the previous best point. With a flat function like the Ackley function it is likely that all of the initial starting points fall into the flat area (especially in higher dimensions where the flat area grows even quicker). The posterior mean of the Gaussian process will then be very flat, and will in turn predict that the underlying objective function is flat as well. This leads to a very flat acquisition function, as most input points will have a small likelihood of improving upon the previous best points. \citet{gramacy2020surrogates} mentions problems when optimizing a flat EI acquisition function that result in the evaluation of points that are not optimal, which is especially problematic when the shape of the objective function is not known and flat regions cannot be ruled out {\em a priori}. If such properties are possible in a particular applied problem, the results suggest that using a different acquisition function, such as an optimistic policy, achieves better results. This again highlights the importance of expert knowledge for applications to specific problems.

Fourthly, the simulations show that Monte Carlo acquisition functions yield comparable results to analytical functions. These functions use Monte Carlo sampling to compute the acquisition functions, instead of analytically solving them. Utilizing sampling to compute a function that can also be solved analytically might not appear useful at first glance as it is essentially just an approximation of the analytical results. However, this approach makes it straightforward to compute batches of candidate points (as outlined in Section~\ref{sec2.2}), which presents the foundation for the next finding.

The fifth finding suggests that multi-point acquisition functions perform comparably to single-point approaches. All methods found good solutions across the range of considered problems, with the exception of improvement-based methods for the Ackley function, as discussed previously. Multi-point approaches are particularly beneficial for cases in which the objective function is expensive to evaluate and allows parallel evaluations (e.g. in high-fidelity turbulence resolving simulations \citep{Mahfoze2019}). For these problems, the total time required to conclude the full experiment can be reduced as multiple points are evaluated in parallel each time. However, more evaluations in total might be required to achieve results comparable to the single-point approach. This method is particularly advantageous for applications involving simulations that can be evaluated in parallel, e.g. the computational fluid dynamics simulations mentioned in the introduction.

Lastly, the results showed no decrease in performance when adding noise to the objective function, in this case, the 6D Hartmann function. When optimizing the function with low and high noise (corresponding to the measurement error range of MEMS sensors to mirror the circumstances of an applied example in fluid dynamics) the same solutions were found, in a comparable number of evaluations, as for the deterministic case. This result can be attributed to the nugget that is added to the deterministic and noisy cases, as discussed in Section~\ref{sec2.1}. These results are promising, as they show that Bayesian optimization can handle noisy objective functions just as efficiently as deterministic functions. This enables the use of Bayesian optimization for physical experiments where measurements cannot be taken without noise, but also for simulations where other noise can be introduced unknowingly. Consider, for example, a physical experiment in which the drag created by air blowing over a surface can be reduced by blowing air upwards orthogonally to the same surface using multiple actuators (for further details refer to the application Section \ref{sec4}). As each actuator has a high, if not infinite, number of settings, finding the optimal overall blowing strategy is a complex problem that is a prime candidate for the use of Bayesian optimization. However, the drag cannot be measured without noise as the measurement errors associated with the MEMS sensors introduced in Section~\ref{sec3} show. As it is impossible to collect noise-free data in such circumstances, and taking the same measurement twice would yield different results, it is critical that Bayesian optimization performs equally well on these problems.

While these findings show that the use of Bayesian optimization to optimize expensive black box functions is promising, some limitations should be noted. Firstly, the acquisition functions considered in this paper represent only a subset of those available in the literature. The general results inferred from this selection might not extend to all acquisition functions. Secondly, the dimensionality of the test functions was selected to be no greater than ten. While Bayesian optimization is generally considered to work best in this range, and even up to 20 input parameters, it would appear unlikely that these results would hold in a much higher-dimensional space \citep{moriconi}. While this is an ongoing area of research, there are multiple algorithms that attempt to make Bayesian optimization viable for higher dimensions. A popular approach is to find a low-dimensional embedding of the high-dimensional input space. Examples of this approach are REMBO \citep{wang2016bayesian}, HeSBO \citep{nayebi2019framework} and ALEBO \citep{letham2020re}. Another approach is to leverage any possible underlying additive structure of the objective function with a Bayesian optimization algorithm based on additive GPs, such as the Add-GP-UCB algorithm \citep{kandasamy2015high}. Other promising approaches are TuRBO \citep{eriksson2019scalable}, which attempts to focus on promising areas of the parameter space via local optimization, and SAASBO \citep{eriksson2021high}, which utilizes a fully Bayesian GP that aims to identify sparse axis-aligned subspaces to extend Bayesian optimization to higher dimensions. Thirdly, the conclusions drawn in this paper are based on the synthetic test functions chosen and their underlying behavior. When encountering objective functions with shapes and challenges that are not similar to those considered in this paper, the findings might not hold. Lastly, the noise added to the Hartmann function could be too low to represent every possible experiment. It is possible that the measurement error, or other sources of noise, from an experiment or simulation, are too large for Bayesian optimization to work effectively. More investigation is needed to find the maximal noise levels that Bayesian optimization can tolerate while still performing well.

\section{Application}
\label{sec4}

\begin{figure}

    \begin{minipage}[t]{0.02\textwidth}
        \vspace{-55mm}
        (A)
    \end{minipage}
    \begin{minipage}[b]{0.98\textwidth}
        \centering
        \includegraphics[width=0.8\linewidth]{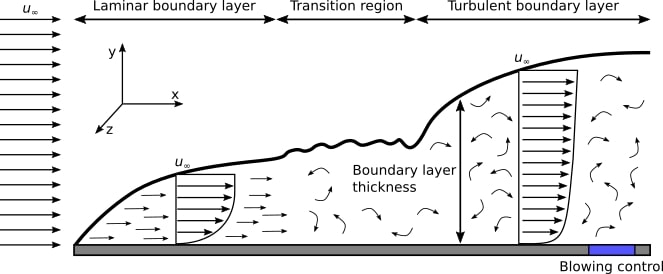}
    \end{minipage}

        \begin{minipage}[t]{0.02\textwidth}
        \vspace{-40mm}
        (B)
    \end{minipage}
    \begin{minipage}[b]{0.48\textwidth}
        \centering
        \includegraphics[width=0.9\linewidth]{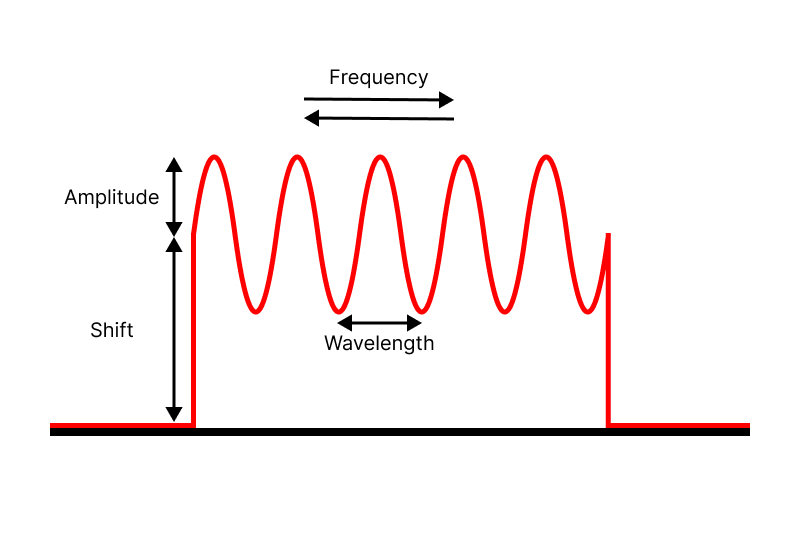}
    \end{minipage}
    \begin{minipage}[t]{0.02\textwidth}
        \vspace{-40mm}
        (C)
    \end{minipage}
    \begin{minipage}[b]{0.48\textwidth}
        \centering
        \includegraphics[width=0.9\linewidth]{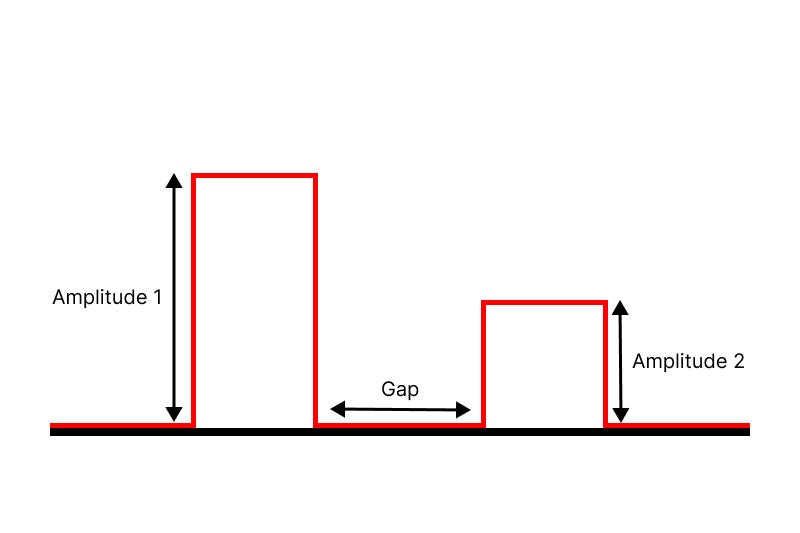}
    \end{minipage}  

    \begin{minipage}[t]{0.02\textwidth}
        \vspace{-50mm}
        (D)
    \end{minipage}
    \begin{minipage}[b]{0.48\textwidth}
        \centering
        \includegraphics[width=0.9\linewidth]{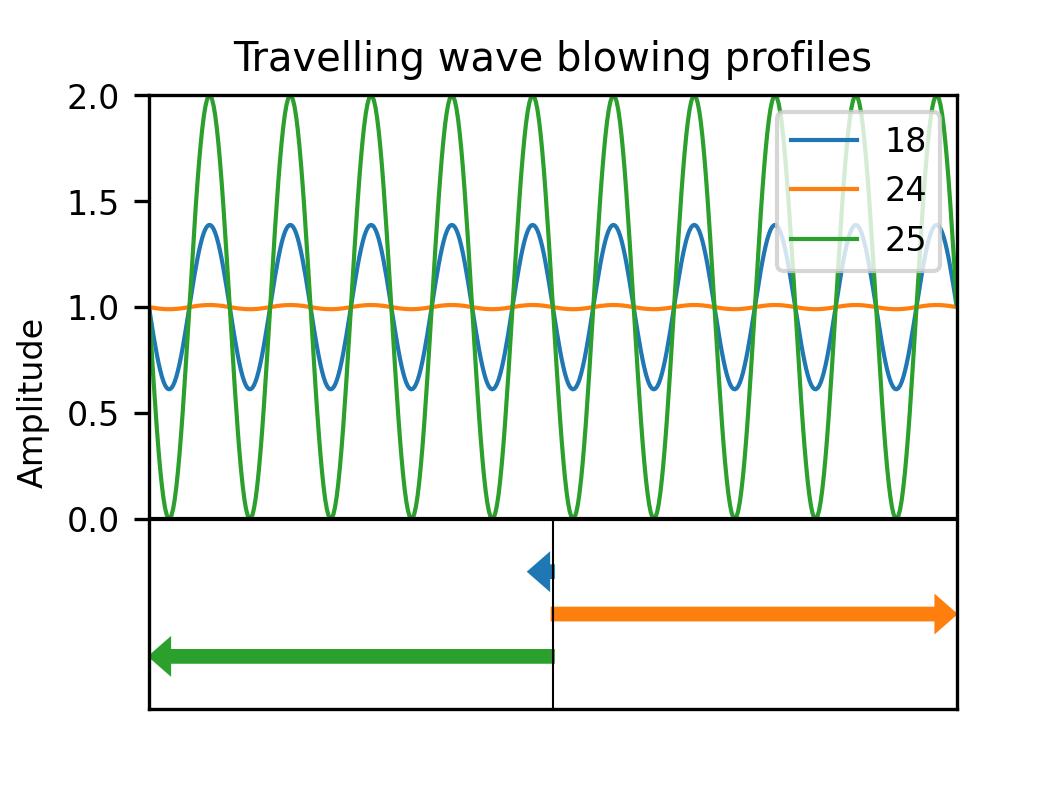}
    \end{minipage}
    \begin{minipage}[t]{0.02\textwidth}
        \vspace{-50mm}
        (E)
    \end{minipage}
    \begin{minipage}[b]{0.48\textwidth}
        \centering
        \includegraphics[width=0.9\linewidth]{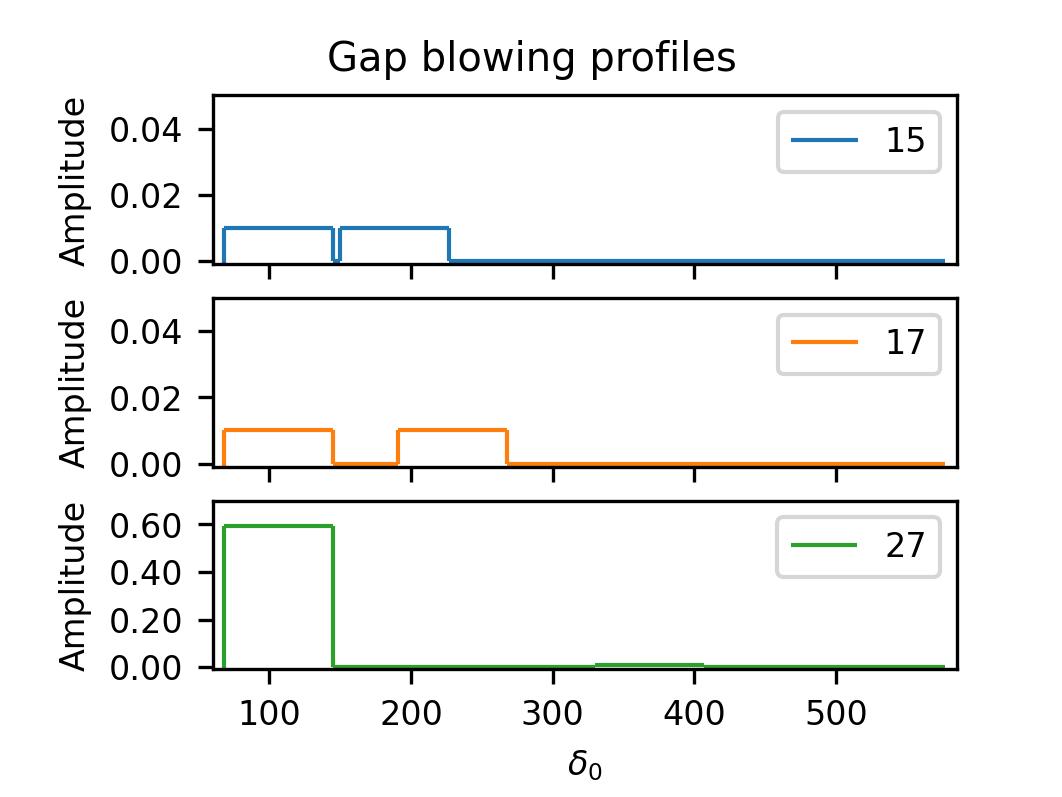}
    \end{minipage}  

    \caption{CFD simulations: (A) illustrates the flow over a flat plate. Initially the boundary layer is laminar. However, at a critical streamwise length from the leading edge, the flow transitions to a turbulent boundary layer, which is characterized by an increase in turbulent activity and skin-friction drag. The blue shaded region illustrates the location of the blowing control region in the present study; (B) shows the travelling wave blowing profile specified by an amplitude, a wavelength, a travelling frequency and a shift parameter; (C) shows the gap blowing profile specified by two blowing areas with individual amplitudes separated by a gap; (D) shows the travelling wave blowing profiles for iterations 18, 24 and 25 where the arrows on the bottom of the plots illustrate the direction and strength of travel; (E) shows the gap blowing profiles for iterations 15, 17 and 27.}
    \label{fig:Figure 7}
\end{figure}

In this section we apply the findings from the investigation on synthetic test functions in Section~\ref{sec3} to the selection and design of Bayesian optimization algorithms to specific simulations, in this case in the area of Computational Fluid Dynamics (CFD). Consider high-fidelity simulations involving the turbulent flow over a flat plate, as illustrated in Figure~\ref{fig:Figure 7}. Initially, the flow within the boundary layer is laminar. However, after a critical streamwise length, the flow transitions into a turbulent regime, characteriszed by an increase in turbulence activity and thereby an increase in skin-friction drag. This setup mimics the flow encountered on many vehicles, e.g. the flow over the wing of an aircraft, a high-speed train or the hull of a ship. The skin-friction drag is a resistive force that opposes the motion of any moving vehicle and is typically responsible for more than half of the vehicle's energy consumption. To place this into context, just a 3\% reduction in the skin-friction drag force experienced by a long-range commercial aircraft would save around £1.2M in jet fuel per aircraft per year, and prevent the annual release of 3,000 tonnes of carbon dioxide \citep{bushnell1990}. There are currently over 25,000 aircraft in active service around the world. Yet despite this significance, the complexity of turbulence has prevented the realisation of any functional and economical system to reduce the effects of turbulent skin-friction drag on any transportation vehicle. This in part, is due to our inability to find optimum solutions in parameter space to control the turbulence effectively and efficiently. The aim of these simulations is to minimize the turbulent skin-friction drag by utilizing active control, via actuators, which are seen as a key upstream technology approach for the aerospace sector, that allows us to either blow fluid away from the plate or suck fluid towards the plate. These actuators are located towards the beginning of the plate, but after the transition region, where the flow is fully turbulent. An averaged skin-friction drag coefficient is then measured along the plate from the blowing region. Within the blowing region a very large number of blowing setups are possible. For example, a simple setup could be a 1D problem where fluid is blown away from the plate uniformly over the entire blowing region, with a constant amplitude. In this case, the only parameter to optimize would be the amplitude of the blowing. However, many more complex setups are possible (e.g. see \citet{Mahfoze2019}). These numerical simulations are a prime candidate for Bayesian optimization as they fulfil the characteristics of an expensive-to-evaluate black box function: the underlying mathematical expression is unknown and each evaluation of the objective function is expensive. Indeed, one evaluation (a high-fidelity simulation with converged statistics) can take up to 12 hours on thousands of CPU cores and requires the use of specialist software, since the full turbulence activity covering the flat plate must be simulated in order to correctly resolve the quantity of interest (i.e. skin-friction drag). To perform these numerical simulations, the open-source flow solver Xcompact3D \citep{xcompact3d} is utilized on the UK supercomputing facility ARCHER2. This section optimizes two blowing profiles that follow this setup: a travelling wave with four degrees of freedom and a gap problem with three degrees of freedom, where two blowing areas with individual amplitudes are separated by a gap.

The computational setup consists of a laminar Blasius solution at the inlet, a convective condition at the outlet, a homogenous Neumann condition in the far-field, and periodic conditions in the spanwise direction. The domain dimensions are $L_x \times L_y \times L_z = 750 \delta_0 \times 80 \delta_0 \times 30 \delta_0$, where $\delta_0$ is the boundary-layer thickness at the inlet. The Reynolds number at the inlet is $Re_{\delta_0} = 1250$, based on the boundary-layer thickness (i.e. the thin layer of fluid above the plate where the flow velocity is reduced due to the presence of the plate) and free-stream velocity ($u_\infty$. i.e. the speed of the moving vehicle) at the inlet of the simulation domain. This corresponds to a momentum Reynolds number of $Re_{\theta_0} \approx 169$ to 2025, for the canonical case (no control), from the inlet to the outlet. The mesh size is chosen to be $n_x \times n_y \times n_z = 1537 \times 257 \times 128$, with a uniform spacing in the streamwise (x-axis in Figure \ref{fig:Figure 7} (A)) and spanwise (z-axis) directions and non-uniform spacing in the wall-normal direction (y-axis) to properly resolve the near-wall effects. This results in a mesh resolution of $\Delta x^+ = 31$, $0.54 \leq \Delta y^+ \leq 705$, and $\Delta z^+ = 15$ in viscous (inner) units, where the inner scaling is with respect to the friction velocity (i.e. scaled by the wall-shear stress generated by the skin-friction drag force) at the start of the control region for the canonical case. The control region extends from $x = 68 \delta_0$ to $145 \delta_0$ in the streamwise direction, corresponding to a Reynolds number range of $Re_\theta \approx 479$ to 703, for the canonical case. To accelerate the transition-to-turbulence, a random volume forcing approach \citep{Schlatter2010}, located at $x = 3.5 \delta_0$, is used to trip the boundary layer.

\begin{table}
    \centering
    \caption{Results for travelling wave experiment. Horizontal line separates initial training points and points suggested by Bayesian optimization. Points with global drag reduction over 22\% in bold.}
    \begin{tabular}{ c c c c c c c }
        Evaluation & Amplitude & Shift & Wavenumber & Frequency & GDR [in \%] \\[0.2cm]
        \hline \hline \\
        1 & 0.50 & -0.74 & 0.02 & -0.22 & -25.71 \\[0.15cm] 
        2 & 0.96 & 0.50 & 0.00 & 0.20 & 12.26 \\[0.15cm] 
        3 & 0.85 & 0.67 & 0.01 & 0.23 & 15.91 \\[0.15cm] 
        4 & 0.09 & -0.48 & 0.02 & -0.07 & -15.63 \\[0.15cm] 
        5 & 0.42 & -0.16 & 0.01 & -0.02 & -4.67 \\[0.15cm] 
        6 & 0.62 & 0.89 & 0.00 & -0.05 & 19.69 \\[0.15cm] 
        7 & 0.44 & -0.50 & 0.00 & -0.22 & -16.73 \\[0.15cm] 
        8 & 0.19 & 0.31 & 0.01 & -0.19 & 8.33 \\[0.15cm] 
        9 & 0.02 & -0.34 & 0.00 & 0.17 & -10.94 \\[0.15cm] 
        10 & 0.68 & -0.98 & 0.01 & -0.12 & -34.70 \\[0.15cm] 
        11 & 0.31 & 0.62 & 0.01 & 0.12 & 14.86 \\[0.15cm] 
        12 & 0.73 & 0.85 & 0.01 & -0.15 & 19.52 \\[0.15cm] 
        13 & 0.34 & 0.16 & 0.02 & 0.08 & 4.28 \\[0.15cm] 
        14 & 0.81 & 0.06 & 0.02 & 0.04 & 1.94 \\[0.15cm] 
        15 & 0.90 & -0.02 & 0.01 & 0.01 & -1.22 \\[0.15cm] 
        16 & 0.24 & -0.77 & 0.01 & 0.15 & -26.98 \\[0.2cm] 
        \hline \\
        17 & 0.84 & 1.00 & 0.01 & 0.04 & 21.43 \\[0.15cm] 
        \textbf{18} & \textbf{0.39} & \textbf{1.00} & \textbf{0.01} & \textbf{-0.00} & \textbf{22.07} \\[0.15cm] 
        19 & 1.00 & 1.00 & 0.00 & -0.25 & 21.77 \\[0.15cm] 
        20 & 1.00 & 1.00 & 0.02 & 0.00 & 21.83 \\[0.15cm] 
        21 & 0.48 & 1.00 & 0.02 & 0.13 & 21.90 \\[0.15cm] 
        \textbf{22} & \textbf{0.01} & \textbf{1.00} & \textbf{0.02} & \textbf{-0.25} & \textbf{22.06} \\[0.15cm] 
        23 & 0.01 & 1.00 & 0.00 & -0.25 & 21.88 \\[0.15cm] 
        \textbf{24} & \textbf{0.01} & \textbf{1.00} & \textbf{0.02} & \textbf{0.25} & \textbf{22.19} \\[0.15cm] 
        \textbf{25} & \textbf{1.00} & \textbf{1.00} & \textbf{0.02} & \textbf{-0.25} & \textbf{22.03} \\[0.15cm] 
        26 & 0.01 & 1.00 & 0.00 & 0.25 & 21.95 \\[0.15cm] 
        \textbf{27} & \textbf{0.01} & \textbf{1.00} & \textbf{0.02} & \textbf{0.01} & \textbf{22.00} \\[0.15cm] 
        \textbf{28} & \textbf{0.42} & \textbf{1.00} & \textbf{0.01} & \textbf{-0.25} & \textbf{22.07} \\[0.15cm] 
        \textbf{29} & \textbf{0.01} & \textbf{1.00} & \textbf{0.01} & \textbf{0.25} & \textbf{22.04} \\[0.15cm] 
        \textbf{30} & \textbf{0.01} & \textbf{1.00} & \textbf{0.01} & \textbf{0.01} & \textbf{22.17} \\[0.15cm] 
        31 & 0.43 & 0.79 & 0.00 & -0.22 & 18.45 \\[0.15cm] 
        32 & 1.00 & 1.00 & 0.00 & 0.25 & 21.88 \\[0.15cm] 
        33 & 0.01 & 1.00 & 0.01 & -0.25 & 21.93 \\[0.15cm] 
        34 & 0.47 & 1.00 & 0.02 & -0.25 & 21.75 \\[0.15cm] 
        35 & 1.00 & 1.00 & 0.02 & 0.25 & 21.83 \\[0.15cm] 
        36 & 0.83 & 0.94 & 0.02 & -0.02 & 21.02 \\[0.15cm] 
    \end{tabular}
    \label{table:Table 7}
\end{table}

\begin{table}
    \centering
    \caption{Results for gap experiment. Horizontal line separates initial training points and points suggested by Bayesian optimization. Points with net energy savings in bold.}
    \begin{tabular}{ c c c c c c }
        Evaluation & Amplitude 1 & Amplitude 2 & Gap & GDR [in \%] & NES [in \%] \\[0.2cm]
        \hline \hline \\
        1 & 0.97 & 0.55 & 110.35 & 32.86 & -4.31 \\[0.15cm] 
        2 & 0.28 & 0.12 & 191.95 & 10.19 & -1.31 \\[0.15cm] 
        3 & 0.87 & 0.89 & 225.60 & 34.75 & -7.41 \\[0.15cm] 
        4 & 0.40 & 0.41 & 289.40 & 17.75 & -3.36 \\[0.15cm] 
        5 & 0.13 & 0.69 & 241.42 & 16.14 & -4.34 \\[0.15cm] 
        6 & 0.71 & 0.25 & 133.84 & 22.47 & -1.66 \\[0.15cm] 
        7 & 0.66 & 0.80 & 352.52 & 26.74 & -7.97 \\[0.15cm] 
        8 & 0.53 & 0.64 & 179.61 & 25.28 & -3.18 \\[0.15cm] 
        9 & 0.83 & 0.03 & 318.82 & 19.83 & -0.97 \\[0.15cm] 
        10 & 0.19 & 0.96 & 51.96 & 25.34 & -3.53 \\[0.15cm] 
        11 & 0.04 & 0.30 & 92.16 & 8.72 & -0.96 \\[0.15cm] 
        12 & 0.49 & 0.45 & 31.67 & 22.86 & -0.87 \\[0.2cm]
        \hline \\
        13 & 0.27 & 0.01 & 5.00 & 7.72 & -0.28 \\[0.15cm] 
        14 & 1.00 & 0.01 & 5.00 & 22.04 & -2.84 \\[0.15cm] 
        \textbf{15} & \textbf{0.01} & \textbf{0.01} & \textbf{5.00} & \textbf{0.91} & \textbf{0.25} \\[0.15cm] 
        16 & 0.15 & 0.38 & 5.00 & 13.91 & -0.77 \\[0.15cm] 
        \textbf{17} & \textbf{0.01} & \textbf{0.01} & \textbf{45.54} & \textbf{0.98} & \textbf{0.32} \\[0.15cm] 
        \textbf{18} & \textbf{0.01} & \textbf{0.01} & \textbf{355.00} & \textbf{0.72} & \textbf{0.06} \\[0.15cm] 
        19 & 0.01 & 0.10 & 5.00 & 3.01 & -0.44 \\[0.15cm] 
        20 & 0.47 & 0.01 & 355.00 & 11.84 & -0.39 \\[0.15cm] 
        \textbf{21} & \textbf{0.01} & \textbf{0.01} & \textbf{131.32} & \textbf{0.77} & \textbf{0.11} \\[0.15cm] 
        22 & 0.01 & 0.01 & 237.00 & 0.62 & -0.04 \\[0.15cm] 
        23 & 0.01 & 0.63 & 5.00 & 15.33 & -0.15 \\[0.15cm] 
        \textbf{24} & \textbf{0.01} & \textbf{0.01} & \textbf{77.50} & \textbf{0.80} & \textbf{0.13} \\[0.15cm] 
        \textbf{25} & \textbf{0.01} & \textbf{0.01} & \textbf{5.00} & \textbf{0.91} & \textbf{0.25} \\[0.15cm] 
        26 & 0.28 & 0.63 & 5.00 & 21.72 & -1.24 \\[0.15cm] 
        27 & 0.60 & 0.01 & 184.71 & 14.83 & -0.01 \\[0.15cm] 
        28 & 0.38 & 0.67 & 5.00 & 24.86 & -1.31 \\[0.15cm] 
    \end{tabular}
    \label{table:Table 8}
\end{table}

Figure \ref{fig:Figure 7} depicts the two blowing profiles in question, and defines the parameters for the optimization. The travelling wave with four degrees of freedom, given in Subfigure \ref{fig:Figure 7} (B), is a wave defined by an amplitude in the range 0.01 to 1.00\% of the overall free-stream velocity and a wavelength between 0.00 and 0.02 (inner scaling). The angular frequency, restricted to values between -0.25 and 0.25 (inner scaling), allows the wave to travel up- and down-stream. Lastly, a shift parameter displaces the wave vertically up and down. This parameter is restricted to values between -1.00 and 1.00\% of the free-stream velocity. The blowing turns into suction for cases where the blowing profile is negative. The gap configuration with three degrees of freedom, illustrated in Subfigure \ref{fig:Figure 7} (C), includes two blowing regions with individual amplitudes in the range 0.01 to 1.00\% of the overall free-stream velocity and a gap restricted to between $5 \delta_0$ and $355 \delta_0$. While the aim for both problems is the maximisation of the global drag reduction (GDR), defined as the globally averaged skin-friction drag reduction with respect to the canonical case, the gap problem with three degrees of freedom also considers the energy consumption of the actuators and tries to find profiles that reduce the drag while also achieving net energy savings (NES). NES is achieved when the energy used by the blowing device is smaller than the energy saved by the drag reduction (much of the energy expenditure in aerodynamics/hydrodynamics applications is used to overcome the skin-friction drag). In this work, NES is calculated following the approach of \citet{Mahfoze2019}, where the input power for the blowing device is estimated from real-world experimental data and a relationship between input power and blowing velocity is derived (see \citet{Mahfoze2019} for more details). 

The Bayesian optimization algorithm used for both problems is informed by the results of the synthetic test functions from Section~\ref{sec3}. For the surrogate model a Gaussian process with a zero mean function and a Mat\'ern kernel with $\nu = 5/2$ was defined and its hyperparameters were estimated from the training data using maximum likelihood estimation. While CFD simulations are expensive, they allow for parallel evaluations. This is done by running, concurrently, multiple simulations and combining the results once all simulations are completed. Therefore, these setups are well suited for the multi-point approach presented in Section~\ref{sec3.2.4} that, as our investigations showed, has no real disadvantage compared to the single-point approach. Based on previous work, for example \citet{Mahfoze2019}, the possibility that the underlying objective function is characterized by large flat areas similar to the 6D Ackley function cannot be ruled out. Thus, an acquisition function that allows the selection of batches and performs well even when encountering flat areas should be implemented. The sequential Monte Carlo Upper Confidence Bound acquisition function with the trade-off hyperparameter $\beta$=1 yielded very good results for the Ackley function as well as all other test function and is thus chosen with a batch size of four to optimize the CFD problems in this section. Section~\ref{sec3.2.2} showed that Bayesian optimization is able to find good solutions even with a relatively small number of initial training data points. Hence, for both problems, four points per input dimension were randomly selected from a Maximin Latin Hypercube \citep{Husslage2011}.

Table \ref{table:Table 7} present the results of the 16 training points plus 20 points (or five batches of four points), selected using Bayesian optimization, by maximizing the GDR, for the travelling wave problem. While the highest GDR of the initial training data was 19.69\%, Bayesian optimization improves upon this value with each evaluated batch, finding multiple strategies that achieve a GDR above 22\%, with the best strategy, from batch 2, giving a GDR of 22.19\%. Three blowing profiles, including the overall best solution found, are depicted in Figure \ref{fig:Figure 7} (D). Overall, the shift parameter seems to be the main driver behind the drag reduction, as almost all strategies selected by Bayesian optimization implement the upper limit of this parameter, independent of the other parameter values. 

While blowing at a high amplitude yields increased skin-friction drag reduction, it also consumes more energy. The second experiment addresses this point, and accounts for the energy consumption, following \citet{Mahfoze2019}, when optimising the blowing profile. Table \ref{table:Table 8} provides the results for 12 initial training points and four batches, suggested via Bayesian optimization, by maximizing the NES. The initial strategies selected by the Latin Hypercube did not find a solution that achieved both GDR and NES. Bayesian optimization was able to find multiple strategies that achieved both, in which the NES and the GDR were relatively small (0.11 to 0.32\% and 0.77 to 0.98\%). However, the algorithm also found one strategy with NES of -0.01\% and a GDR of 14.83\%. While this strategy did not achieve NES, it did not increase overall energy use, and a small increase in the efficiency of the actuators could yield NES with a considerable GDR. Compared to the high intensity blowing for the travelling wave, in this experiment the amplitudes are clustered towards the lower end of the parameter space. This is a result of the objective to optimize NES, which penalizes high-velocity blowing due to its increased power requirements.  Figure \ref{fig:Figure 7} (E) illustrates this by providing the blowing profiles of three solutions; the two solutions with the greatest net energy savings and the solution with a high GDR, as previously described.

\section{Conclusion}
\label{sec5}

In this paper, Bayesian optimization algorithms, implemented with different types of acquisition functions, were benchmarked regarding their performance and their robustness on synthetic test functions inspired by applications in engineering and machine learning. Synthetic test functions have the advantage that their shape and their global optimum are known. This allows the algorithms to be evaluated on (a) how close their best solutions are to the global optimum and (b) how well they perform on specific challenges such as oscillating functions or functions with steep edges. This evaluation can indicate advantages and shortcomings of the individual acquisition functions and can inform researchers on the best approach to choose for their specific problem.

Four sets of comparisons were conducted in this research article. First, analytical single-point acquisition functions were compared to each other. Second, the effect of varying the number of initial training data points was investigated. Third, the analytical approach was contrasted with acquisition functions based on Monte Carlo sampling. Fourth, the single-point approach was compared to the multi-point or batched approach.

From these experiments six main conclusions could be drawn: (i) While all acquisition functions performed well on simple test functions, optimistic policies, such as the Upper Confidence Bound, dealt best with challenging problems. (ii) Varying the number of initial training data points did not have a significant effect on the performance of the individual methods. (iii) Improvement-based acquisition functions struggled with flat test functions. (iv) The results showed that Monte Carlo acquisition functions and multi-point acquisition functions present a good alternative to the widely used analytical single-point methods. (v) The multi-point approach is particularly advantageous when the objective function takes a long time to evaluate and allows parallel evaluations. (vi) Bayesian optimization performs equally well on noisy objective functions.

Finally, two experiments in computational fluid dynamics were taken as illustrative examples on how the findings of this paper can be used to guide the design of a Bayesian optimization algorithm and tailor it to unique problems. In detail, a multi-point approach was employed that used the Monte Carlo Upper Confidence Bound acquisition function allowing multiple points to be evaluated in parallel with concurrent simulations. For both experiments, Bayesian optimization was able to improve upon the training points straight away and find solutions that implement global drag reduction for the travelling wave and global drag reduction as well as net energy savings for the experiment where two blowing areas were separated by a gap. However, the effects of the second experiment remained relatively small but these optimization studies are nevertheless encouraging in the quest to design a robust and efficient control strategy to reduce drag around moving vehicles.

\newpage

\section*{Abbreviations}
\begin{table} [!h]
    \centering
    \begin{tabular}{ m{3cm} l }
    AUC & Area under the curve \\
    BO & Bayesian optimization \\
    BUCB & Batched Upper Confidence Bound \\
    CDF & Cumulative distribution function \\
    CFD & Computational fluid dynamics \\
    Cl & Constant liar \\
    EI & Expected Improvement \\
    ES & Entropy Search \\
    GDR & Global drag reduction \\
    GP & Gaussian process \\
    GP-UCB & Gaussian Process Upper Confidence Bound \\
    MC & Monte Carlo \\
    MES & Max-value entropy search \\
    MEMS & Micro-Electro-mechanical-Systems \\
    MLE & Maximum likelihood estimation \\
    NES & Net-energy savings \\
    NUBO & Newcastle University Bayesian Optimization \\
    PDF & Probability density function \\
    PES & Predictive Entropy Search \\
    PI & Probability of Improvement \\
    UCB & Upper Confidence Bound \\
    \end{tabular}
\end{table}

\section*{Acknowledgments}
The work has been supported by the Engineering and Physical Sciences Research Council (EPSRC) under grant numbers EP/T020946/1, EP/T021144/1 and the EPSRC Centre for Doctoral Training in Cloud Computing for Big Data under grant number EP/L015358/1. The authors would also like to thank the UK Turbulence Consortium (EP/R029326/1) for the computational time made available on the \href{https://www.archer2.ac.uk}{ARCHER2 UK National Supercomputing Service}.

\section*{Data Availability Statement}
The datasets generated and analyzed for this study can be found in the GitHub repository \href{https://github.com/mikediessner/investigating-BO}{https://github.com/mikediessner/investigating-BO}.

\clearpage

\bibliography{references}  

\clearpage 
\appendix

\section*{Appendix}

\section{Overview of Acquisition Functions}

\begin{table}[h]
    \centering
    
    \caption{Overview of all acquisition functions used in the comparisons. For each function, the group the function belongs to, and whether an analytical or a Monte Carlo implementation is used in this article, is given.}
    \begin{tabular}{ l | c c c }
        Acquisition function & Group & Analytical & Monte Carlo  \\[0.2cm]
        \hline \hline \\
        Probability of Improvement & Improvement-based & Yes & Yes \\[0.15cm]
        Expected Improvement & Improvement-based  & Yes & Yes \\[0.15cm]
        Upper Confidence Bound  & Optimistic & Yes & Yes \\[0.15cm]
        Hedge & Portfolio & Yes & No \\[0.15cm]
        Max-value Entropy Search & Entropy-based & No & Yes \\[0.15cm]
        Constant Liar & Improvement-based & Yes & No \\[0.15cm]
        Batched Upper Confidence Bound & Optimistic & Yes & No \\[0.15cm]
    \end{tabular}
\end{table}

\section{Pseudocode}

\begin{algorithm}

    \caption{Fit a GP to training data. Estimate covariance kernel hyperparameters and nugget.}
    \begin{algorithmic}
    \setstretch{1.2}
        \Require Training data $\mathcal{D}$ with training inputs $\textbf{X}$ and training outputs $\textbf{y}$; GP prior mean function $\mu_0$; GP prior covariance kernel $k$
        \State Estimate parameters by maximizing $\log p(\textbf{y} | \textbf{X}) = - \frac{1}{2} \textbf{y}^T \textbf{K}^{-1} \textbf{y} - \frac{1}{2} \log |\textbf{K}| - \frac{n}{2} \log(2 \pi)$ 
        \State where $\textbf{K} = k(\textbf{x}, \textbf{x}) + \nu I$ 
    \end{algorithmic}
\end{algorithm}

\vspace{20mm}

\begin{algorithm}

    \caption{Compute posterior mean function and variance.}
    \begin{algorithmic}
    \setstretch{1.2}
        \Require Training data $\mathcal{D}$ with training inputs $\textbf{X}$ and training outputs $\textbf{y}$; GP prior mean function $\mu_0$; GP prior covariance kernel $k$
        \State Use Algorithm 1 to estimate hyperparameters of the GP from the training data $\mathcal{D}$
        \State Compute posterior mean function $\mu_n(\textbf{x}) = \mu_0(\textbf{x}) + \textbf{k}(\textbf{x})^T (\textbf{K} + \sigma^2\textbf{I})^{-1} (\textbf{y} - \textbf{m})$
        \State Compute posterior variance $\sigma^2_n(\textbf{x}) = k(\textbf{x}, \textbf{x}) - \textbf{k}(\textbf{x})^T(\textbf{K} + \sigma^2\textbf{I})^{-1} \textbf{k}(\textbf{x})$
    \end{algorithmic}
\end{algorithm}

\begin{algorithm}

    \caption{The standard Bayesian optimisation algorithm.}
    \begin{algorithmic}
    \setstretch{1.2}
        \Require Input space $\mathcal{X}$; Objective function $f$; Training data $\mathcal{D}$ with training inputs $\textbf{X}$ and training outputs $\textbf{y}$; GP prior mean function $\mu_0$; GP prior covariance kernel $k$; Acquisition function $\alpha$; Iterations $N$
        \For{$n \leq N$}
            \State Use Algorithm 2 to update the mean function $\mu_n$ and the variance $\sigma_n$
            \State Find $\textbf{x}_n = \underset{\textbf{x} \in \mathcal{X}}{argmax} \ \ \alpha(\textbf{x}; \mathcal{D}_{n-1})$
            \State Sample $y_n = f(\textbf{x}_n)$
            \State Add $\textbf{x}_n$ and $y_n$ to $\textbf{X}$ and $\textbf{y}$
        \EndFor
    \end{algorithmic}
\end{algorithm}

\begin{algorithm}

    \caption{The Hedge portfolio Bayesian optimisation algorithm.}
    \begin{algorithmic}
    \setstretch{1.2}
        \Require Input space $\mathcal{X}$; Objective function $f$; Training data $\mathcal{D}$ with training inputs $\textbf{X}$ and training outputs $\textbf{y}$; GP prior mean function $\mu_0$; GP prior covariance kernel $k$; Acquisition functions $A$; Iterations $N$; Hedge parameter $\eta$
        \State Initialise gains $g_0^{\alpha} = 0$ for all $\alpha$ in $A$
        \State Use Algorithm 2 to update the mean function $\mu_n$ and the variance $\sigma_n$
        \For{$n \leq N$}
            \State Find $\textbf{x}^{\alpha}_n = \underset{\textbf{x} \in \mathcal{X}}{argmax} \ \ \alpha(\textbf{x}; \mathcal{D}_{n-1})$ for all $\alpha$ in $A$
            \State Select point $\textbf{x}_n = \textbf{x}^{\alpha}_n$ with probability $p_n(\alpha) = \exp\left(\eta g^{\alpha}_{n-1}\right)$ / $\sum^k_{i=1} \exp\left(\eta g^i_{n-1}\right)$
            \State Sample $y_n = f(\textbf{x}_n)$
            \State Add $\textbf{x}_n$ and $y_n$ to $\textbf{X}$ and $\textbf{y}$
            \State Use Algorithm 1 to update the mean function $\mu_{n+1}$ and the covariance kernel $K_{n+1}$
            \State Get rewards $r_n^{\alpha} = \mu_{n+1}(\textbf{x}_n^{\alpha})$
            \State Get gains $g_n^{\alpha} = g_{n-1}^{\alpha} + r_n^{\alpha}$
        \EndFor
    \end{algorithmic}
\end{algorithm}

\begin{algorithm}

    \caption{The Max-value Entropy Search Bayesian optimisation algorithm.}
    \begin{algorithmic}
    \setstretch{1.2}
        \Require Input space $\mathcal{X}$; Objective function $f$; Training data $\mathcal{D}$ with training inputs $\textbf{X}$ and training outputs $\textbf{y}$; GP prior mean function $\mu_0$; GP prior covariance kernel $k$; Acquisition function $\alpha$; Iterations $N$; Samples $M$
        \For{$n \leq N$}
            \State Use Algorithm 2 to update the mean function $\mu_n$ and the variance $\sigma_n$
            \State Draw $M$ samples form $\textbf{m} \sim \text{Unif}([0, 1])$
            \State Compute $\alpha_{n-1}(\cdot) = a - b \log(-\log(\textbf{m}))$ \Comment{approx. with Gumbel distribution $\mathcal{G}(a, b)$}
            
            \State Find $\textbf{x}_n = \underset{\textbf{x} \in \mathcal{X}}{argmax} \ \ \alpha_{n-1}(\textbf{x})$
            \State Sample $y_n = f(\textbf{x}_n)$
            \State Add $\textbf{x}_n$ and $y_n$ to $\textbf{X}$ and $\textbf{y}$
        \EndFor
    \end{algorithmic}
\end{algorithm}

\begin{algorithm}

    \caption{The Monte Carlo Bayesian optimisation algorithm.}
    \begin{algorithmic}
    \setstretch{1.2}
        \Require Input space $\mathcal{X}$; Objective function $f$; Training data $\mathcal{D}$ with training inputs $\textbf{X}$ and training outputs $\textbf{y}$; GP prior mean function $\mu_0$; GP prior covariance kernel $k$; Acquisition function $\alpha$; Iterations $N$; Monte Carlo samples $M$
        \For{$n \leq N$}
            \State Use Algorithm 2 to update the mean function $\mu_n$ and the variance $\sigma_n$
            \State Draw $M$ samples $\textbf{z} \sim \mathcal{N}(0, \textbf{I})$
            \State Find $\textbf{x}_n = \underset{\textbf{x} \in \mathcal{X}}{argmax} \ \ \frac{1}{M} \ \alpha_{MC}(\textbf{x}; \textbf{z}, \mathcal{D}_{n-1})$
            \State Sample $y_n = f(\textbf{x}_n)$
            \State Add $\textbf{x}_n$ and $y_n$ to $\textbf{X}$ and $\textbf{y}$
        \EndFor
    \end{algorithmic}
\end{algorithm}

\begin{algorithm}

    \caption{The Constant Liar Bayesian optimisation algorithm.}
    \begin{algorithmic}
    \setstretch{1.2}
        \Require Input space $\mathcal{X}$; Objective function $f$; Training data $\mathcal{D}$ with training inputs $\textbf{X}$ and training outputs $\textbf{y}$; GP prior mean function $\mu_0$; GP prior covariance kernel $k$; Acquisition function $\alpha$; Iterations $N$; Batch size $B$
        \For{$n \leq N$}
            \For{b $\leq$ B} \Comment{Compute batch points}
                \State Use Algorithm 2 to update the mean function $\mu_n$ and the variance $\sigma_n$
                \State Find $\textbf{x}_b = \underset{\textbf{x} \in \mathcal{X}}{argmax} \ \ \alpha_{EI}(\textbf{x}; \mathcal{D}_{n-1})$
                \State Compute lie $y^{L}_b = \max{\textbf{y}}$ or $y^{L}_b = \min{\textbf{y}}$
                \State Add $\textbf{x}_b$ and $y^{L}_b$ to $\textbf{X}$ and $\textbf{y}$
            \EndFor
            \For{$b \leq B$} \Comment{Evaluate batch points}
                \State Sample $y_b = f(\textbf{x}_b)$
                \State Replace all $y^{L}_b$ in $\textbf{y}$ with $y_b$ 
            \EndFor
        \EndFor
    \end{algorithmic}
\end{algorithm}

\begin{algorithm}

    \caption{The Batched Upper Confidence Bound Bayesian optimisation algorithm.}
    \begin{algorithmic}
    \setstretch{1.2}
        \Require Input space $\mathcal{X}$; Objective function $f$; Training data $\mathcal{D}$ with training inputs $\textbf{X}$ and training outputs $\textbf{y}$; GP prior mean function $\mu_0$; GP prior covariance kernel $k$; Acquisition function $\alpha$; Iterations $N$; Batch size $B$
        \For{$n \leq N$}
            \State Use Algorithm 2 to update the mean function $\mu_n$ and the variance $\sigma_n$
            \For{$b \leq B$} \Comment{Compute batch points}
                \State Find $\textbf{x}_b = \underset{\textbf{x} \in \mathcal{X}}{argmax} \ \ \alpha_{UCB}(\textbf{x}; \mathcal{D}_{n-1})$
                \State Use batch points $\textbf{x}_b$ to update the posterior variance $\sigma_n$ as shown in Algorithm 2
            \EndFor
            \For{$b \leq B$} \Comment{Evaluate batch points}
                \State Sample $y_b = f(\textbf{x}_b)$
                \State Add $\textbf{x}_b$ and $y_b$ to $\textbf{X}$ and $\textbf{y}$
            \EndFor            
        \EndFor
    \end{algorithmic}
\end{algorithm}

\FloatBarrier

\section{Mathematical Definition of Synthetic Test Functions}

\subsection{6D Ackley}
Function: $f(\textbf{x}) = -a \exp \left(-b \sqrt{\frac{1}{6} \sum^6_{i=1} x^2_i}\right) - \exp \left( \frac{1}{6} \sum^6_{i=1} \cos(c x_i)\right) + a + \exp(1)$

\hspace{5mm} with $a=20.0$, $b=0.5$ and $c=0.0$

Bounds: $x_i \in [-32.768, 32.768] \quad \forall \ i = 1, 2, ..., 6$

Global Minimum: $f(\textbf{x}^*) = 0 \quad $ at $\quad \textbf{x}^* = (0, 0, 0, 0, 0, 0)$

\vspace{5mm}

\subsection{10D Dixon-Price}
Function: $f(\textbf{x}) = (x_1 - 1)^2 + \sum^{10}_{i=2} i (2 x_i^2 - x_{i-1})^2$

Bounds: $x_i \in [-10, 10] \quad \forall \ i = 1, 2, ..., 10$

Global Minimum: $f(\textbf{x}^*) = 0 \quad $ at $\quad x_i^* = 2^{-\frac{2^i-2}{2^i}} \quad \forall \ i = 1, ..., 10$

\vspace{5mm}

\subsection{8D Griewank}
Function: $f(\textbf{x}) = \sum^{8}_{i=1} \frac{x_i^2}{4000} - \prod^8_{i=1} \cos \left(\frac{x_i}{\sqrt{i}} \right) + 1$

Bounds: $x_i \in [-600, 600] \quad \forall i = 1, 2, ..., 8$

Global Minimum: $f(\textbf{x}^*) = 0 \quad $ at $\quad \textbf{x}^* = (0, 0, 0, 0, 0, 0, 0, 0)$

\vspace{5mm}

\subsection{6D Hartmann}
Function: $f(\textbf{x}) = - \sum^4_{i=1} \alpha_i \exp \left(- \sum^6_{j=1} \textbf{A}_{ij} \left(x_j - \textbf{P}_{ij}\right)^2 \right)$

\hspace{20mm} where $\alpha = (1.0, 1.2, 3.0, 3.2)^T$

\hspace{20mm} and $\textbf{A} = \begin{pmatrix}
                                    10 & 3 & 17 & 3.5 & 1.7 & 8\\
                                    0.05 & 10 & 17 & 0.1 & 8 & 14\\
                                    3 & 3.5 & 1.7 & 10 & 17 & 8\\
                                    17 & 8 & 0.05 & 10 & 0.1 & 14
                                \end{pmatrix}$

\hspace{20mm} and $\textbf{P} = 10^{-4}     \begin{pmatrix}
                                                1312 & 1696 & 5569 & 124 & 8283 & 5886\\
                                                2329 & 4135 & 8307 & 3736 & 1004 & 9991\\
                                                2348 & 1451 & 3522 & 2883 & 3047 & 6650\\
                                                4047 & 8828 & 8732 & 5743 & 1091 & 381
                                            \end{pmatrix}$

Bounds: $x_i \in (0, 1) \quad \forall i = 1, 2, ..., 6$

Global Minimum: $f(\textbf{x}^*) = -3.32237 \quad $ at $\quad \textbf{x}^* = (0.20169, 0.150011, 0.476874, 0.275332, 0.311652, 0.6573)$

\vspace{5mm}

\subsection{5D Michalewicz}
Function: $f(\textbf{x}) = - \sum^{5}_{i=1} \sin (x_i) \sin^{20} \left( \frac{i x_i^2}{\pi}\right)$

Bounds: $x_i \in [0, \pi] \quad \forall i = 1, 2, ..., 5$

Global Minimum: $f(\textbf{x}^*) = -4.687658 \quad $

\vspace{5mm}

\subsection{10D Sphere}
Function: $f(\textbf{x}) = \sum^{10}_{i=1} x_i^2$

Bounds: $x_i \in [-5.12, 5.12] \quad \forall i = 1, 2, ..., 10$

Global Minimum: $f(\textbf{x}^*) = 0 \quad $ at $\quad \textbf{x}^* = (0, 0, 0, 0, 0, 0, 0, 0, 0, 0)$

\vspace{10mm}

\section{Supplementary Statistical Analysis}

In this section we provide a formal statistical comparison of the algorithms of the analytical single-point acquisition functions and the multi-point acquisition functions both with five initial starting points per input dimension - we see these as the two most important comparisons in the paper. The other investigations either represent a similar setup (analytical single-point comparison with different numbers of initial training points as presented in Section 3.2.2 of the main article) or are used to facilitate the comparison of the multi-point acquisition functions (Monte Carlo comparison in Section 3.2.3 of the main article).

\begin{figure}
    \begin{minipage}[t]{0.02\textwidth}
        \vspace{-45mm}
        (A)
    \end{minipage}
    \begin{minipage}[t]{0.48\textwidth}
        \centering
        \includegraphics[width=0.8\linewidth]{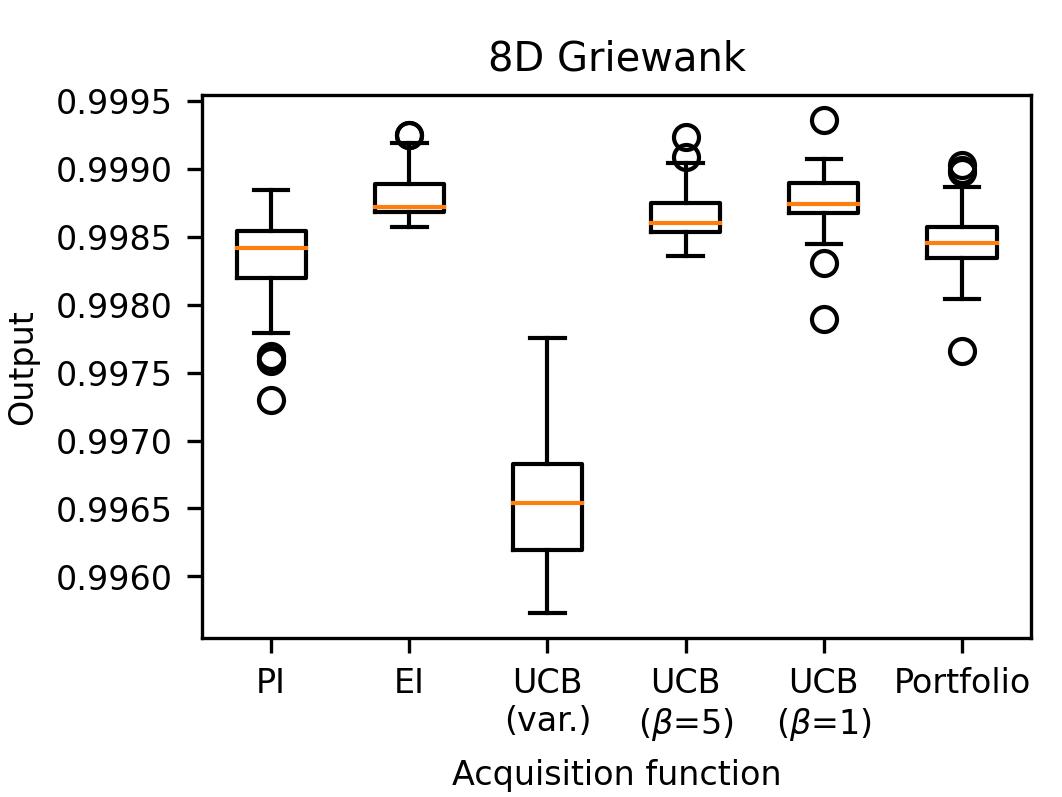}
    \end{minipage}
    \begin{minipage}[t]{0.02\textwidth}
        \vspace{-45mm}
        (B)
    \end{minipage}
    \begin{minipage}[t]{0.48\textwidth}
    \centering
        \includegraphics[width=0.8\linewidth]{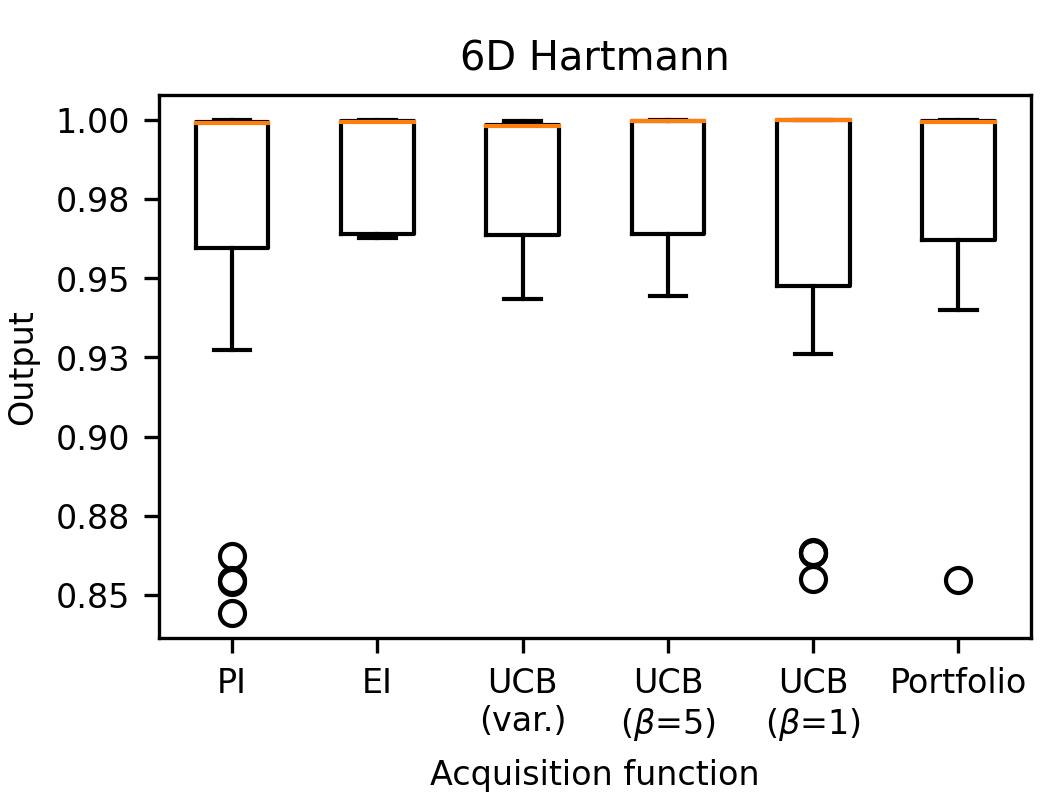}
    \end{minipage}
    \begin{minipage}[t]{0.02\textwidth}
        \vspace{-45mm}
        (C)
    \end{minipage}
    \begin{minipage}[t]{0.48\textwidth}
        \centering
        \includegraphics[width=0.8\linewidth]{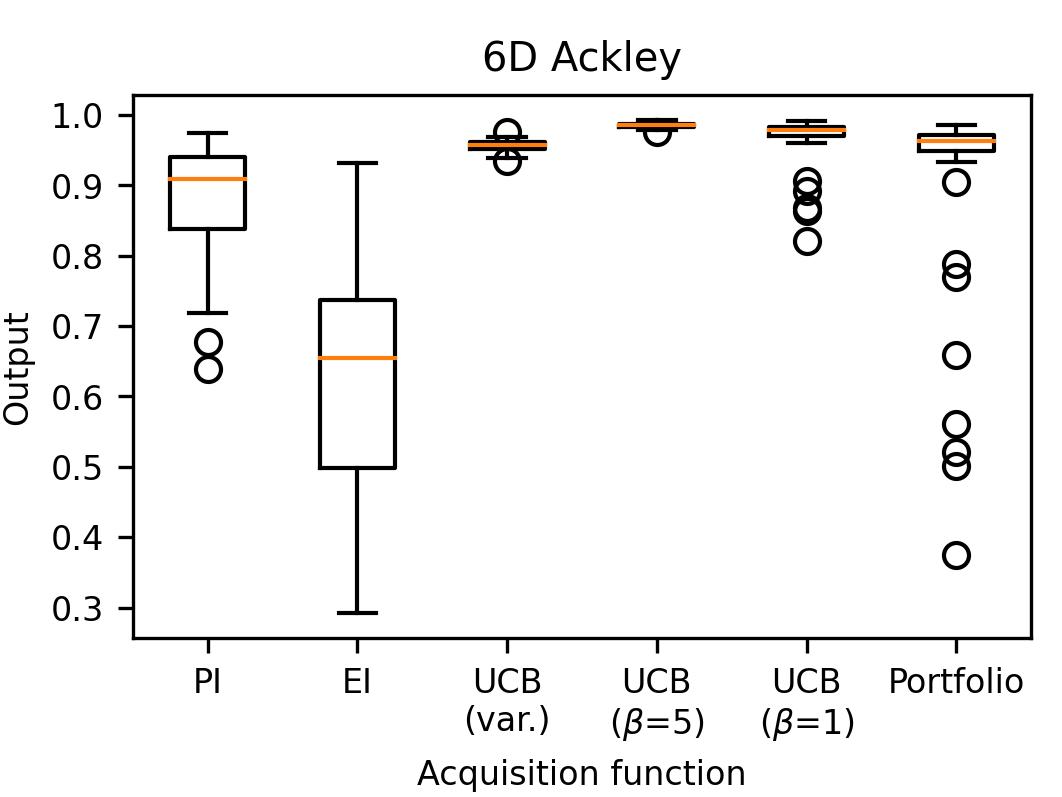}
    \end{minipage}
    \begin{minipage}[t]{0.02\textwidth}
        \vspace{-45mm}
        (D)
    \end{minipage}
    \begin{minipage}[t]{0.48\textwidth}
    \centering
        \includegraphics[width=0.8\linewidth]{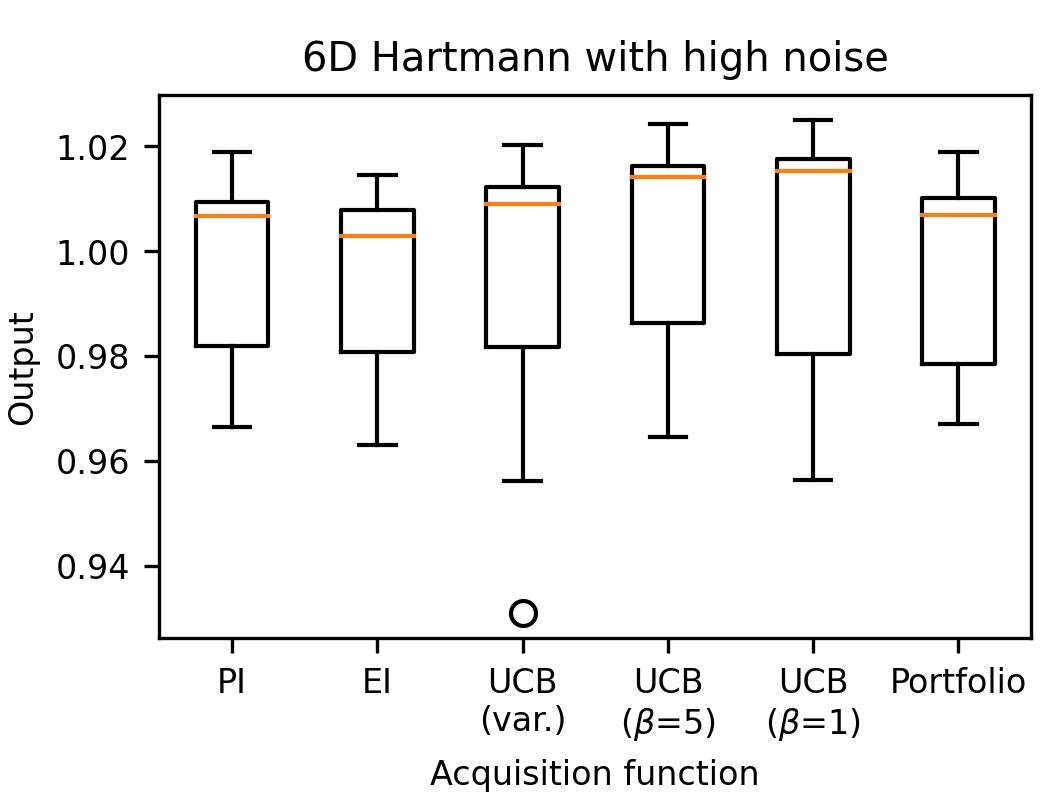}
    \end{minipage}

    \caption{Box plots of best values found for analytical single-point acquisition functions with five initial starting points per input dimension. The orange line indicates the median, the box extends from the lower to the upper quartile range and the whiskers indicate the range of the best values without the outliers shown as circles.}
    \label{fig:SFigure 1}
\end{figure}

\begin{figure}
    \begin{minipage}[t]{0.02\textwidth}
        \vspace{-45mm}
        (A)
    \end{minipage}
    \begin{minipage}[t]{0.48\textwidth}
        \centering
        \includegraphics[width=0.8\linewidth]{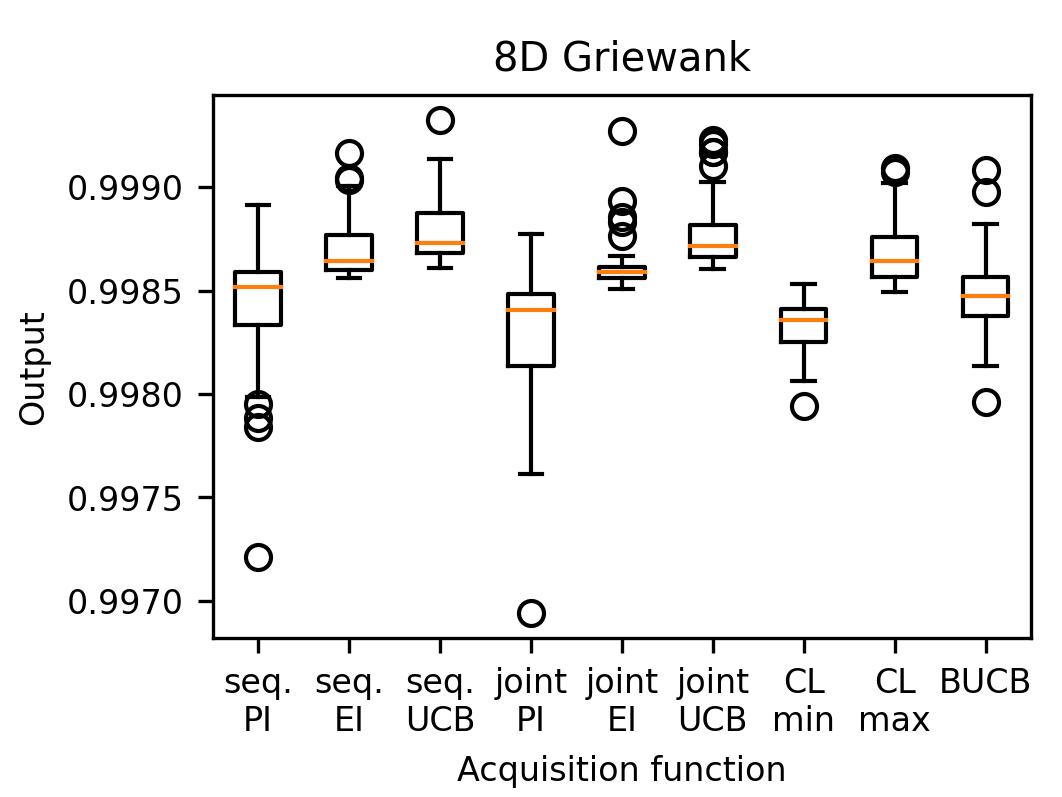}
    \end{minipage}
    \begin{minipage}[t]{0.02\textwidth}
        \vspace{-45mm}
        (B)
    \end{minipage}
    \begin{minipage}[t]{0.48\textwidth}
    \centering
        \includegraphics[width=0.8\linewidth]{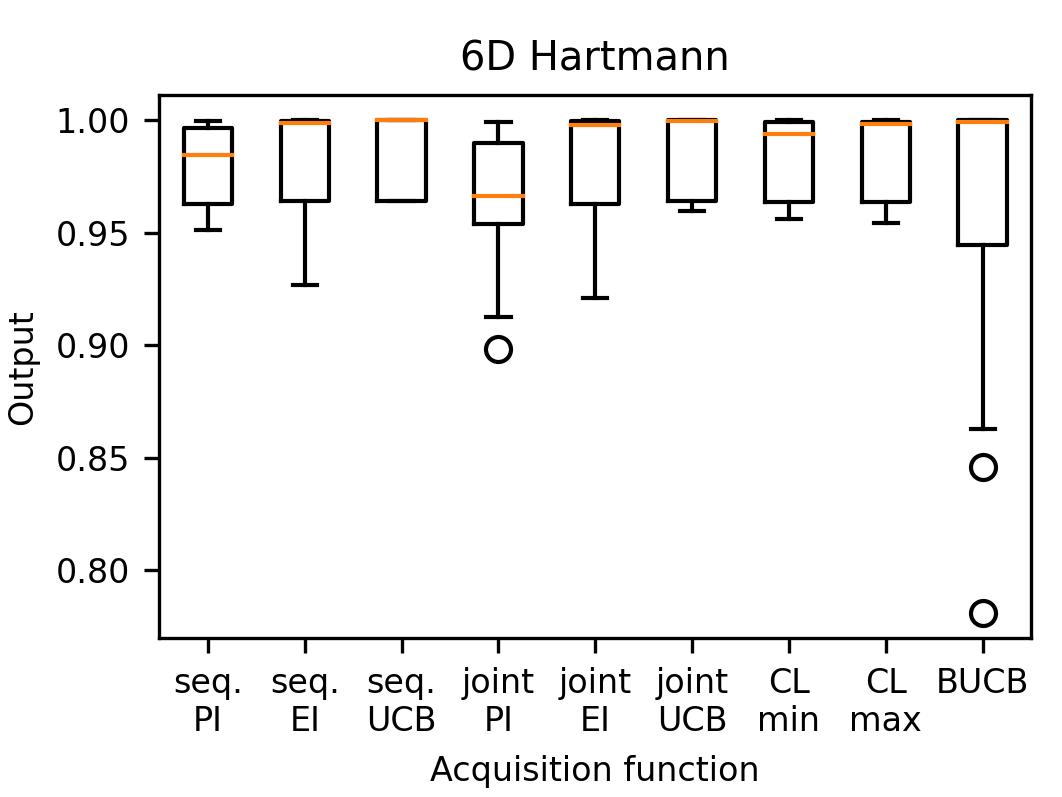}
    \end{minipage}
    \begin{minipage}[t]{0.02\textwidth}
        \vspace{-45mm}
        (C)
    \end{minipage}
    \begin{minipage}[t]{0.48\textwidth}
        \centering
        \includegraphics[width=0.8\linewidth]{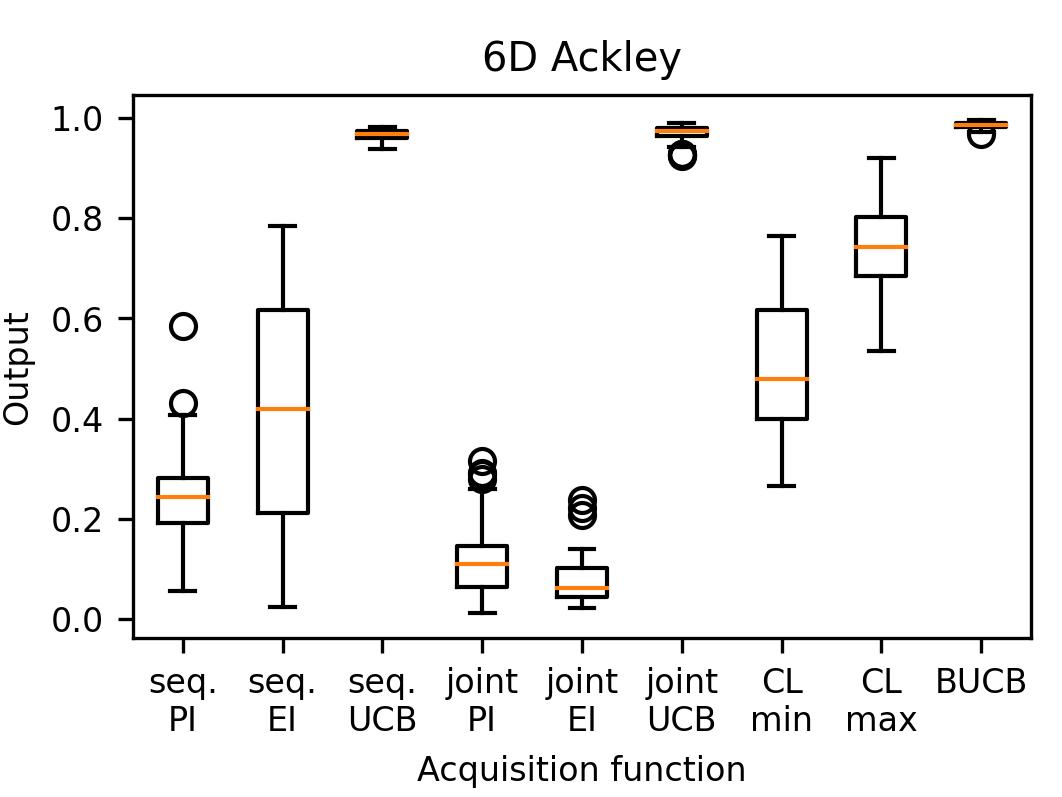}
    \end{minipage}
    \begin{minipage}[t]{0.02\textwidth}
        \vspace{-45mm}
        (D)
    \end{minipage}
    \begin{minipage}[t]{0.48\textwidth}
    \centering
        \includegraphics[width=0.8\linewidth]{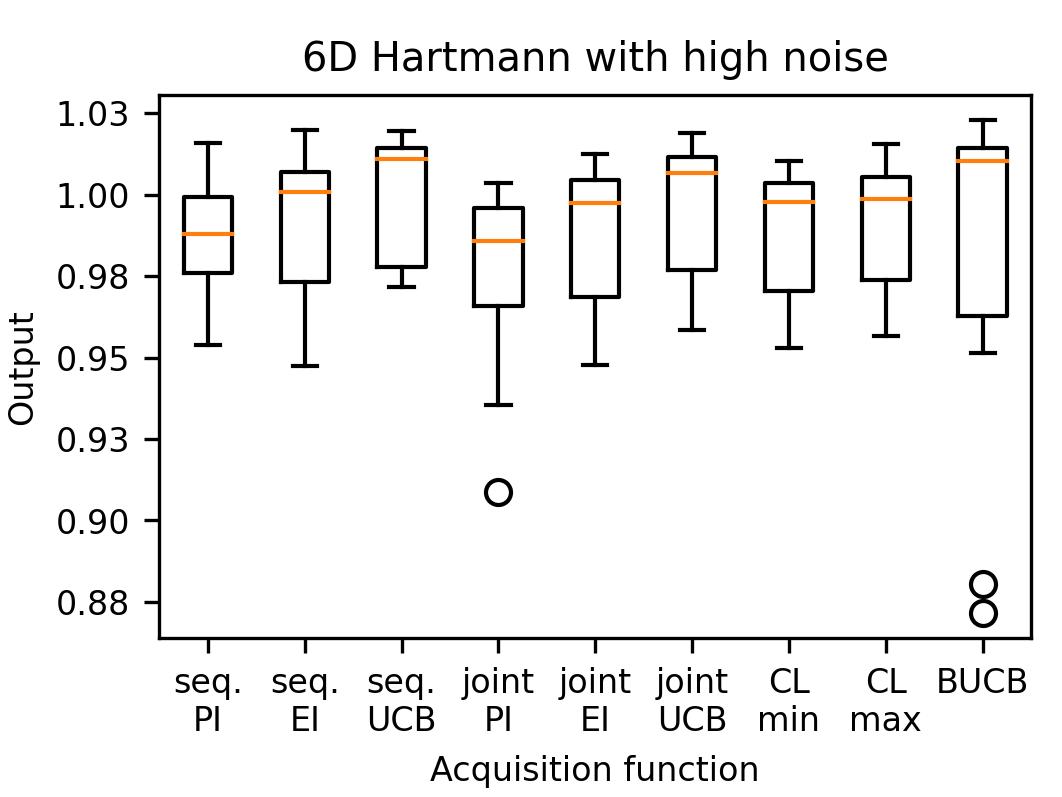}
    \end{minipage}

    \caption{Box plots of best values found for multi-point acquisition functions with five initial starting point per input dimension. The orange line indicates the median, the box extends from the lower to the upper quartile range and the whiskers indicate the range of the best values without the outliers shown as circles.}
    \label{fig:SFigure 2}
\end{figure}

To investigate if there are statistically significant differences between the methods, the non-parametric Friedman test is performed on the best output value found for every replication for each algorithm. This conducts a hypothesis test to see if there is a difference in performance between the algorithms. First, we focus on the analytical single-point algorithms of Section 3.2.1 of the main article. The tests reported in Table \ref{table:STable 1} provide p-values which are all well below the 0.1\% significance level, indicating that we have very strong evidence of a difference in the performance of the different methods. We can investigate these differences using the box plots in Figure \ref{fig:SFigure 1}, that show where the differences between the algorithms lie. We see especially large differences for the Ackley function. The results are similar for the multi-point approaches of Section 3.2.4 of the main article. We again see significant differences between the optimal output found by the different methods (at the 0.1\% level) using the Friedman tests reported in Table \ref{table:STable 2}, and the box plots in Figure \ref{fig:SFigure 2} reinforce these results.

\begin{table}[h]
    \centering
    \caption{Friedman test for analytical single-point acquisition function with five initial starting points per input dimension.}
    \begin{tabular}{ l | r r }
        Test function & Test statistic & p-value \\[0.2cm]
        \hline \hline \\
        Griewank & 169.10 & 1.14e-34 \\[0.15cm]
        Hartmann & 81.86 & 3.42e-16 \\[0.15cm]
        Noisy Hartmann & 78.80 & 1.50e-15 \\[0.15cm]
        Ackley & 199.10 &4.43e-41 \\[0.15cm]
    \end{tabular}
    \label{table:STable 1}
\end{table}

\begin{table}[h]
    \vspace{10mm}
    \centering
    \caption{Friedman test for multi-point acquisition function with five initial starting points per input dimension.}
    \begin{tabular}{ l | r r }
        Test function & Test statistic & p-value \\[0.2cm]
        \hline \hline \\
        Griewank & 254.97 & 1.53e-50 \\[0.15cm]
        Hartmann & 159.59 & 1.94e-30 \\[0.15cm]
        Noisy Hartmann & 137.30 & 8.64e-26 \\[0.15cm]
        Ackley & 362.75 & 1.72e-73 \\[0.15cm]
    \end{tabular}
    \label{table:STable 2}
\end{table}

\color{black}

\newpage
\FloatBarrier
\section{Supplementary Figures}

\begin{figure}[h]
\vspace{10mm}
    \begin{minipage}[t]{0.02\textwidth}
        \vspace{-45mm}
        (A)
    \end{minipage}
    \begin{minipage}[t]{0.48\textwidth}
        \centering
        \includegraphics[width=0.8\linewidth]{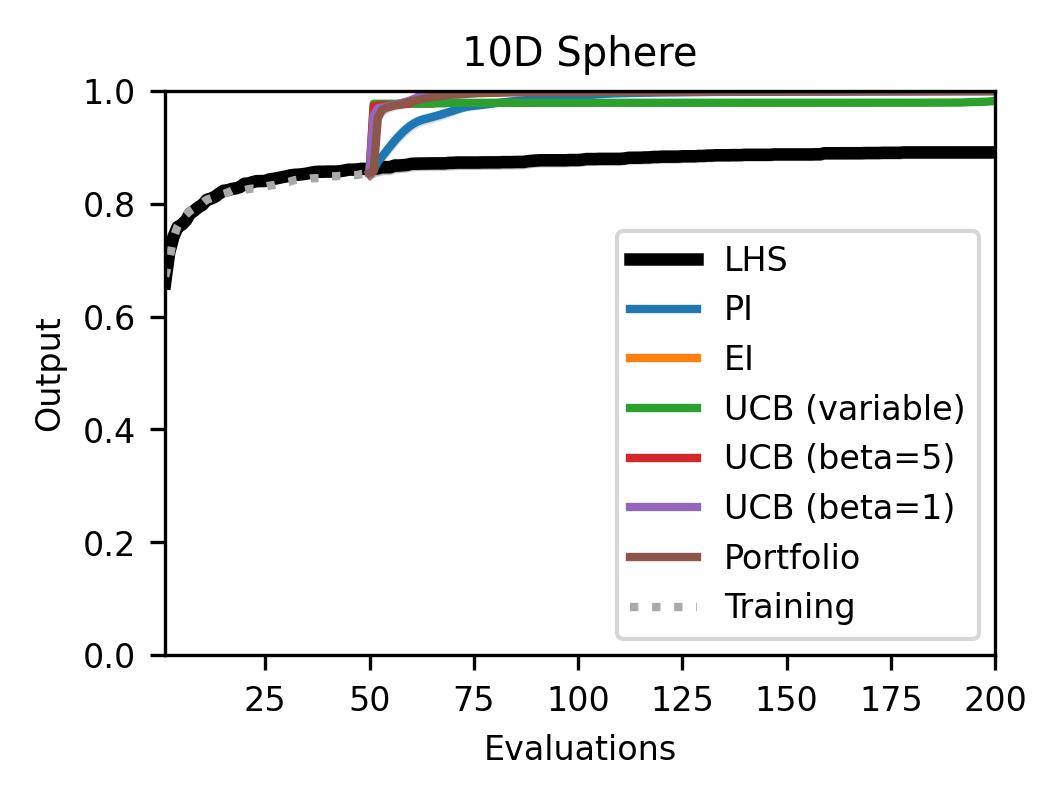}
    \end{minipage}
    \begin{minipage}[t]{0.02\textwidth}
        \vspace{-45mm}
        (B)
    \end{minipage}
    \begin{minipage}[t]{0.48\textwidth}
    \centering
        \includegraphics[width=0.8\linewidth]{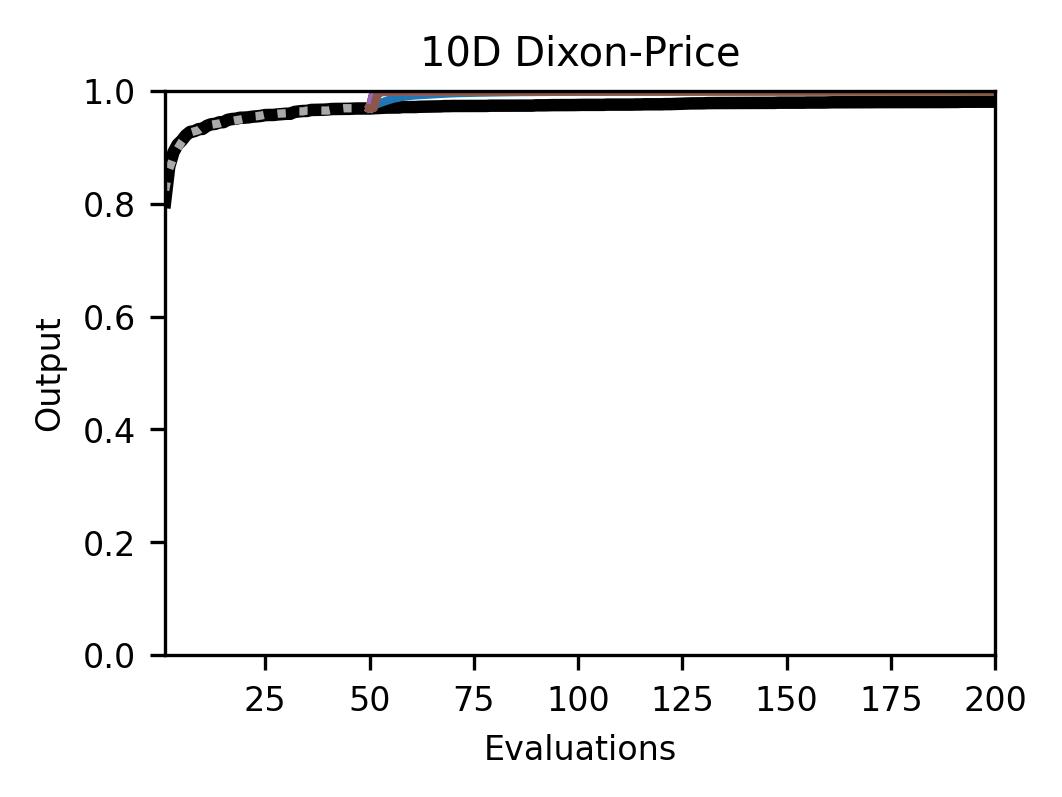}
    \end{minipage}
    \begin{minipage}[t]{0.02\textwidth}
        \vspace{-45mm}
        (C)
    \end{minipage}
    \begin{minipage}[t]{0.48\textwidth}
        \centering
        \includegraphics[width=0.8\linewidth]{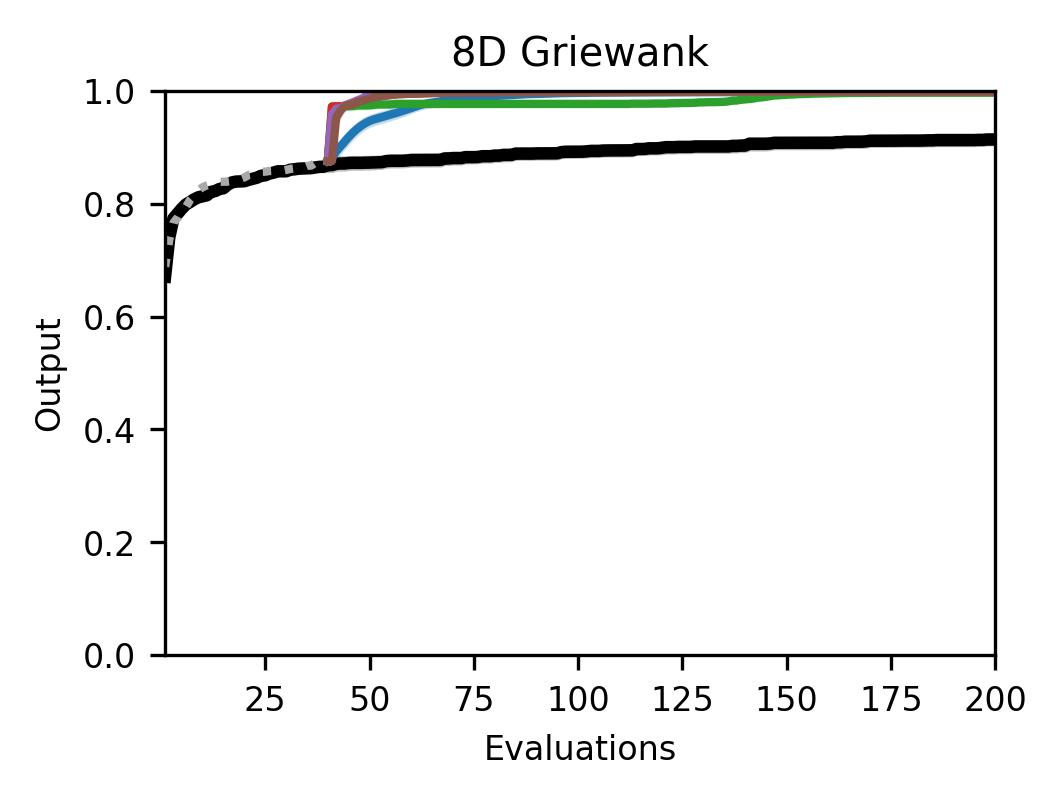}
    \end{minipage}
    \begin{minipage}[t]{0.02\textwidth}
        \vspace{-45mm}
        (D)
    \end{minipage}
    \begin{minipage}[t]{0.48\textwidth}
    \centering
        \includegraphics[width=0.8\linewidth]{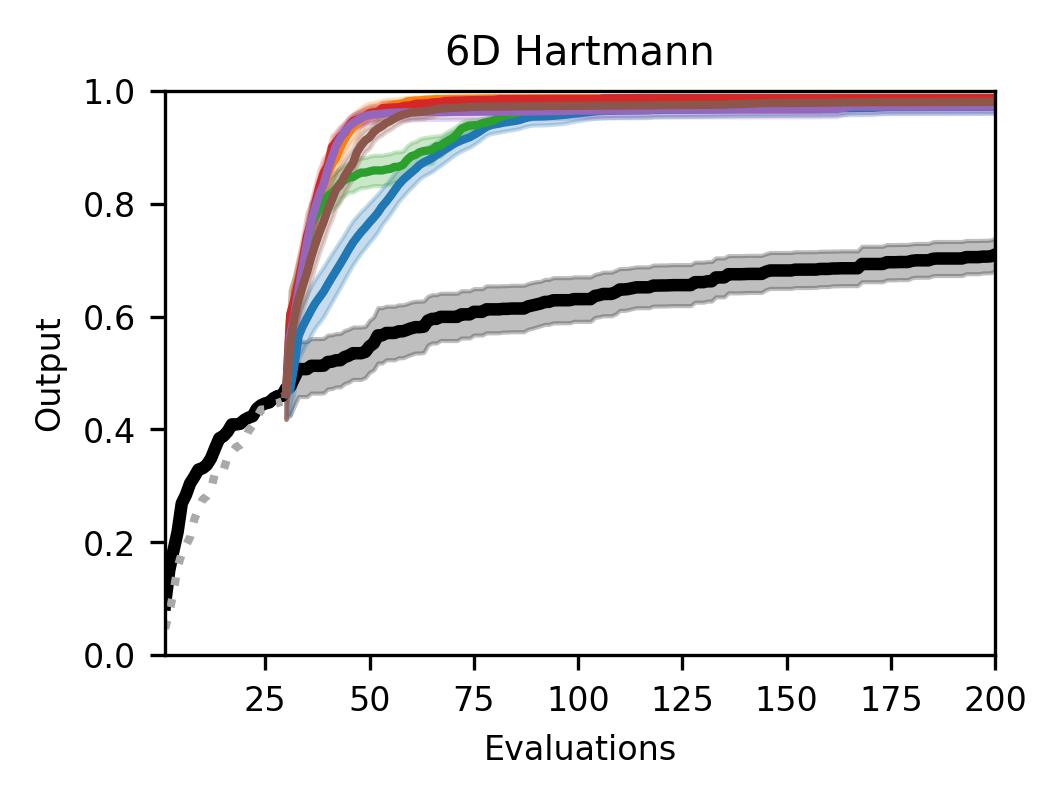}
    \end{minipage}
    \begin{minipage}[t]{0.02\textwidth}
        \vspace{-45mm}
        (E)
    \end{minipage}
    \begin{minipage}[t]{0.48\textwidth}
        \centering
        \includegraphics[width=0.8\linewidth]{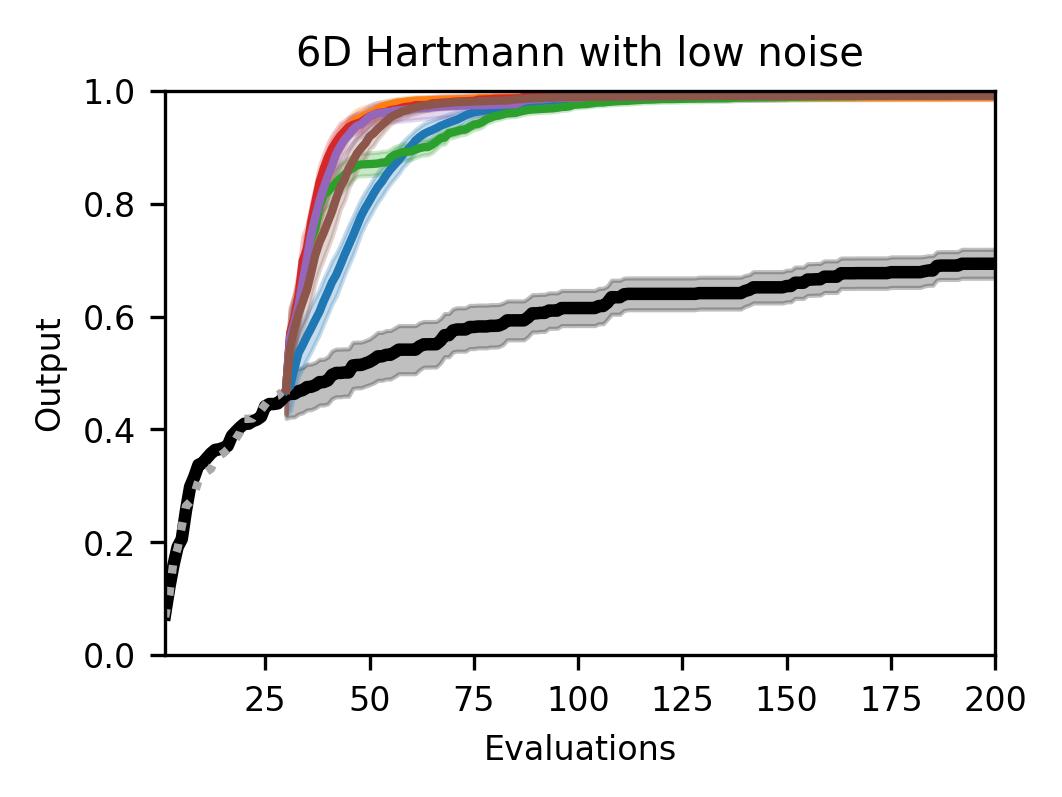}
    \end{minipage}
    \begin{minipage}[t]{0.02\textwidth}
        \vspace{-45mm}
        (F)
    \end{minipage}
    \begin{minipage}[t]{0.48\textwidth}
    \centering
        \includegraphics[width=0.8\linewidth]{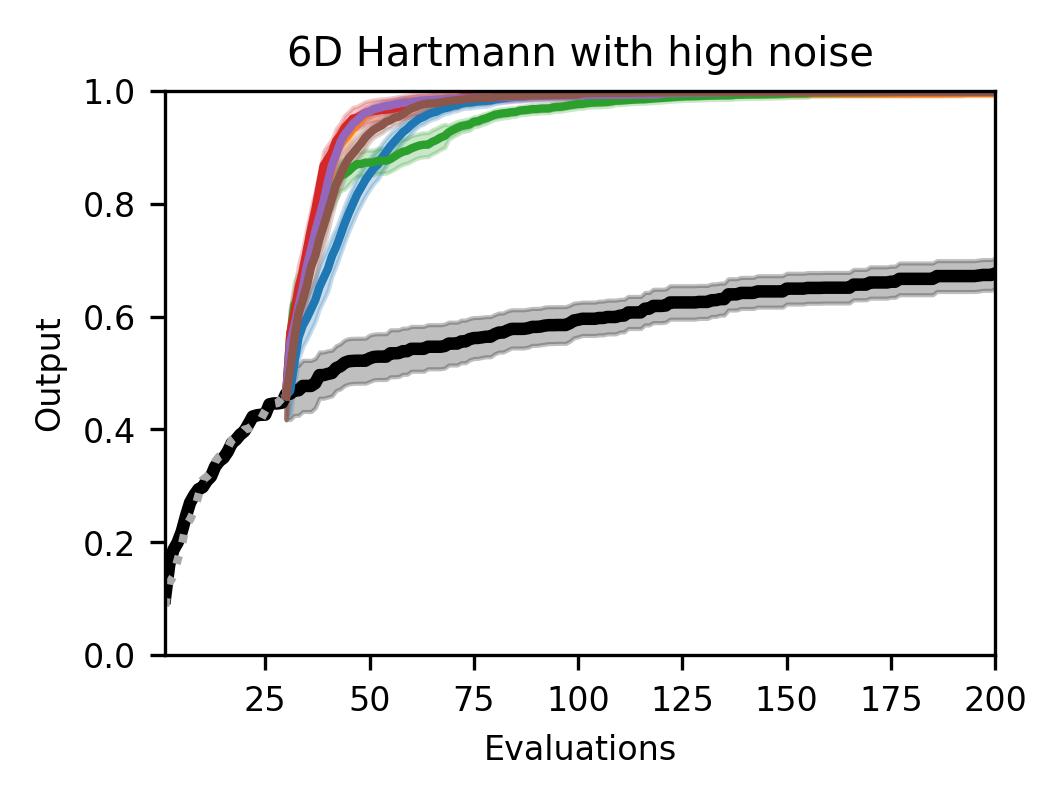}
    \end{minipage}
    \begin{minipage}[t]{0.02\textwidth}
        \vspace{-45mm}
        (G)
    \end{minipage}
    \begin{minipage}[t]{0.48\textwidth}
        \centering
        \includegraphics[width=0.8\linewidth]{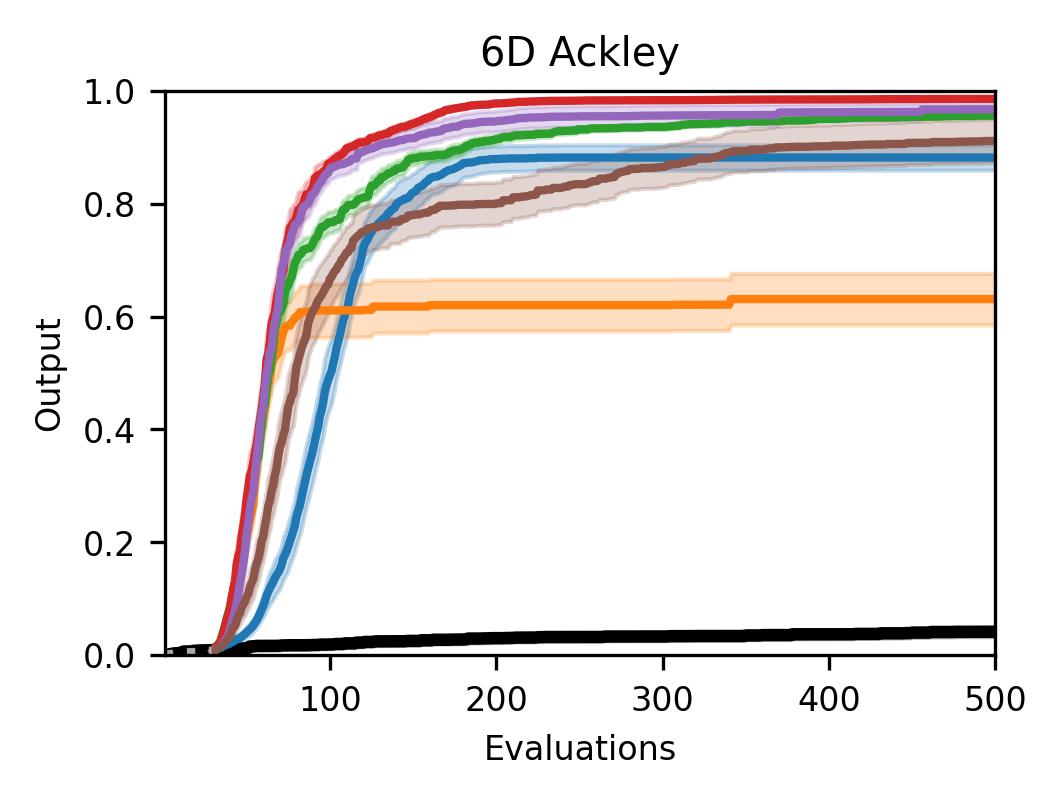}
    \end{minipage}
    \begin{minipage}[t]{0.02\textwidth}
        \vspace{-45mm}
        (H)
    \end{minipage}
    \begin{minipage}[t]{0.48\textwidth}
    \centering
        \includegraphics[width=0.8\linewidth]{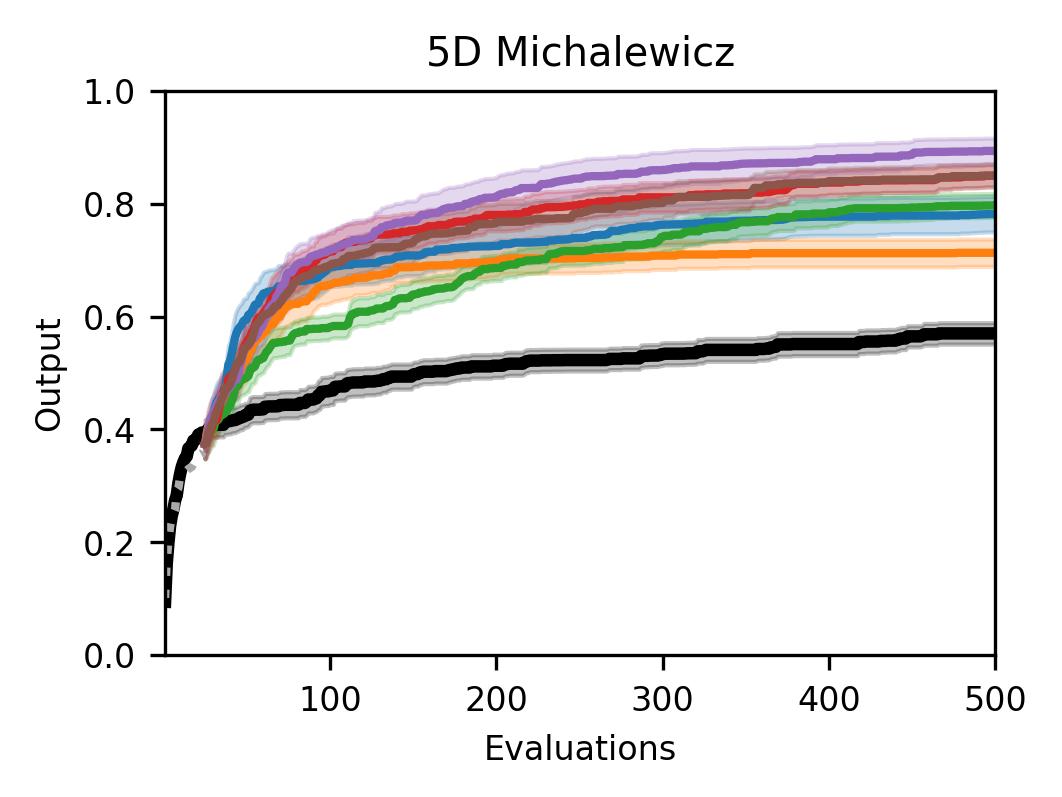}
    \end{minipage}

    \caption{Performance plots for analytical single-point acquisition functions with five initial starting points per input dimension. Solid lines represent the mean over the 50 runs while the shaded area represents the 95\% confidence intervals.}
\end{figure}

\begin{figure}
    \begin{minipage}[t]{0.02\textwidth}
        \vspace{-45mm}
        (A)
    \end{minipage}
    \begin{minipage}[t]{0.48\textwidth}
        \centering
        \includegraphics[width=0.8\linewidth]{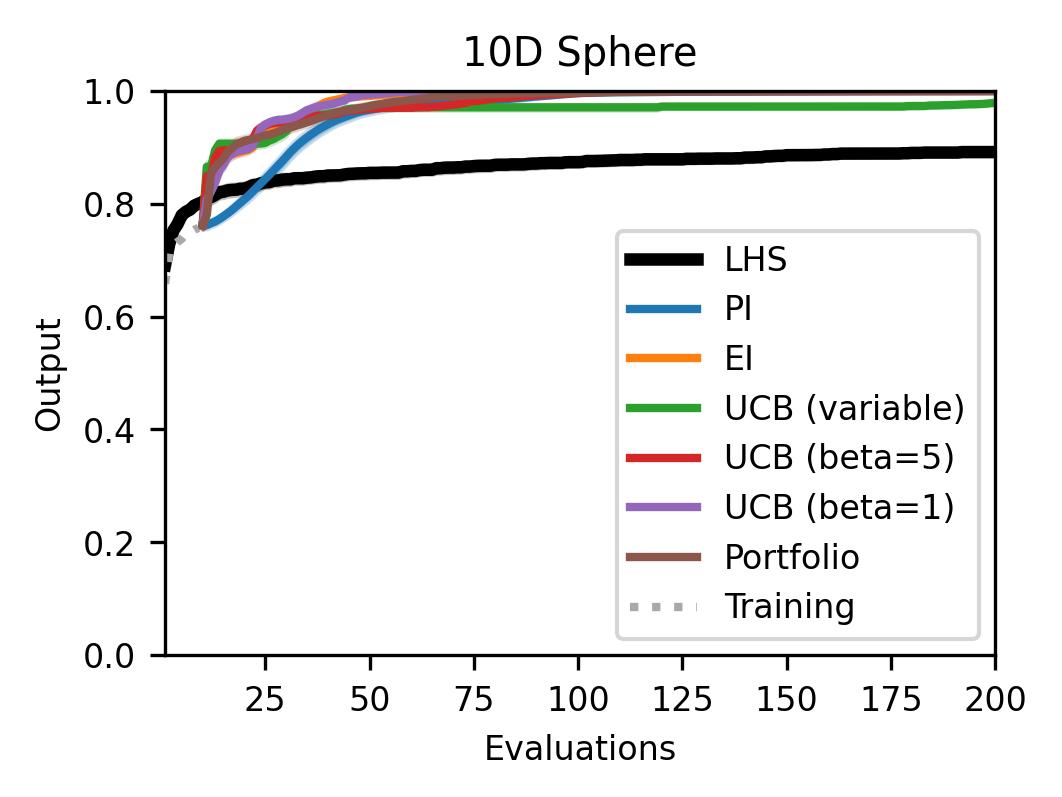}
    \end{minipage}
    \begin{minipage}[t]{0.02\textwidth}
        \vspace{-45mm}
        (B)
    \end{minipage}
    \begin{minipage}[t]{0.48\textwidth}
    \centering
        \includegraphics[width=0.8\linewidth]{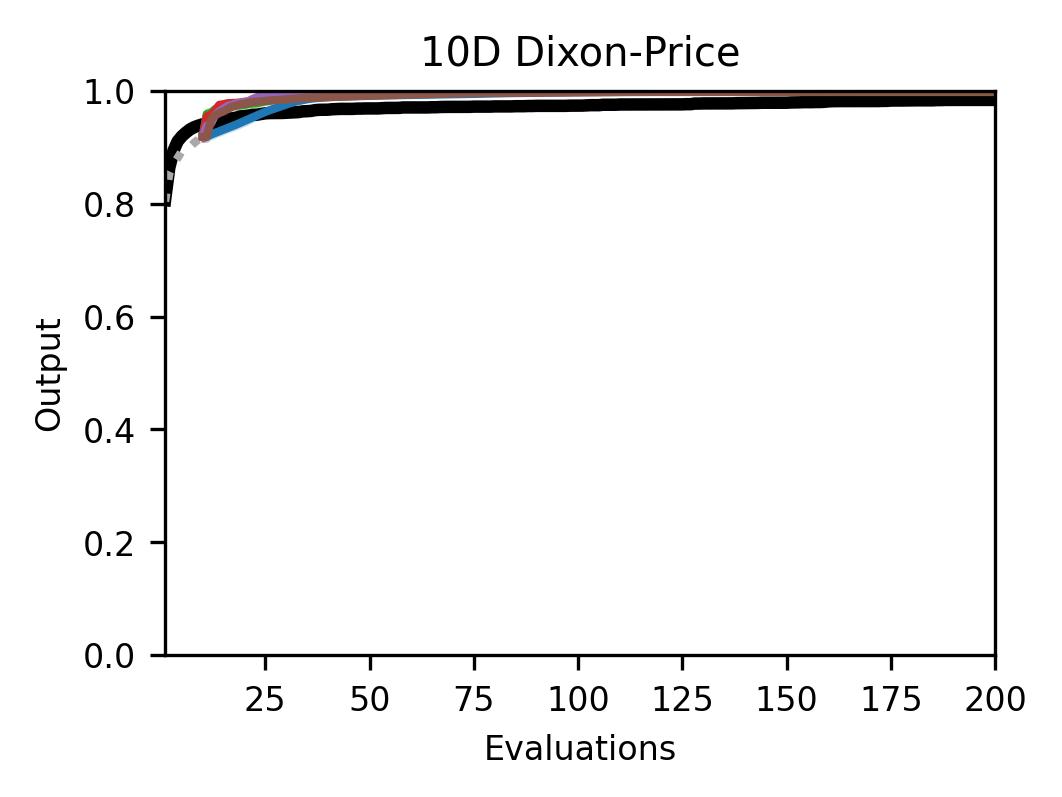}
    \end{minipage}
    \begin{minipage}[t]{0.02\textwidth}
        \vspace{-45mm}
        (C)
    \end{minipage}
    \begin{minipage}[t]{0.48\textwidth}
        \centering
        \includegraphics[width=0.8\linewidth]{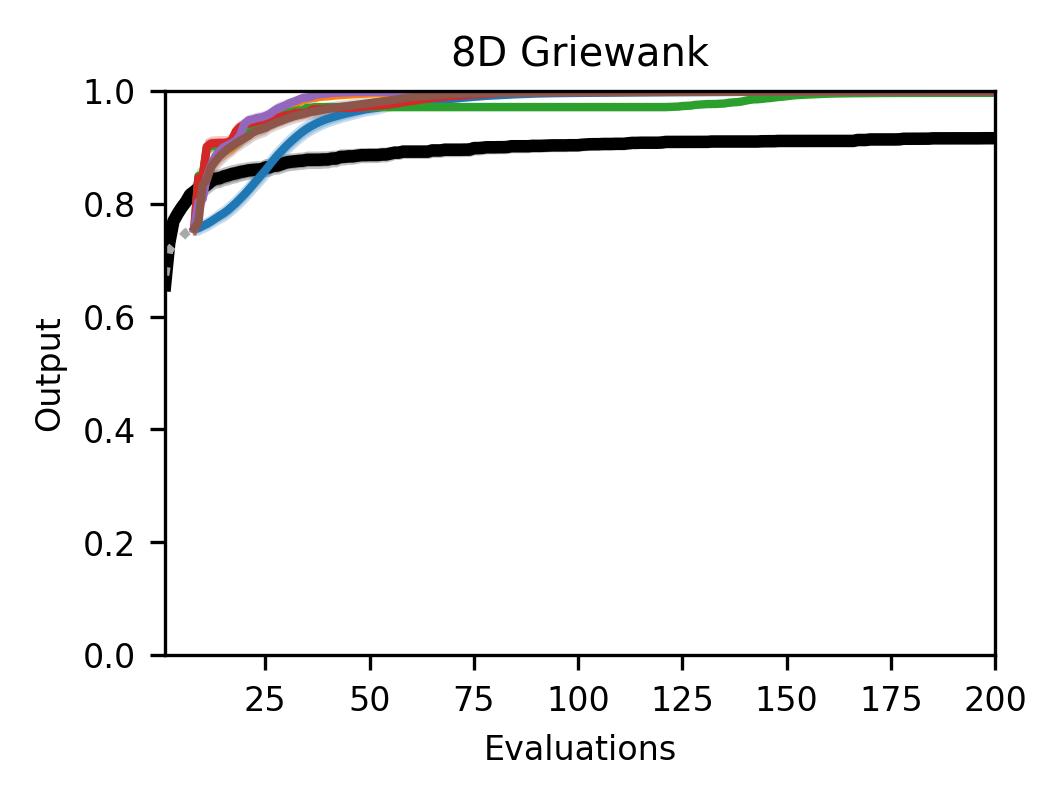}
    \end{minipage}
    \begin{minipage}[t]{0.02\textwidth}
        \vspace{-45mm}
        (D)
    \end{minipage}
    \begin{minipage}[t]{0.48\textwidth}
    \centering
        \includegraphics[width=0.8\linewidth]{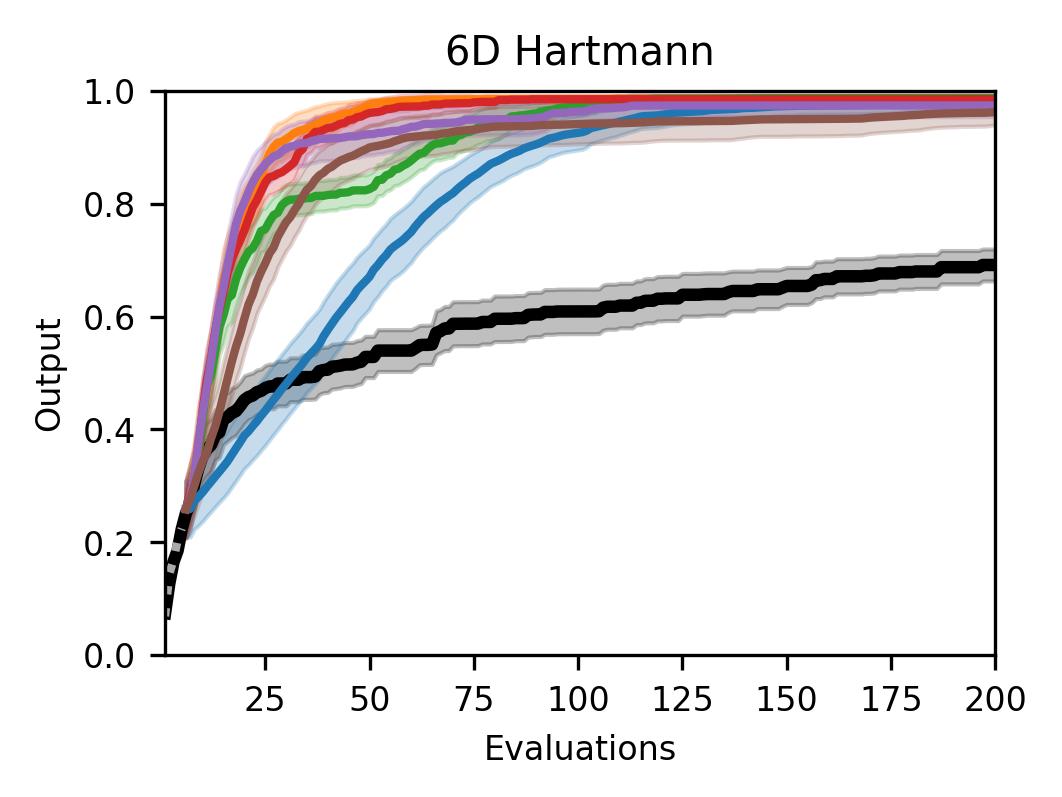}
    \end{minipage}
    \begin{minipage}[t]{0.02\textwidth}
        \vspace{-45mm}
        (E)
    \end{minipage}
    \begin{minipage}[t]{0.48\textwidth}
        \centering
        \includegraphics[width=0.8\linewidth]{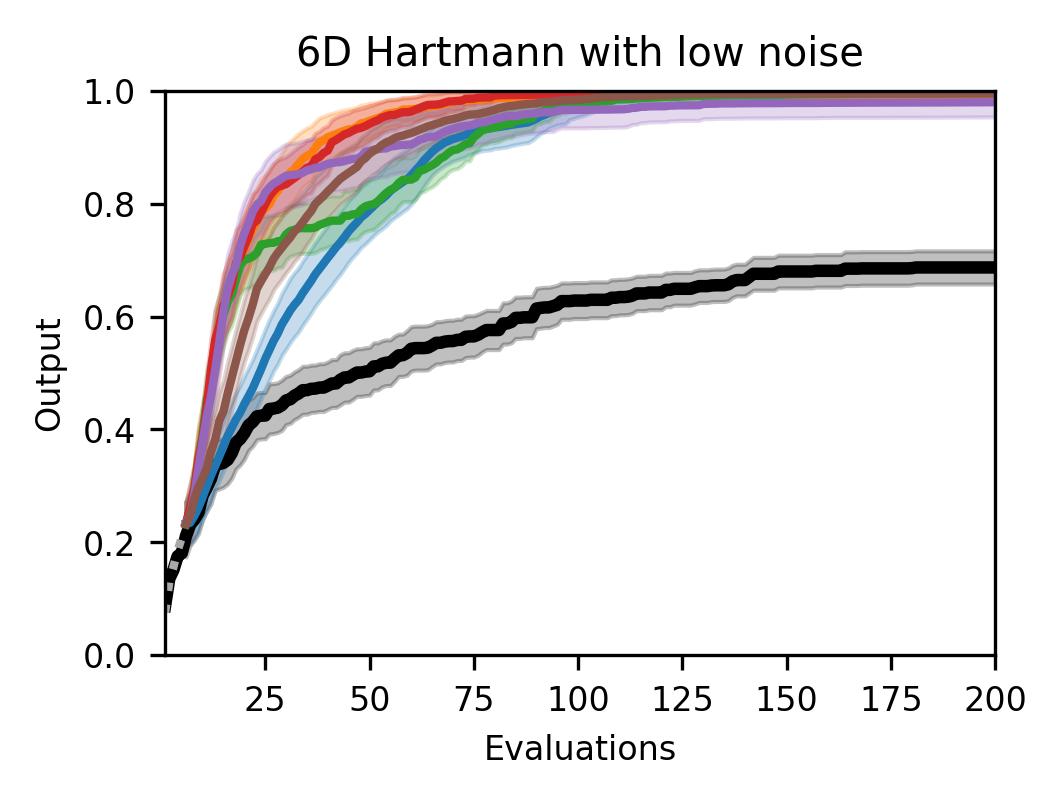}
    \end{minipage}
    \begin{minipage}[t]{0.02\textwidth}
        \vspace{-45mm}
        (F)
    \end{minipage}
    \begin{minipage}[t]{0.48\textwidth}
    \centering
        \includegraphics[width=0.8\linewidth]{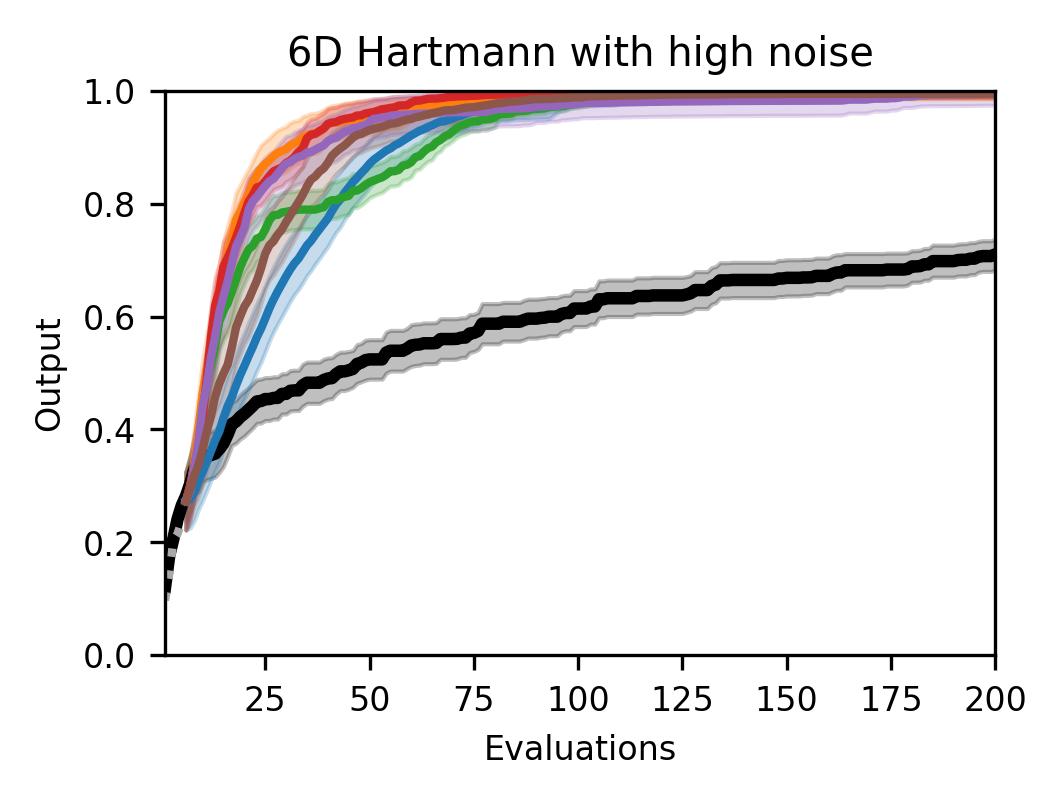}
    \end{minipage}
    \begin{minipage}[t]{0.02\textwidth}
        \vspace{-45mm}
        (G)
    \end{minipage}
    \begin{minipage}[t]{0.48\textwidth}
        \centering
        \includegraphics[width=0.8\linewidth]{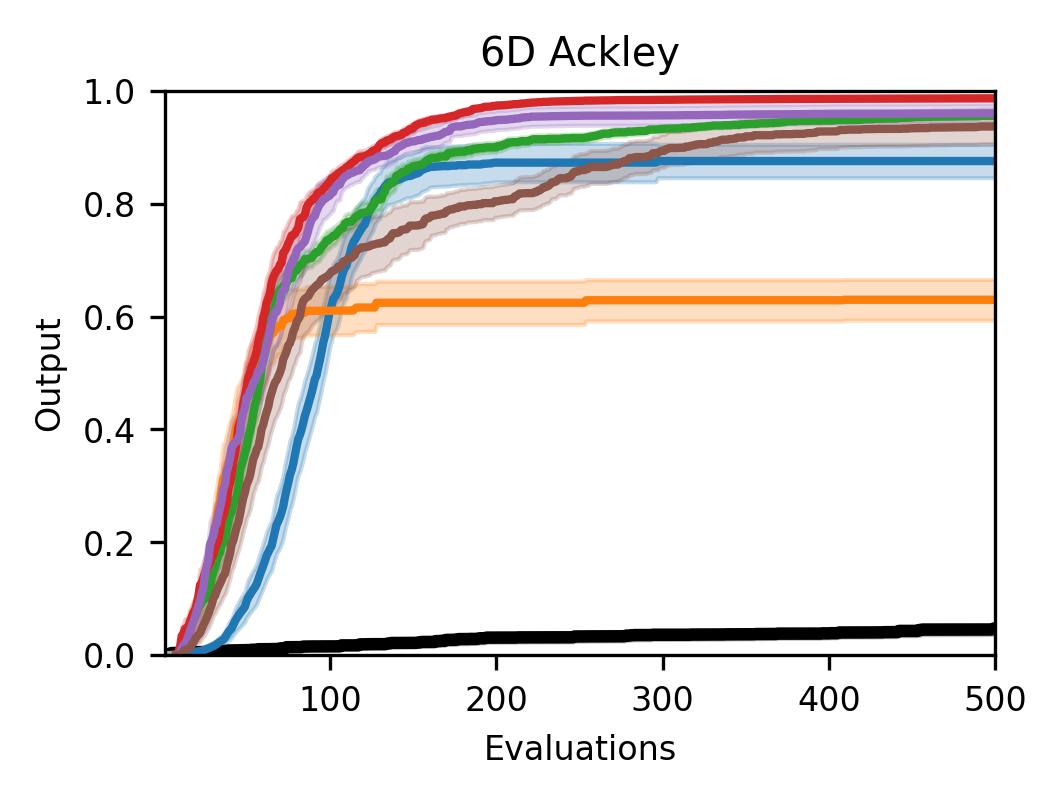}
    \end{minipage}
    \begin{minipage}[t]{0.02\textwidth}
        \vspace{-45mm}
        (H)
    \end{minipage}
    \begin{minipage}[t]{0.48\textwidth}
    \centering
        \includegraphics[width=0.8\linewidth]{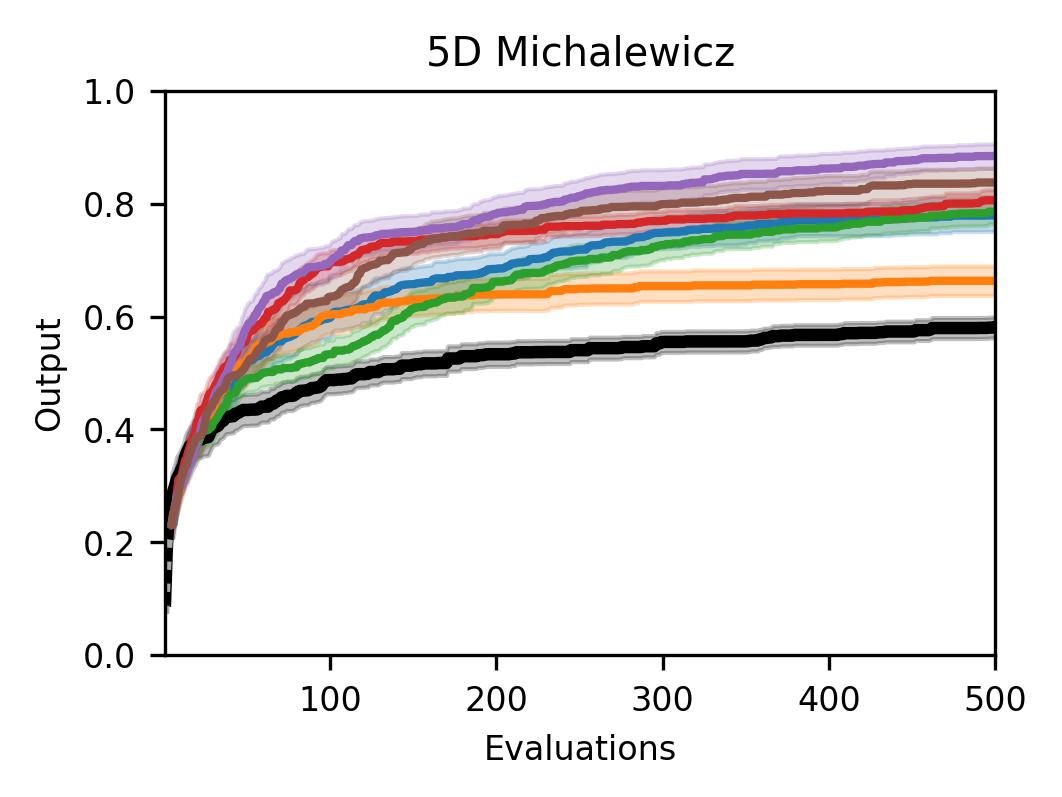}
    \end{minipage}

    \caption{Performance plots for analytical single-point acquisition functions with one initial starting point per input dimension. Solid lines represent the mean over the 50 runs while the shaded area represents the 95\% confidence intervals.}
\end{figure}

\begin{figure}
    \begin{minipage}[t]{0.02\textwidth}
        \vspace{-45mm}
        (A)
    \end{minipage}
    \begin{minipage}[t]{0.48\textwidth}
        \centering
        \includegraphics[width=0.8\linewidth]{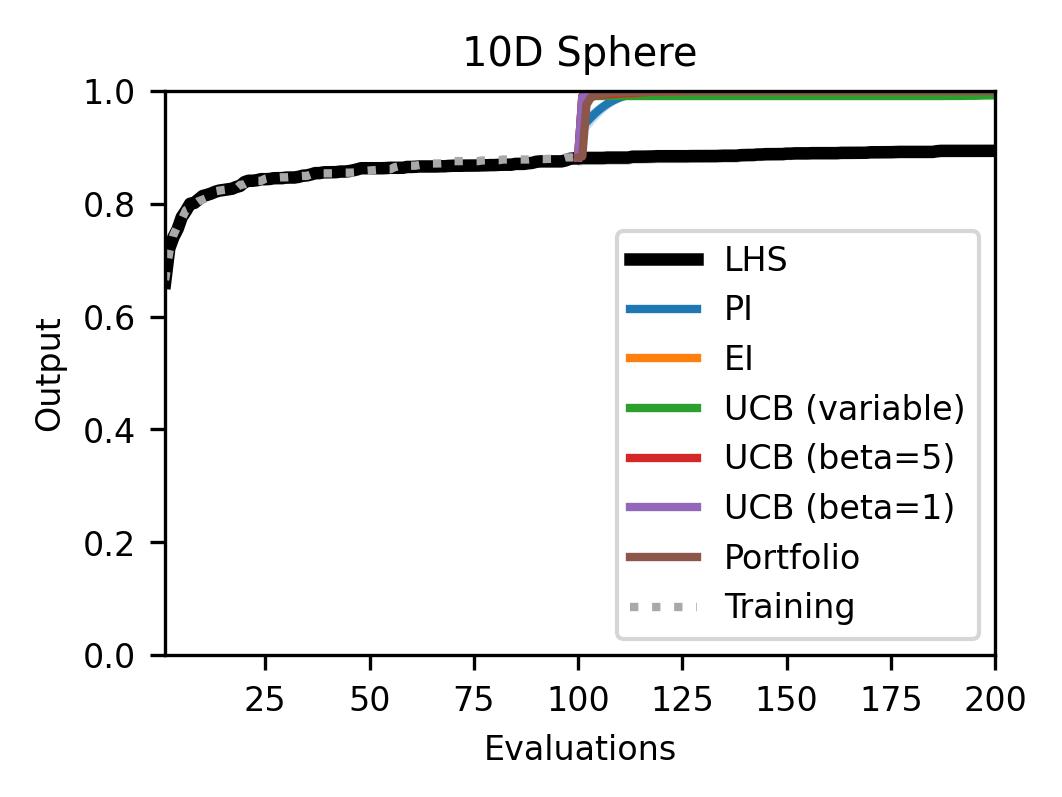}
    \end{minipage}
    \begin{minipage}[t]{0.02\textwidth}
        \vspace{-45mm}
        (B)
    \end{minipage}
    \begin{minipage}[t]{0.48\textwidth}
    \centering
        \includegraphics[width=0.8\linewidth]{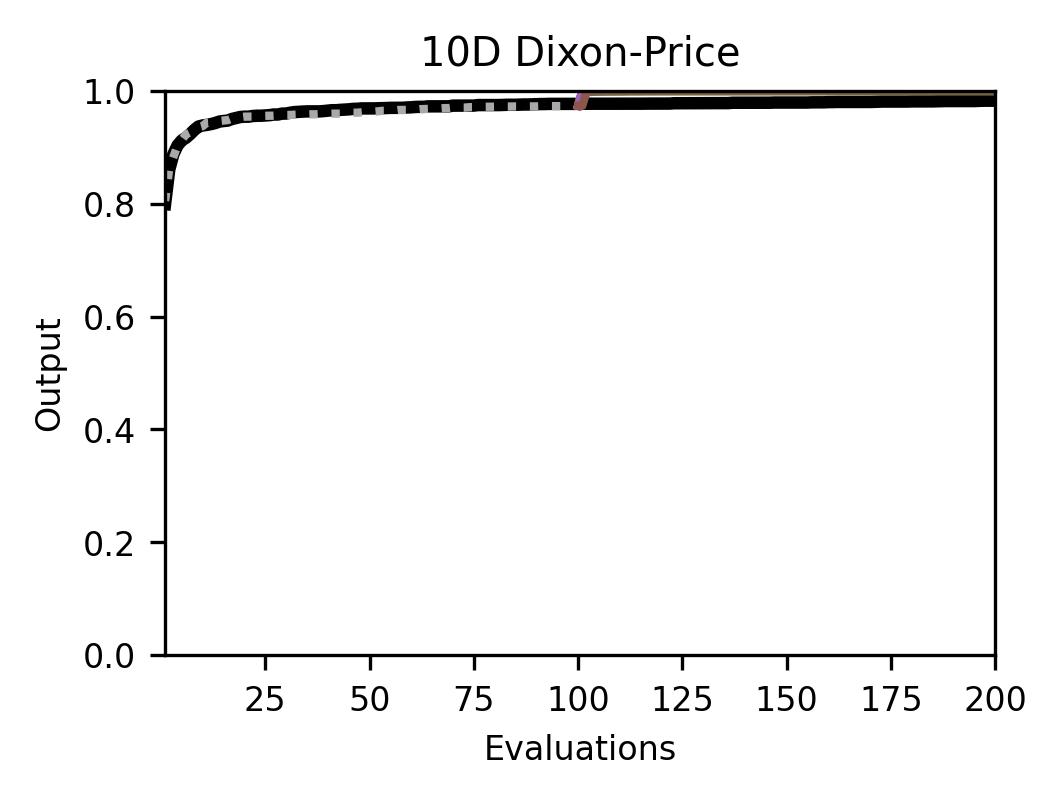}
    \end{minipage}
    \begin{minipage}[t]{0.02\textwidth}
        \vspace{-45mm}
        (C)
    \end{minipage}
    \begin{minipage}[t]{0.48\textwidth}
        \centering
        \includegraphics[width=0.8\linewidth]{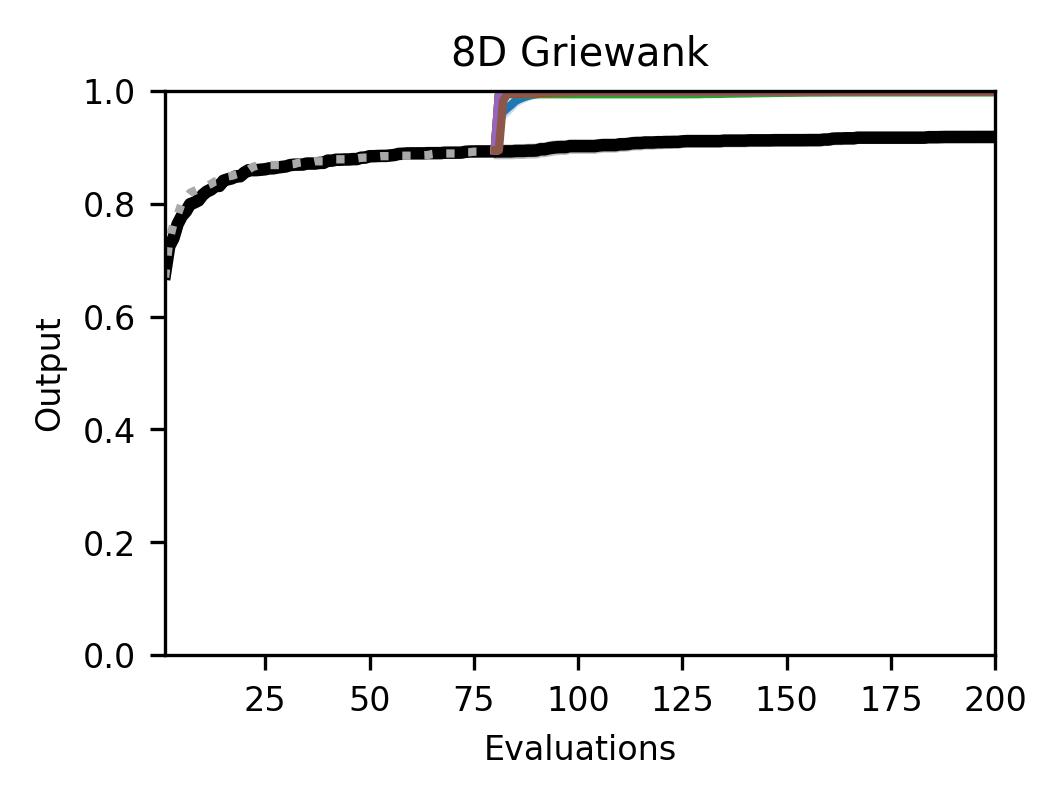}
    \end{minipage}
    \begin{minipage}[t]{0.02\textwidth}
        \vspace{-45mm}
        (D)
    \end{minipage}
    \begin{minipage}[t]{0.48\textwidth}
    \centering
        \includegraphics[width=0.8\linewidth]{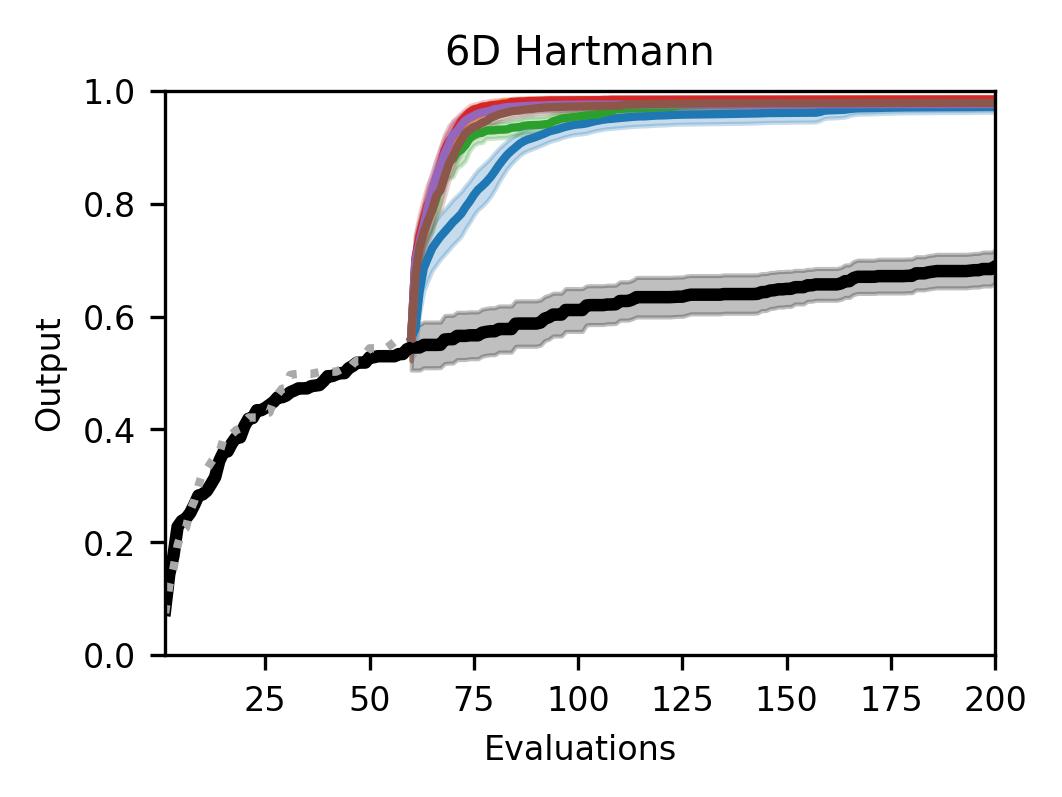}
    \end{minipage}
    \begin{minipage}[t]{0.02\textwidth}
        \vspace{-45mm}
        (E)
    \end{minipage}
    \begin{minipage}[t]{0.48\textwidth}
        \centering
        \includegraphics[width=0.8\linewidth]{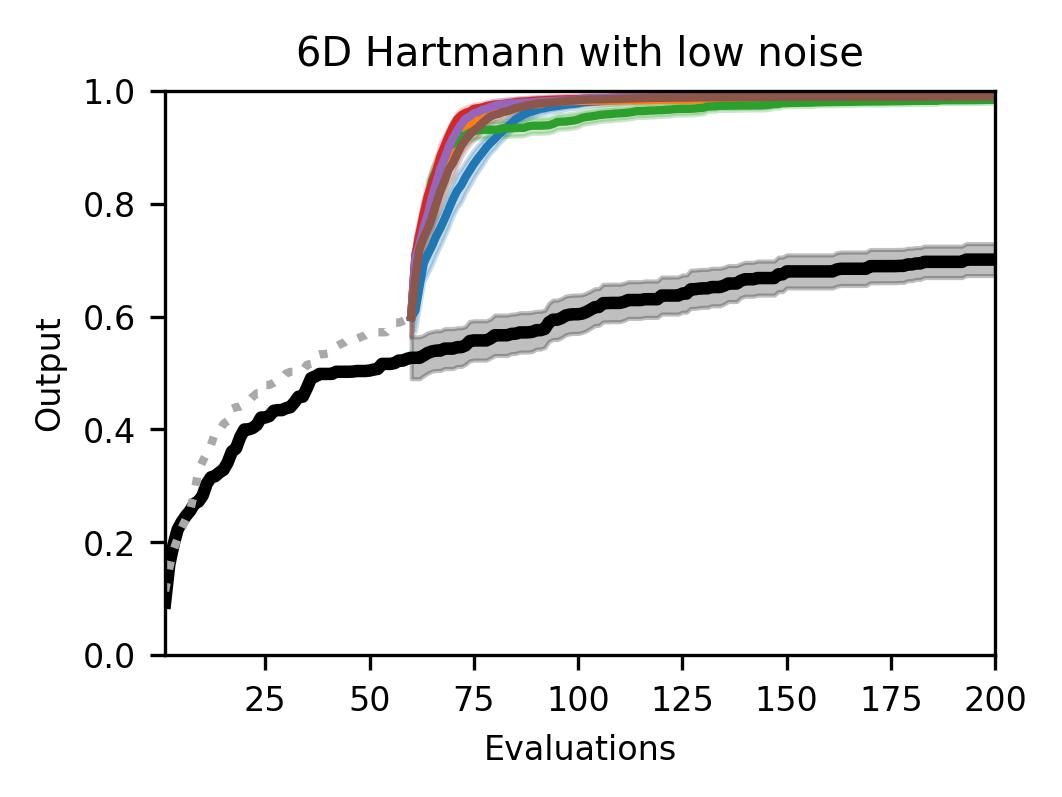}
    \end{minipage}
    \begin{minipage}[t]{0.02\textwidth}
        \vspace{-45mm}
        (F)
    \end{minipage}
    \begin{minipage}[t]{0.48\textwidth}
    \centering
        \includegraphics[width=0.8\linewidth]{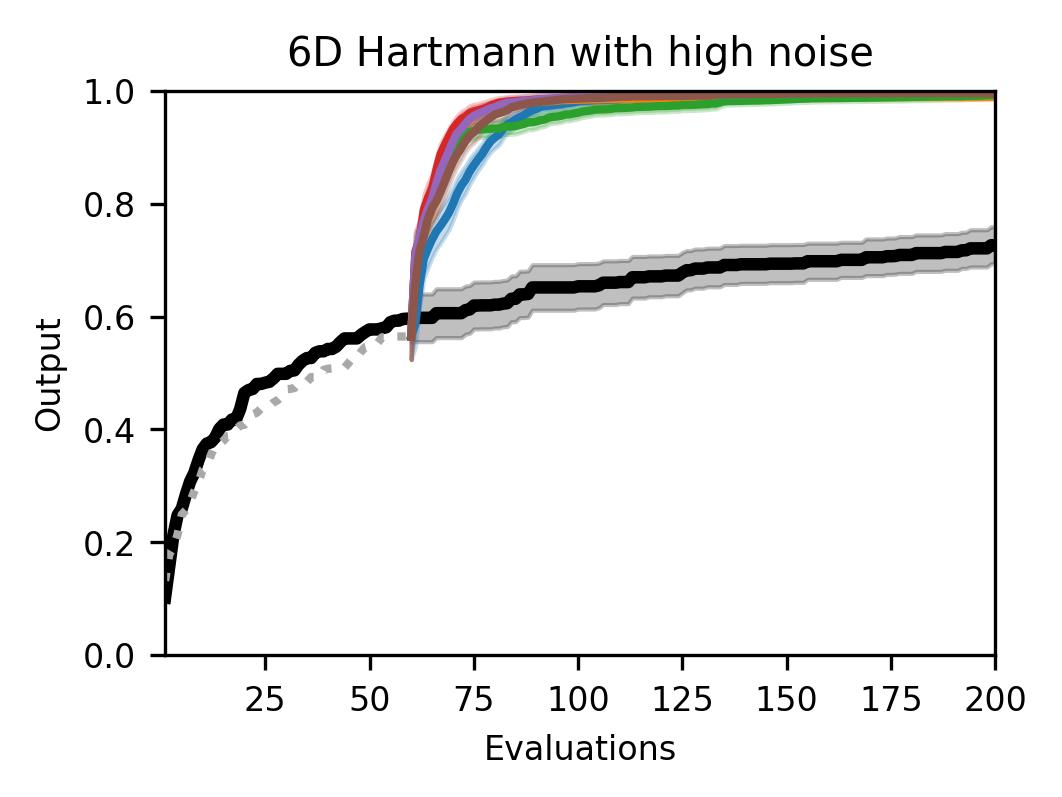}
    \end{minipage}
    \begin{minipage}[t]{0.02\textwidth}
        \vspace{-45mm}
        (G)
    \end{minipage}
    \begin{minipage}[t]{0.48\textwidth}
        \centering
        \includegraphics[width=0.8\linewidth]{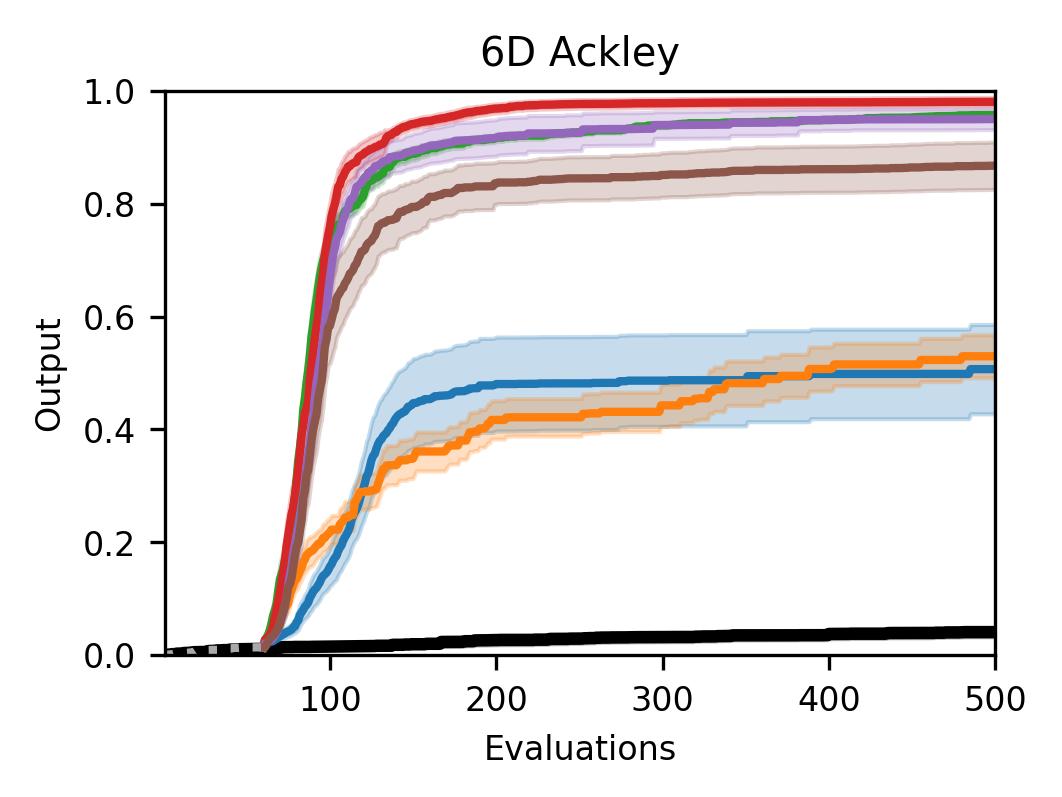}
    \end{minipage}
    \begin{minipage}[t]{0.02\textwidth}
        \vspace{-45mm}
        (H)
    \end{minipage}
    \begin{minipage}[t]{0.48\textwidth}
    \centering
        \includegraphics[width=0.8\linewidth]{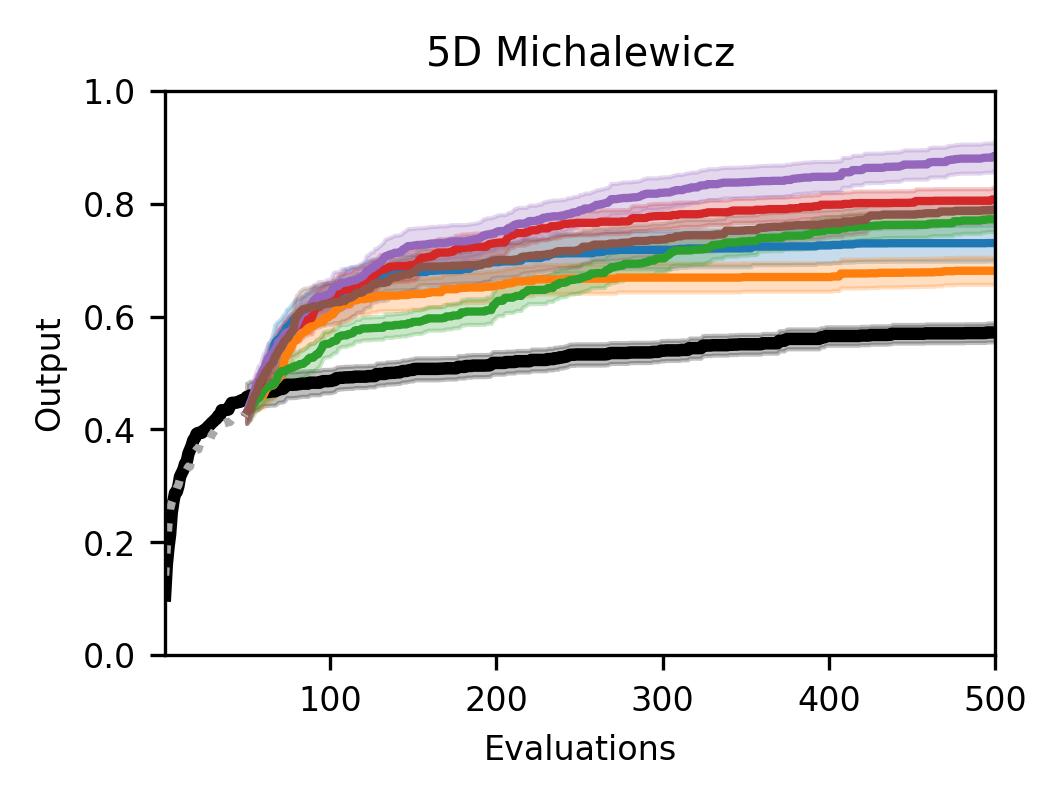}
    \end{minipage}

    \caption{Performance plots for analytical single-point acquisition functions with ten initial starting points per input dimension. Solid lines represent the mean over the 50 runs while the shaded area represents the 95\% confidence intervals.}
\end{figure}

\begin{figure}
    \begin{minipage}[t]{0.02\textwidth}
        \vspace{-45mm}
        (A)
    \end{minipage}
    \begin{minipage}[t]{0.48\textwidth}
        \centering
        \includegraphics[width=0.8\linewidth]{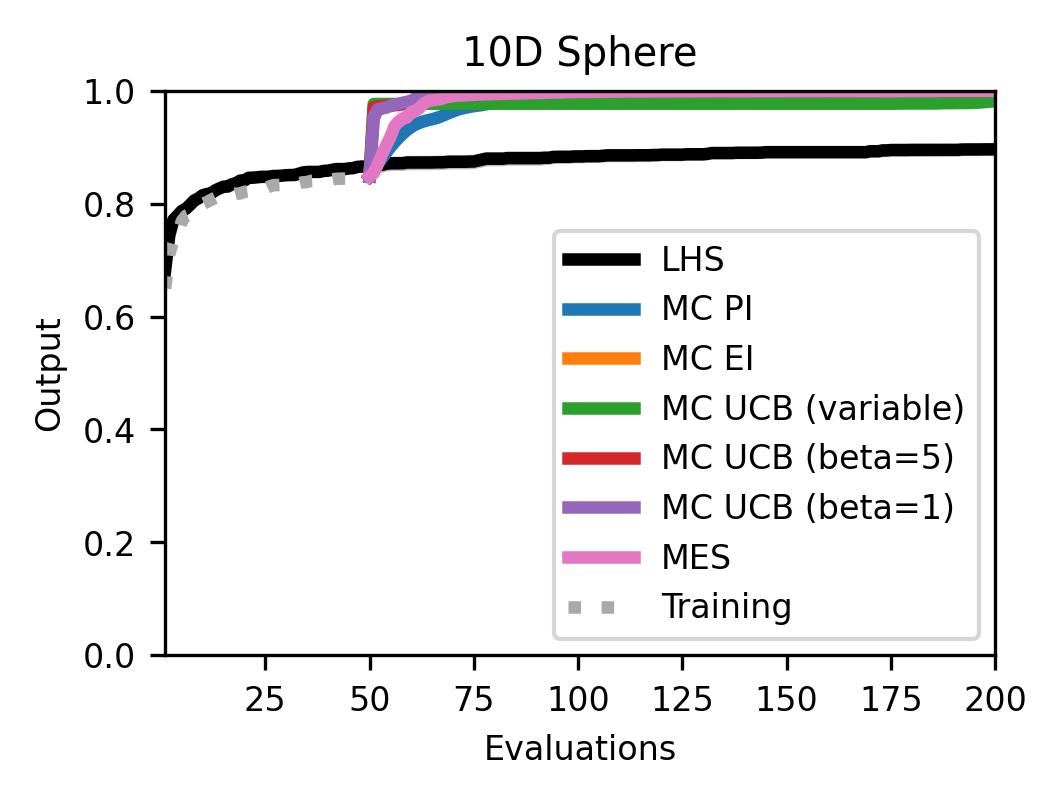}
    \end{minipage}
    \begin{minipage}[t]{0.02\textwidth}
        \vspace{-45mm}
        (B)
    \end{minipage}
    \begin{minipage}[t]{0.48\textwidth}
    \centering
        \includegraphics[width=0.8\linewidth]{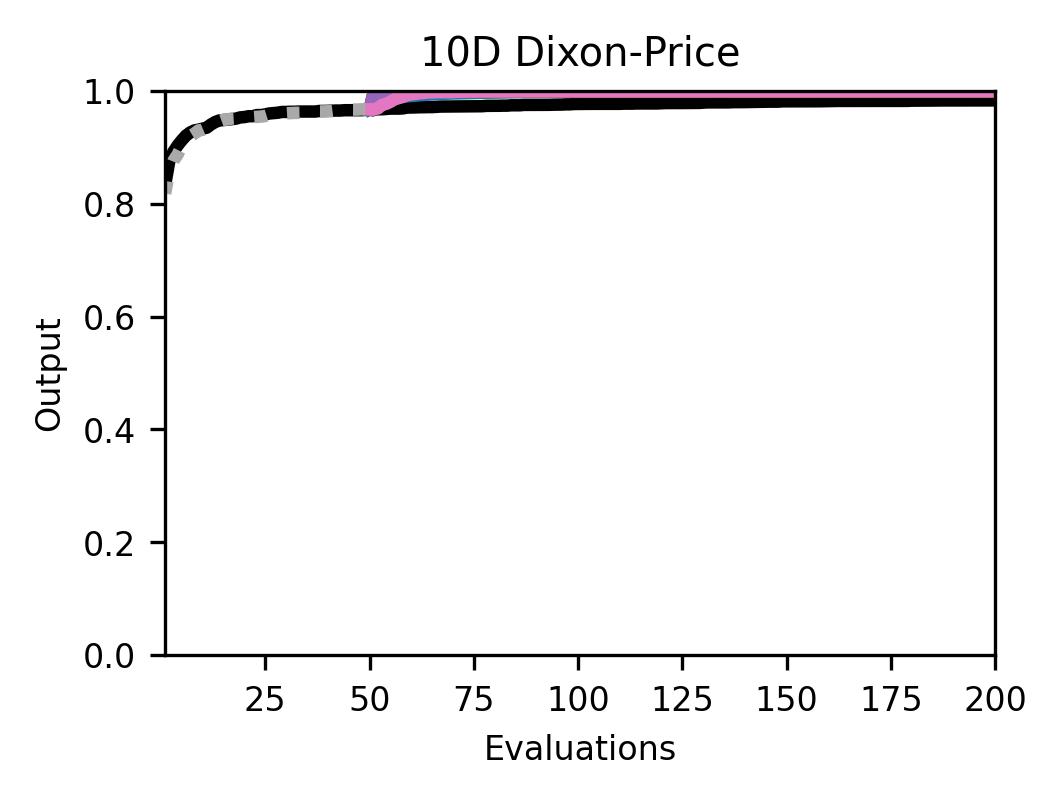}
    \end{minipage}
    \begin{minipage}[t]{0.02\textwidth}
        \vspace{-45mm}
        (C)
    \end{minipage}
    \begin{minipage}[t]{0.48\textwidth}
        \centering
        \includegraphics[width=0.8\linewidth]{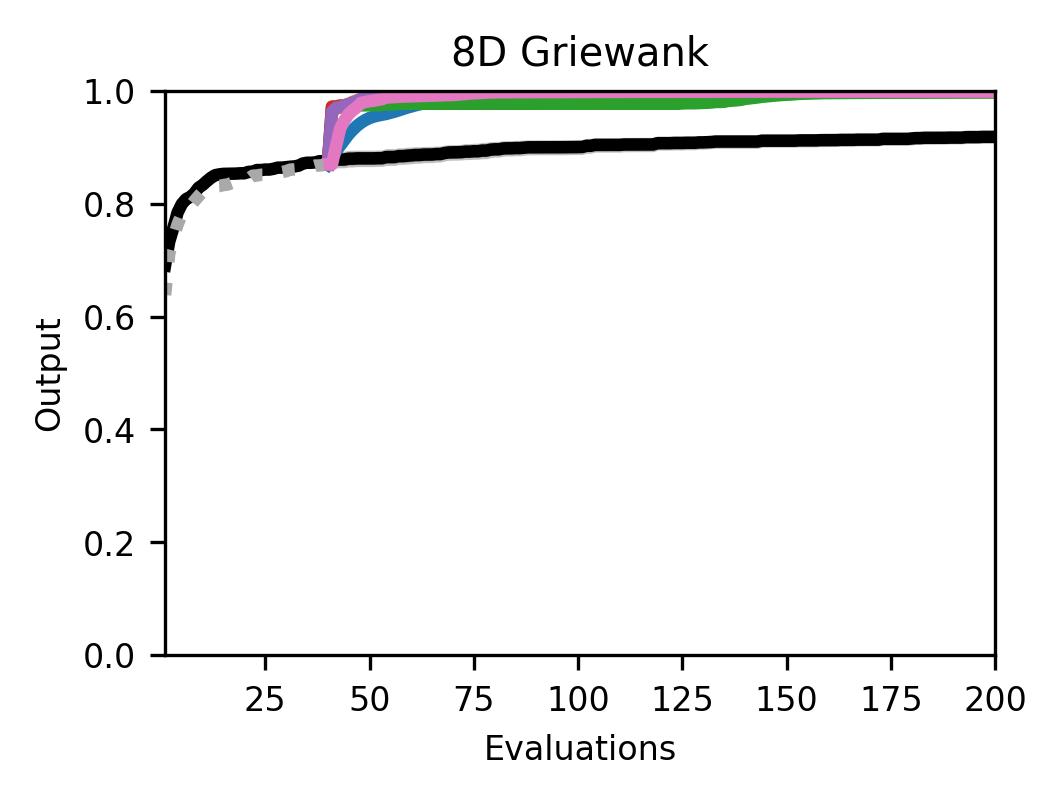}
    \end{minipage}
    \begin{minipage}[t]{0.02\textwidth}
        \vspace{-45mm}
        (D)
    \end{minipage}
    \begin{minipage}[t]{0.48\textwidth}
    \centering
        \includegraphics[width=0.8\linewidth]{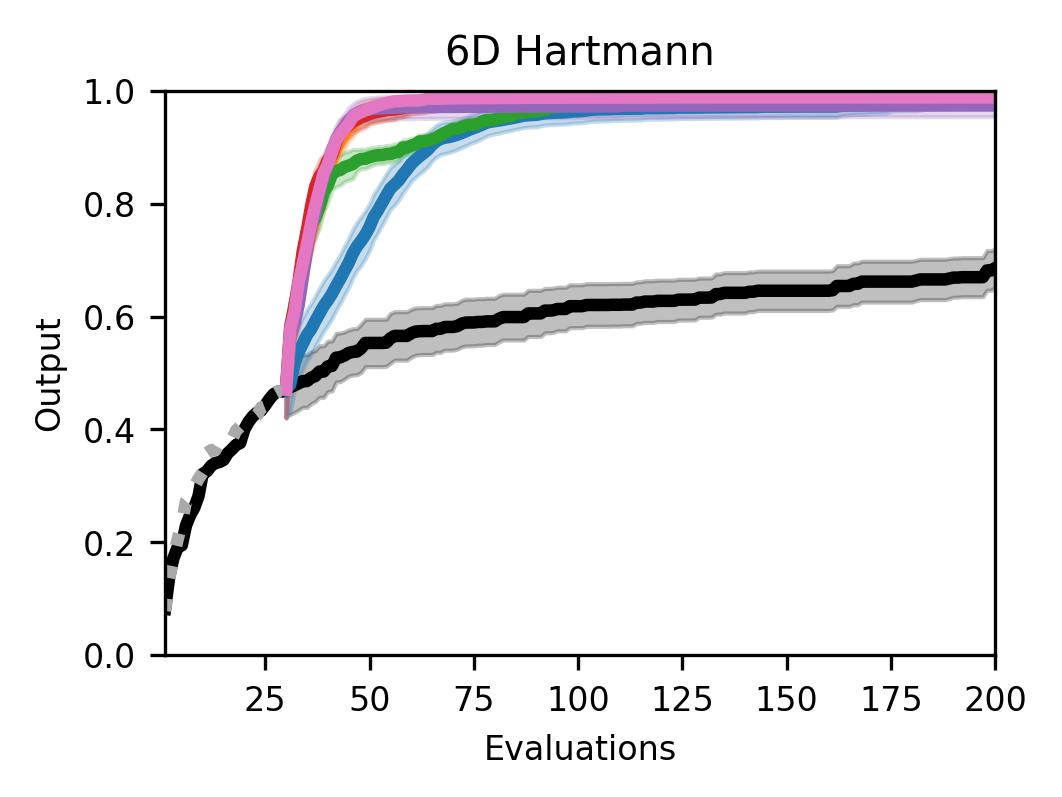}
    \end{minipage}
    \begin{minipage}[t]{0.02\textwidth}
        \vspace{-45mm}
        (E)
    \end{minipage}
    \begin{minipage}[t]{0.48\textwidth}
        \centering
        \includegraphics[width=0.8\linewidth]{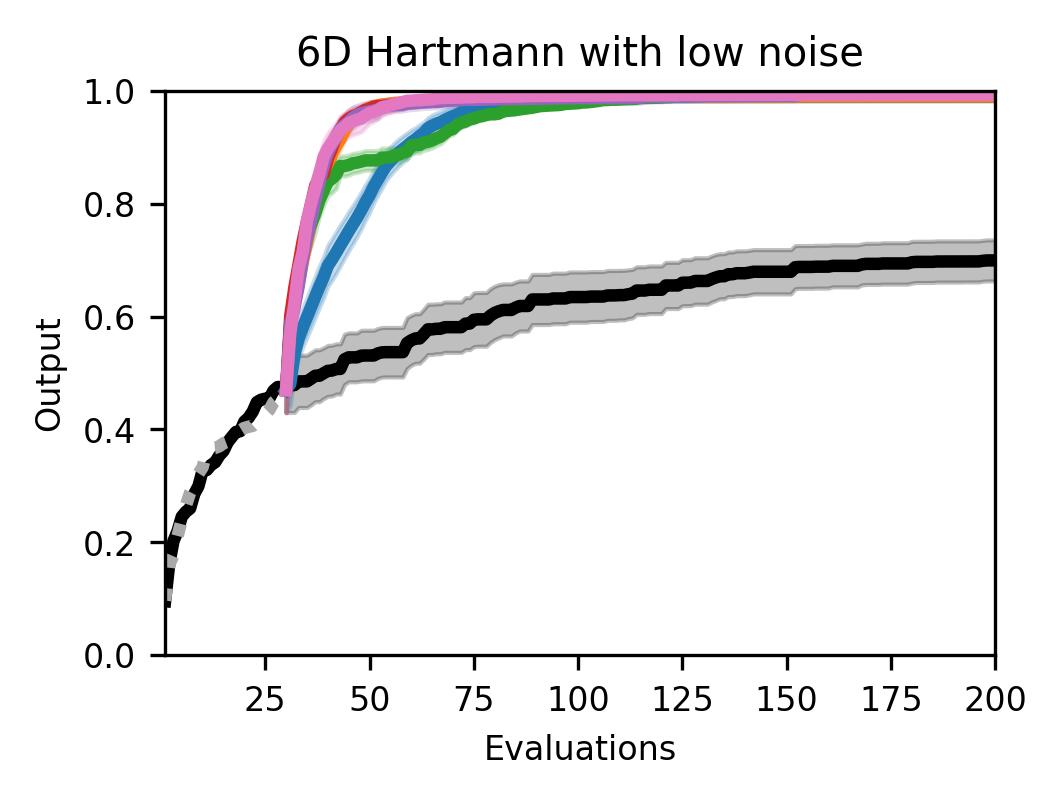}
    \end{minipage}
    \begin{minipage}[t]{0.02\textwidth}
        \vspace{-45mm}
        (F)
    \end{minipage}
    \begin{minipage}[t]{0.48\textwidth}
    \centering
        \includegraphics[width=0.8\linewidth]{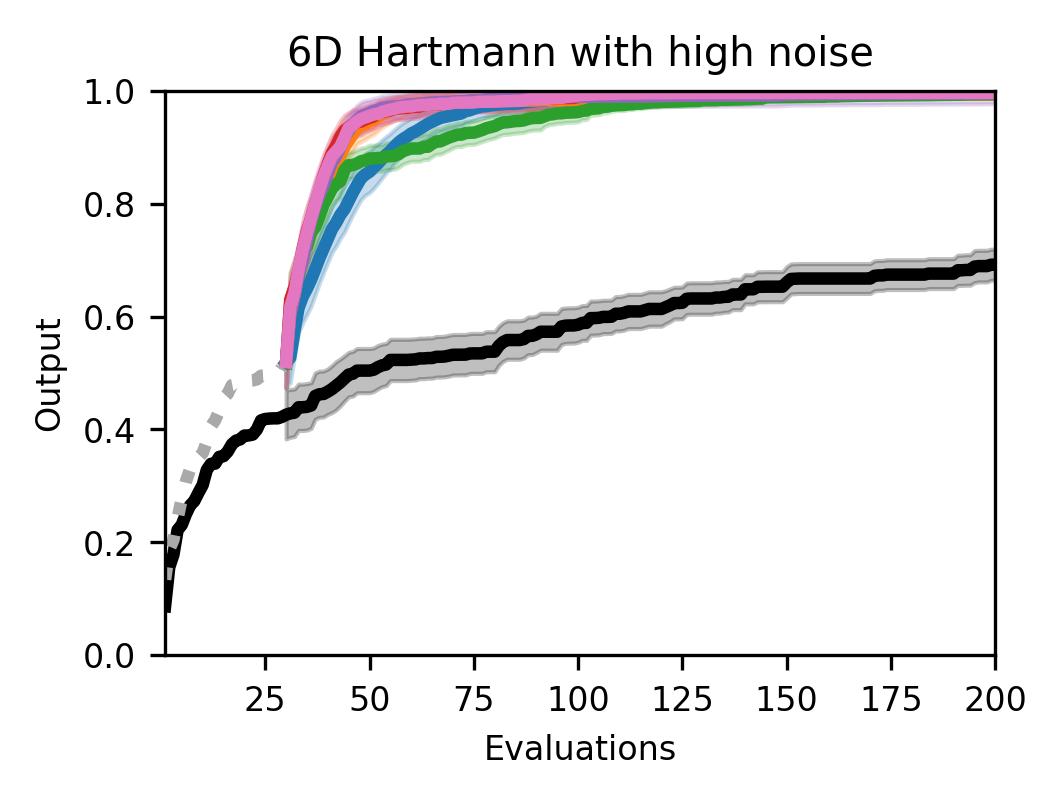}
    \end{minipage}
    \begin{minipage}[t]{0.02\textwidth}
        \vspace{-45mm}
        (G)
    \end{minipage}
    \begin{minipage}[t]{0.48\textwidth}
        \centering
        \includegraphics[width=0.8\linewidth]{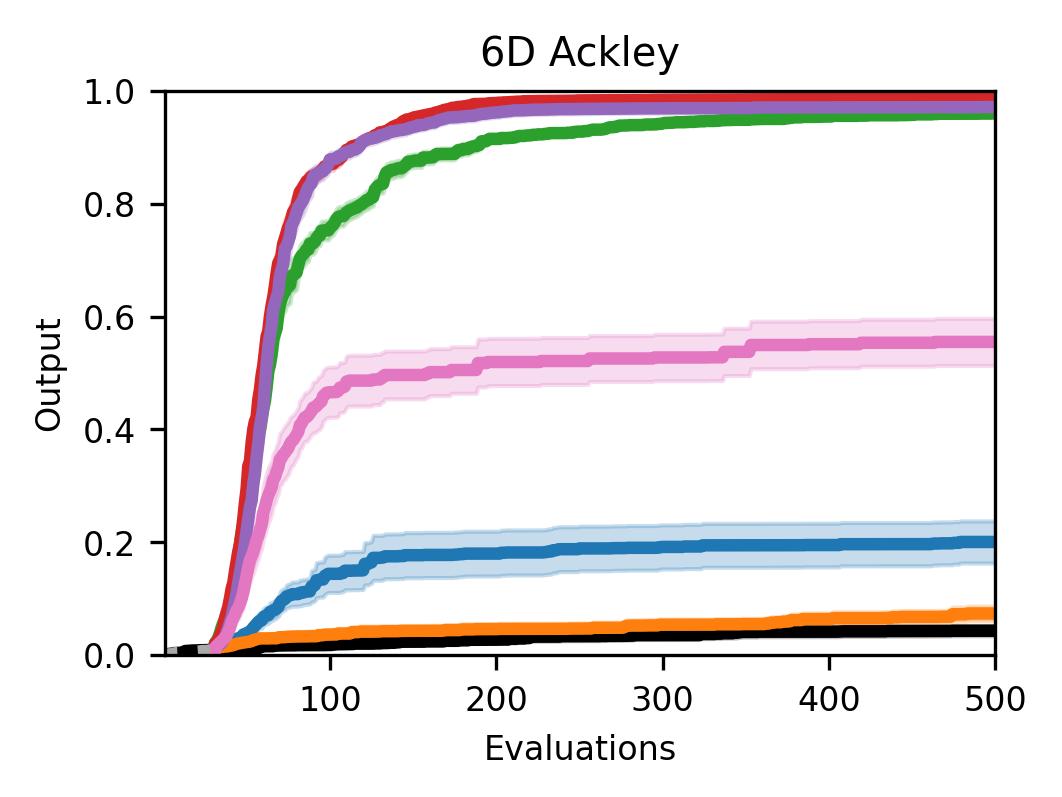}
    \end{minipage}
    \begin{minipage}[t]{0.02\textwidth}
        \vspace{-45mm}
        (H)
    \end{minipage}
    \begin{minipage}[t]{0.48\textwidth}
    \centering
        \includegraphics[width=0.8\linewidth]{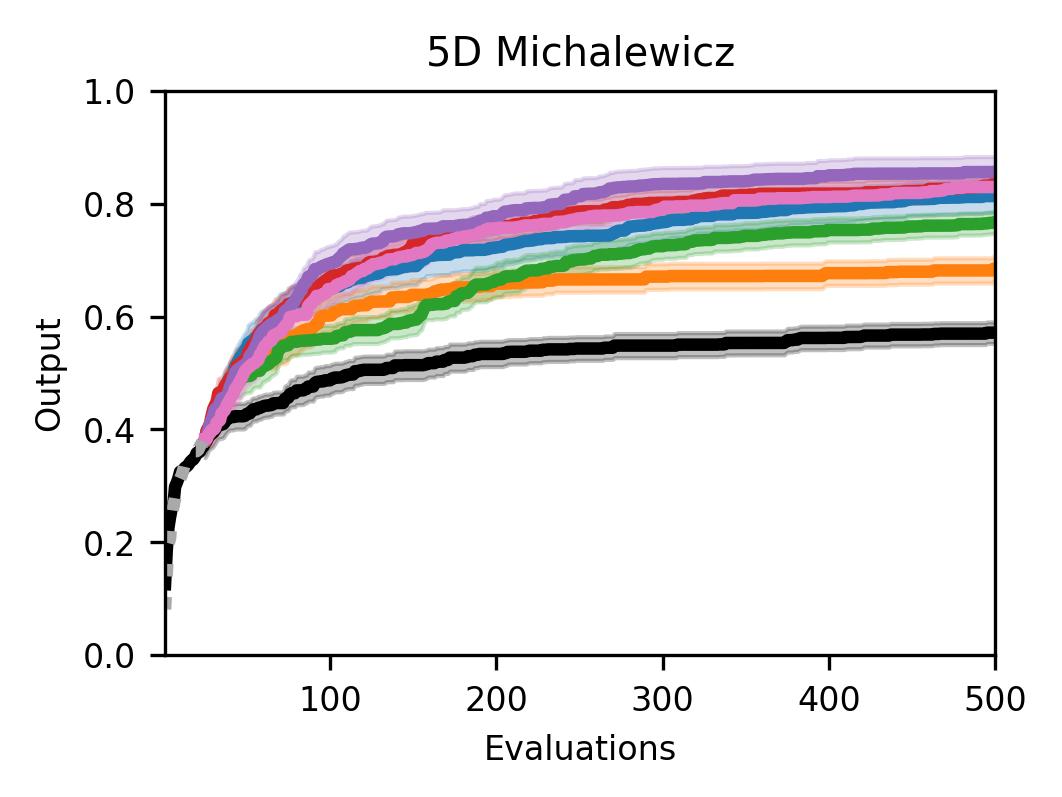}
    \end{minipage}

    \caption{Performance plots for Monte Carlo single-point acquisition functions  with five initial starting points per input dimension. Solid lines represent the mean over the 50 runs while the shaded areas represent the 95\% confidence intervals.}
\end{figure}

\begin{figure}
    \begin{minipage}[t]{0.02\textwidth}
        \vspace{-45mm}
        (A)
    \end{minipage}
    \begin{minipage}[t]{0.48\textwidth}
        \centering
        \includegraphics[width=0.8\linewidth]{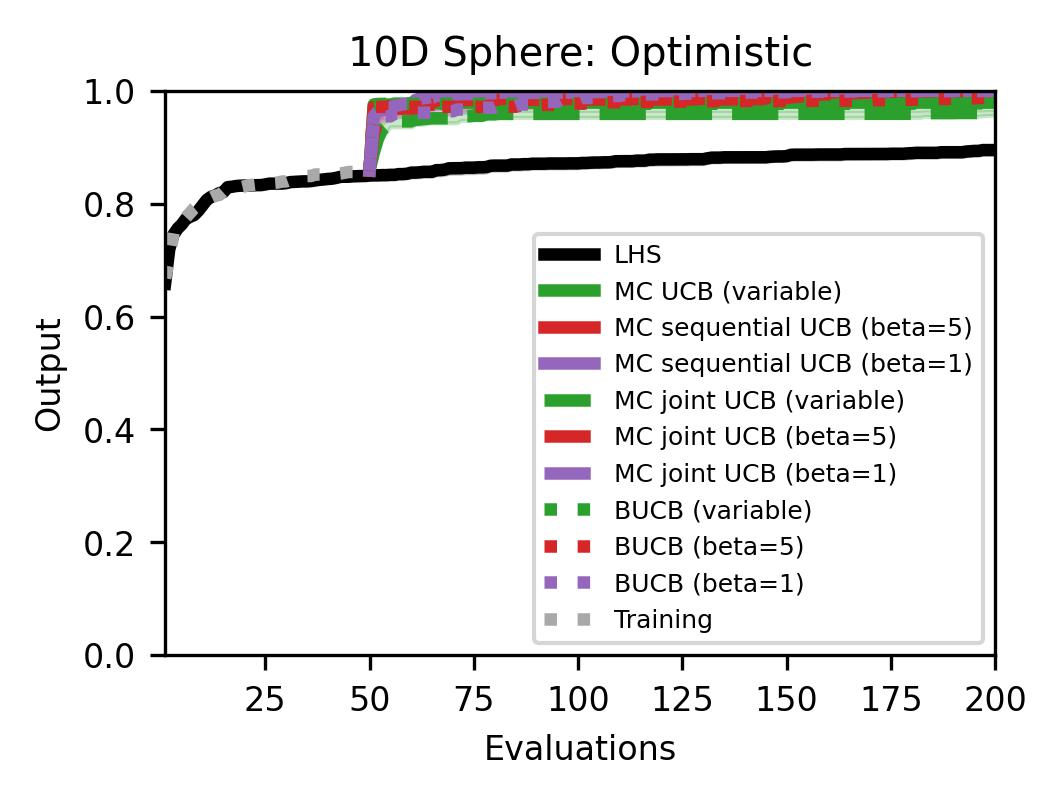}
    \end{minipage}
    \begin{minipage}[t]{0.02\textwidth}
        \vspace{-45mm}
        (B)
    \end{minipage}
    \begin{minipage}[t]{0.48\textwidth}
    \centering
        \includegraphics[width=0.8\linewidth]{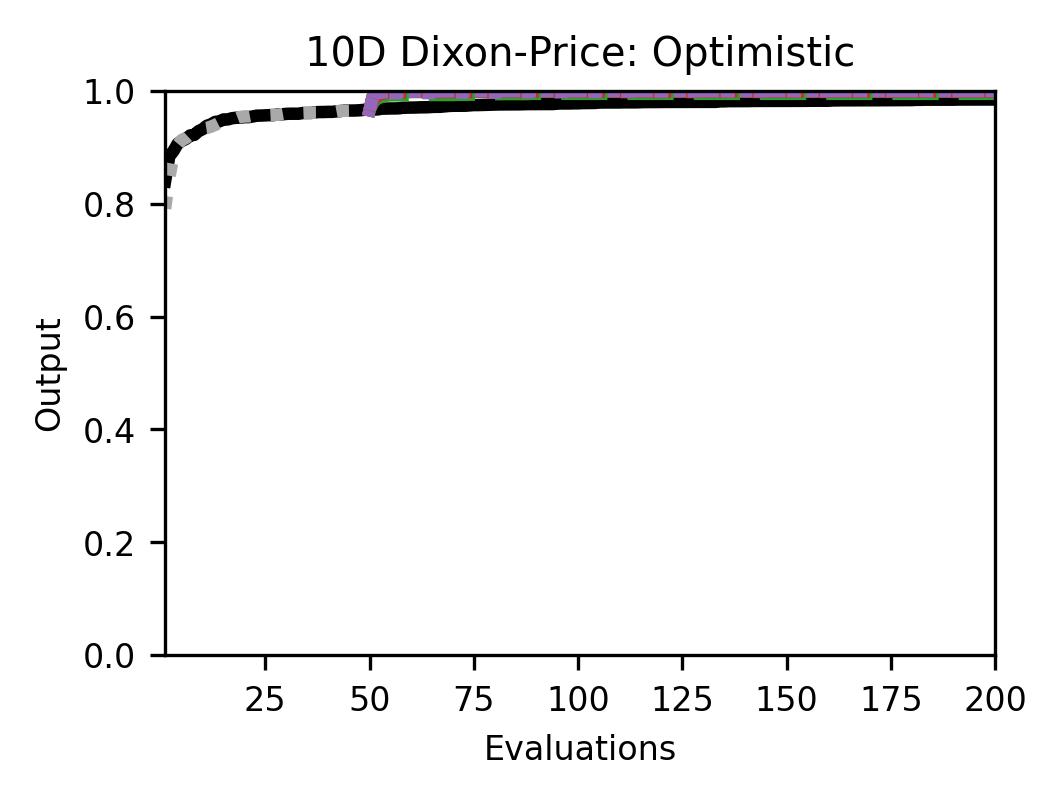}
    \end{minipage}
    \begin{minipage}[t]{0.02\textwidth}
        \vspace{-45mm}
        (C)
    \end{minipage}
    \begin{minipage}[t]{0.48\textwidth}
        \centering
        \includegraphics[width=0.8\linewidth]{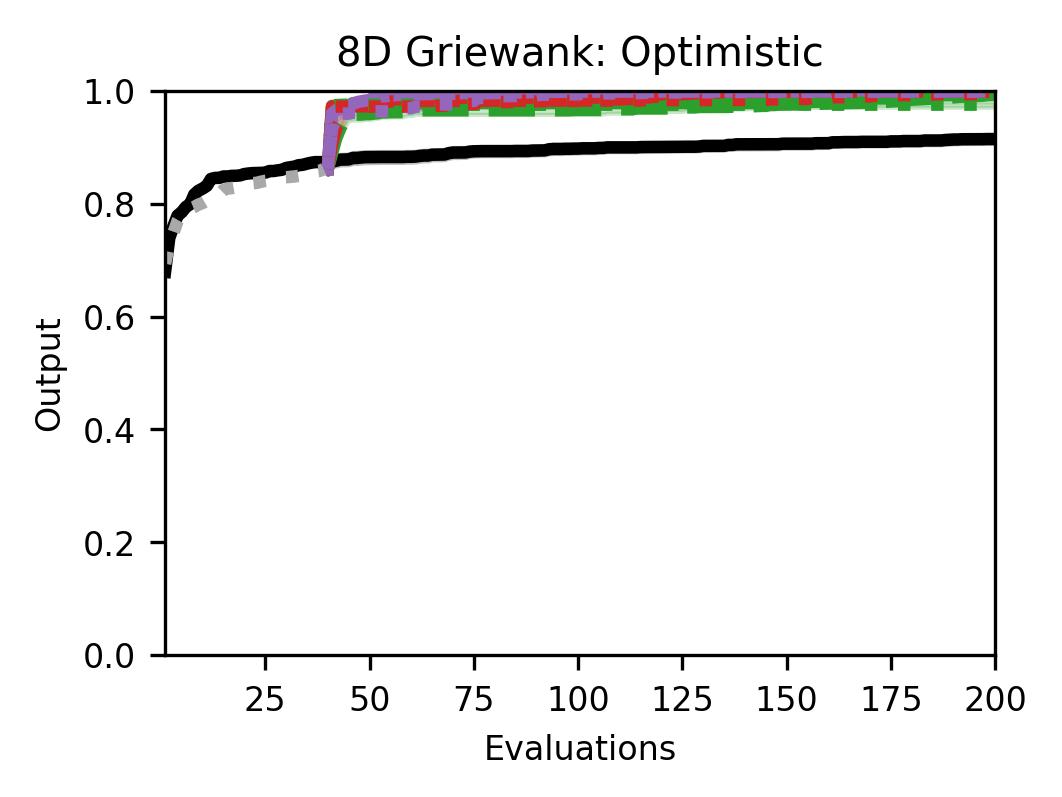}
    \end{minipage}
    \begin{minipage}[t]{0.02\textwidth}
        \vspace{-45mm}
        (D)
    \end{minipage}
    \begin{minipage}[t]{0.48\textwidth}
    \centering
        \includegraphics[width=0.8\linewidth]{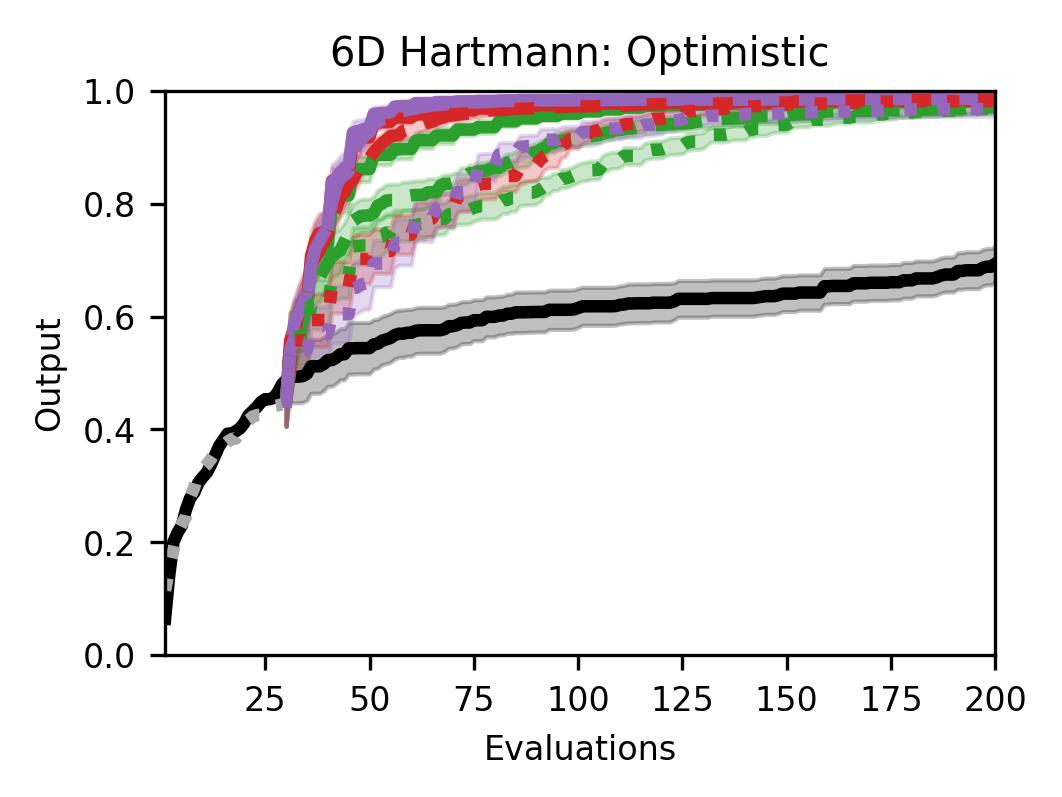}
    \end{minipage}
    \begin{minipage}[t]{0.02\textwidth}
        \vspace{-45mm}
        (E)
    \end{minipage}
    \begin{minipage}[t]{0.48\textwidth}
        \centering
        \includegraphics[width=0.8\linewidth]{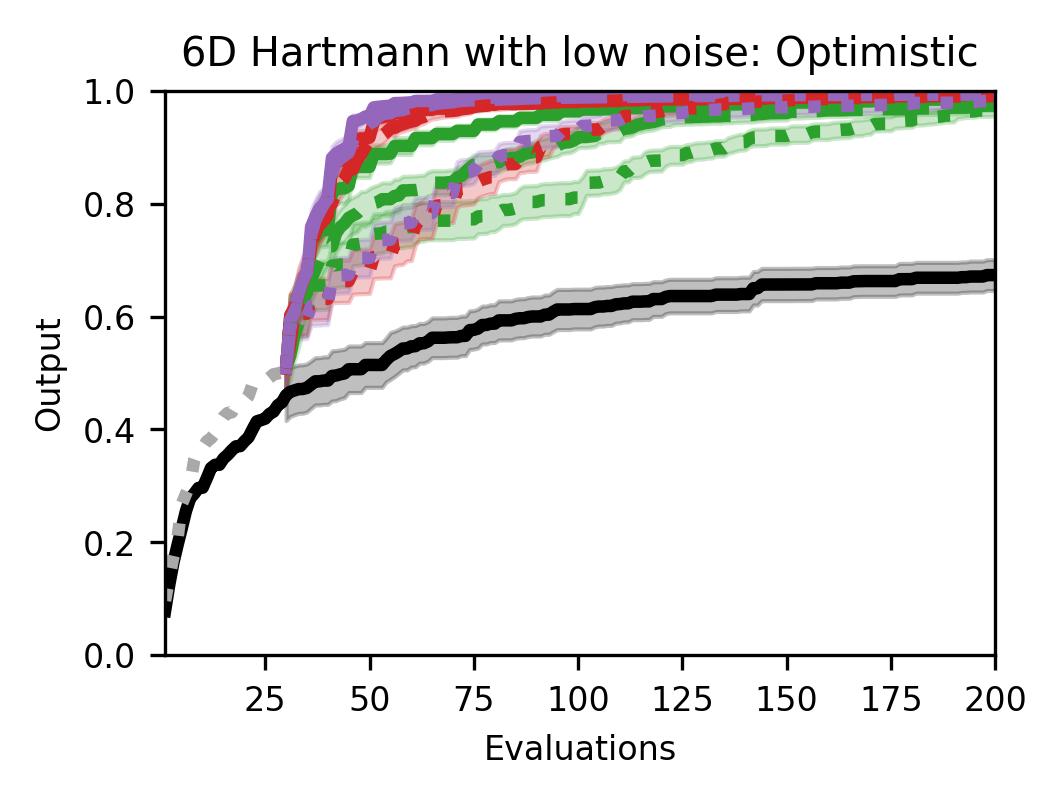}
    \end{minipage}
    \begin{minipage}[t]{0.02\textwidth}
        \vspace{-45mm}
        (F)
    \end{minipage}
    \begin{minipage}[t]{0.48\textwidth}
    \centering
        \includegraphics[width=0.8\linewidth]{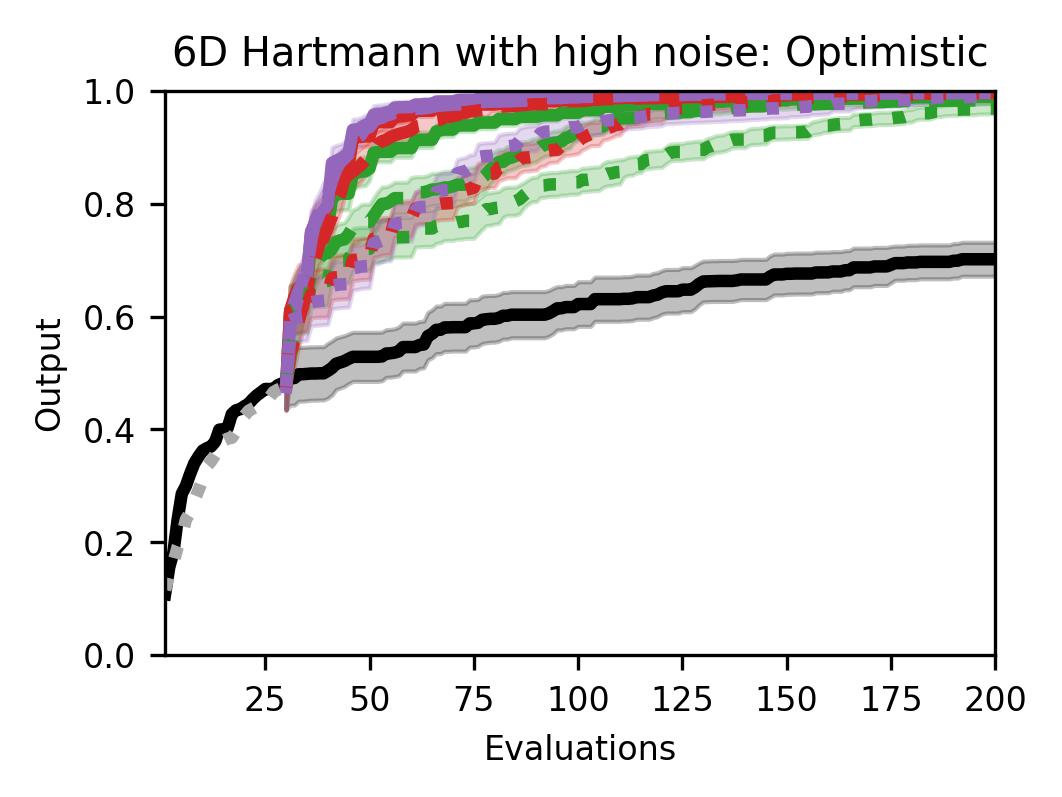}
    \end{minipage}
    \begin{minipage}[t]{0.02\textwidth}
        \vspace{-45mm}
        (G)
    \end{minipage}
    \begin{minipage}[t]{0.48\textwidth}
        \centering
        \includegraphics[width=0.8\linewidth]{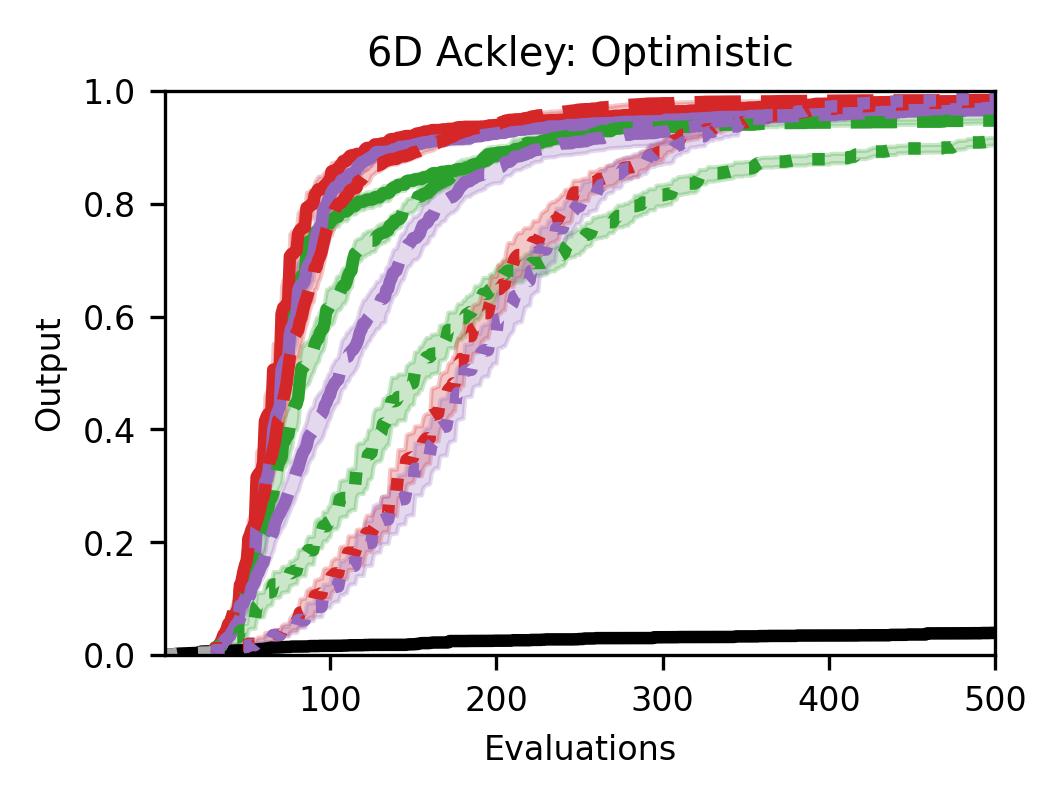}
    \end{minipage}
    \begin{minipage}[t]{0.02\textwidth}
        \vspace{-45mm}
        (H)
    \end{minipage}
    \begin{minipage}[t]{0.48\textwidth}
    \centering
        \includegraphics[width=0.8\linewidth]{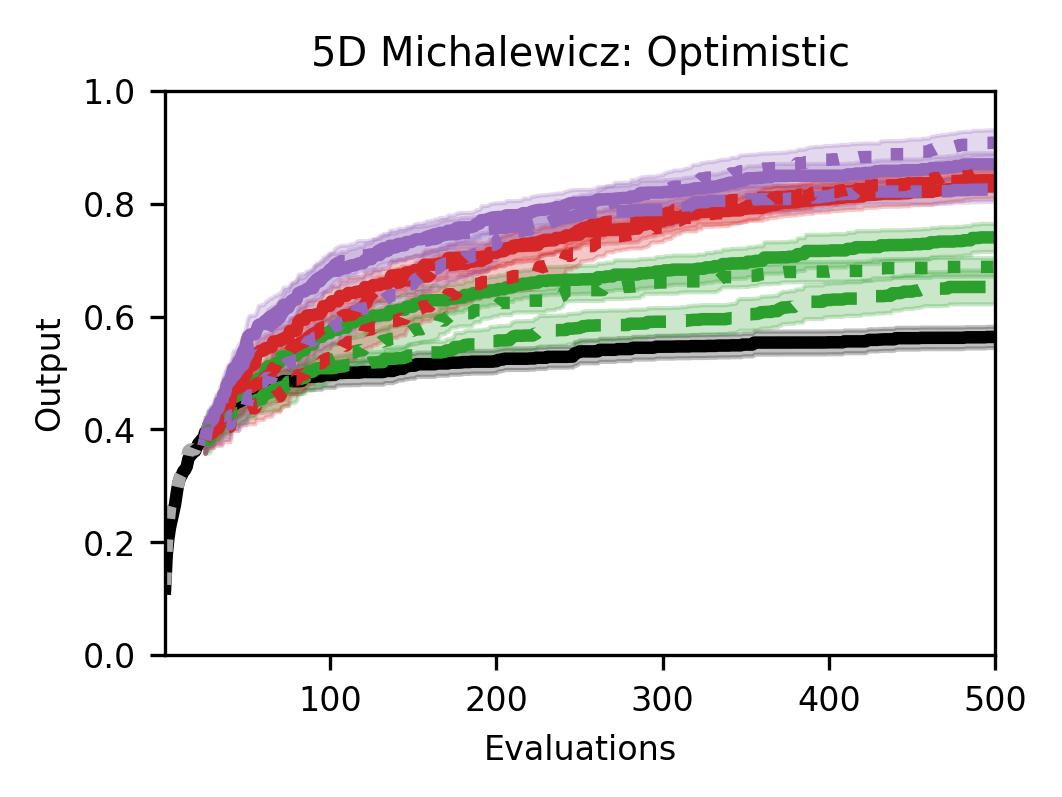}
    \end{minipage}

    \caption{Performance plots for optimistic multi-point acquisition functions with five initial training points per input dimension. Solid lines represent the mean over the 50 runs while the shaded area represents the 95\% confidence intervals.}
\end{figure}

\begin{figure}
    \begin{minipage}[t]{0.02\textwidth}
        \vspace{-45mm}
        (A)
    \end{minipage}
    \begin{minipage}[t]{0.48\textwidth}
        \centering
        \includegraphics[width=0.8\linewidth]{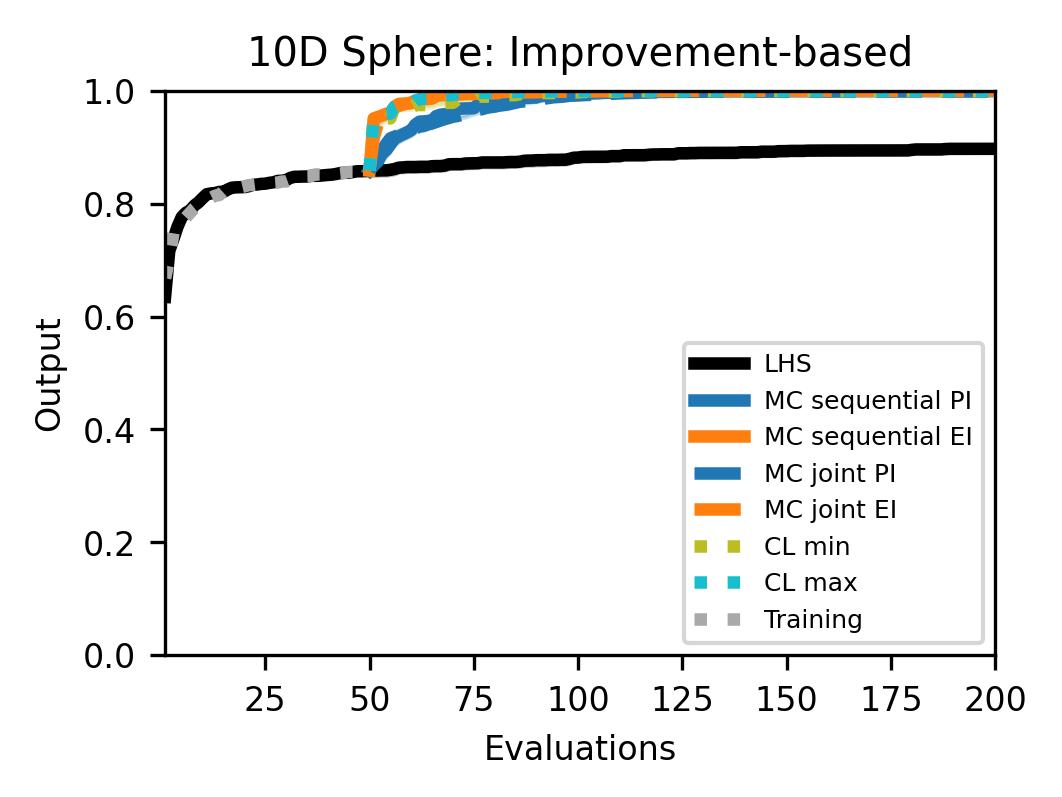}
    \end{minipage}
    \begin{minipage}[t]{0.02\textwidth}
        \vspace{-45mm}
        (B)
    \end{minipage}
    \begin{minipage}[t]{0.48\textwidth}
    \centering
        \includegraphics[width=0.8\linewidth]{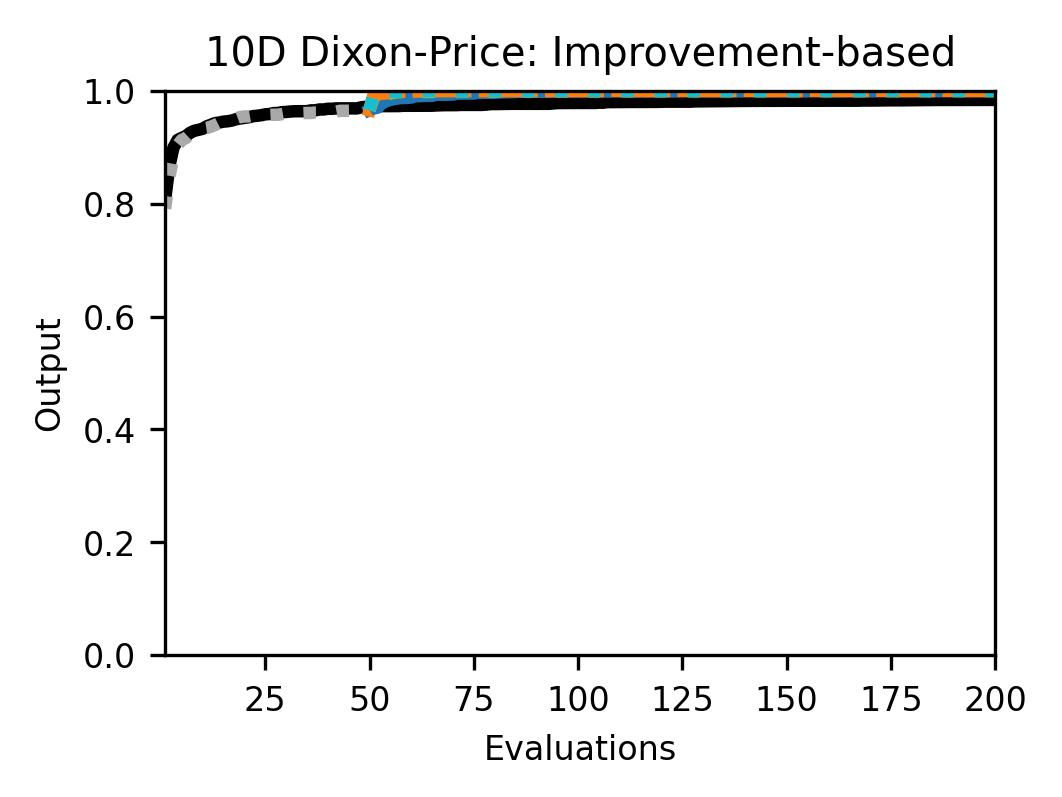}
    \end{minipage}
    \begin{minipage}[t]{0.02\textwidth}
        \vspace{-45mm}
        (C)
    \end{minipage}
    \begin{minipage}[t]{0.48\textwidth}
        \centering
        \includegraphics[width=0.8\linewidth]{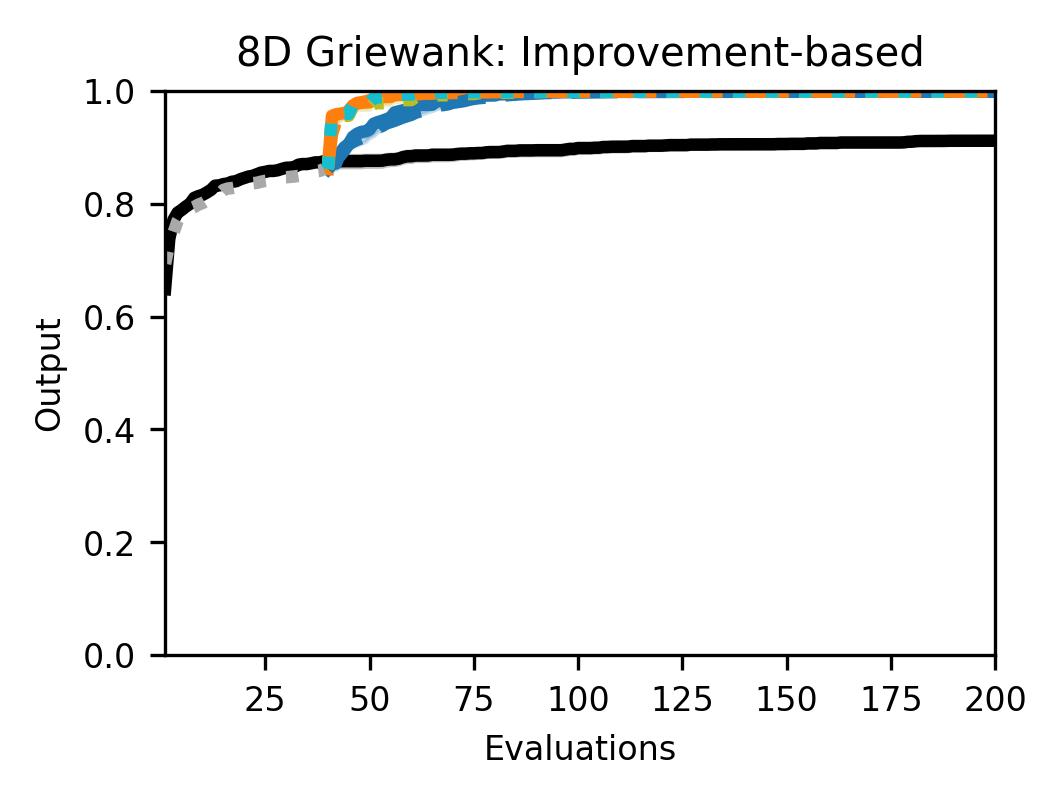}
    \end{minipage}
    \begin{minipage}[t]{0.02\textwidth}
        \vspace{-45mm}
        (D)
    \end{minipage}
    \begin{minipage}[t]{0.48\textwidth}
    \centering
        \includegraphics[width=0.8\linewidth]{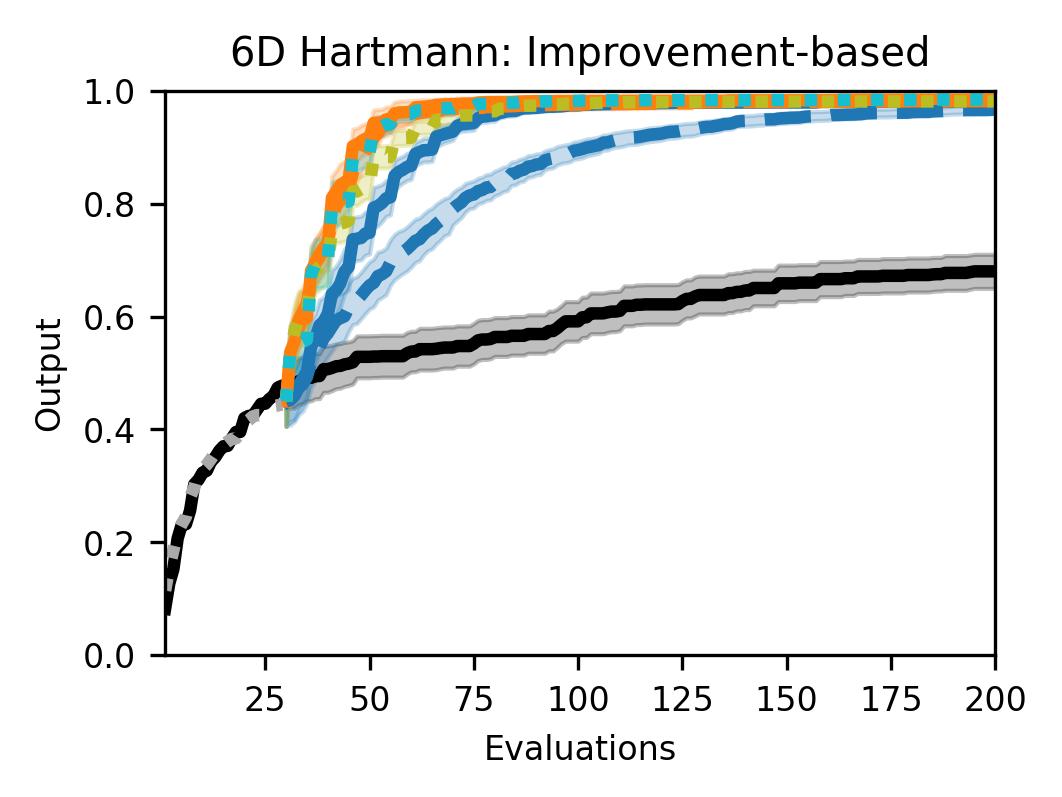}
    \end{minipage}
    \begin{minipage}[t]{0.02\textwidth}
        \vspace{-45mm}
        (E)
    \end{minipage}
    \begin{minipage}[t]{0.48\textwidth}
        \centering
        \includegraphics[width=0.8\linewidth]{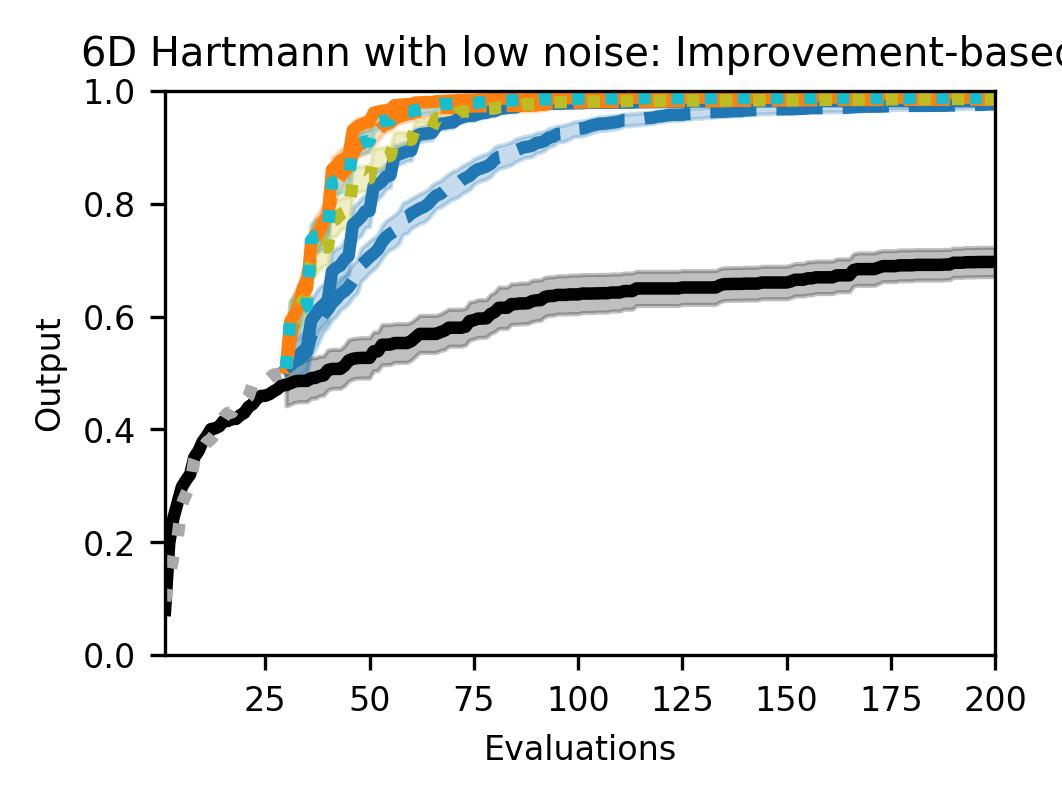}
    \end{minipage}
    \begin{minipage}[t]{0.02\textwidth}
        \vspace{-45mm}
        (F)
    \end{minipage}
    \begin{minipage}[t]{0.48\textwidth}
    \centering
        \includegraphics[width=0.8\linewidth]{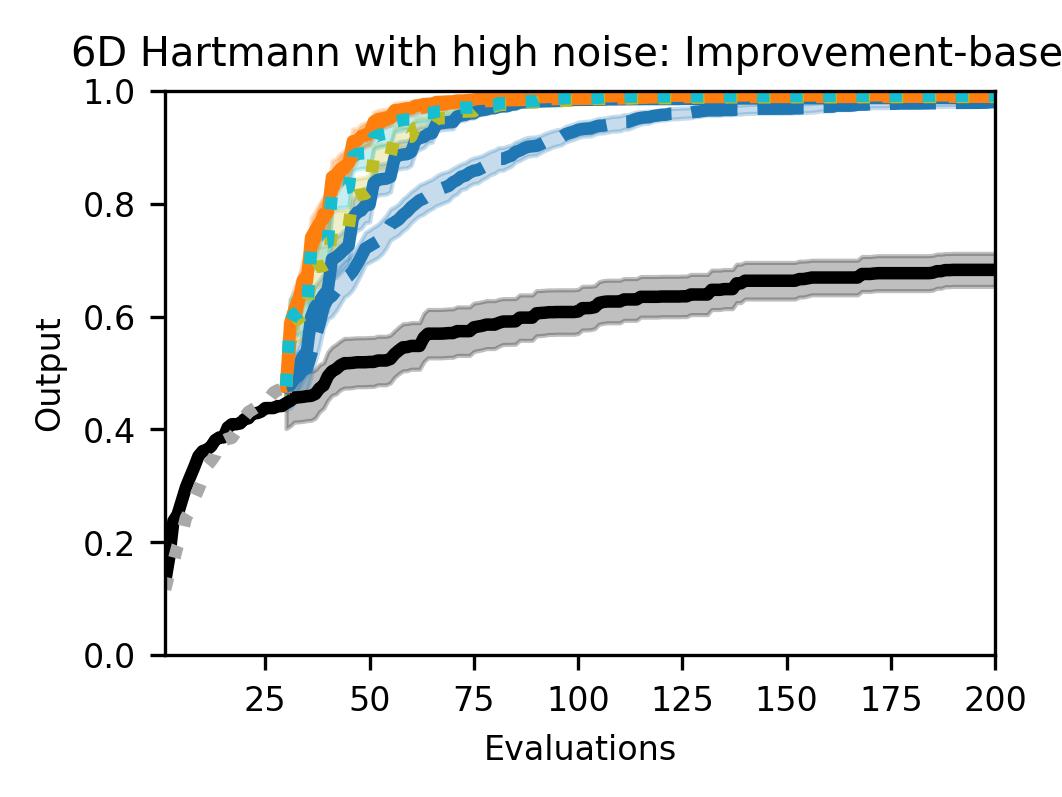}
    \end{minipage}
    \begin{minipage}[t]{0.02\textwidth}
        \vspace{-45mm}
        (G)
    \end{minipage}
    \begin{minipage}[t]{0.48\textwidth}
        \centering
        \includegraphics[width=0.8\linewidth]{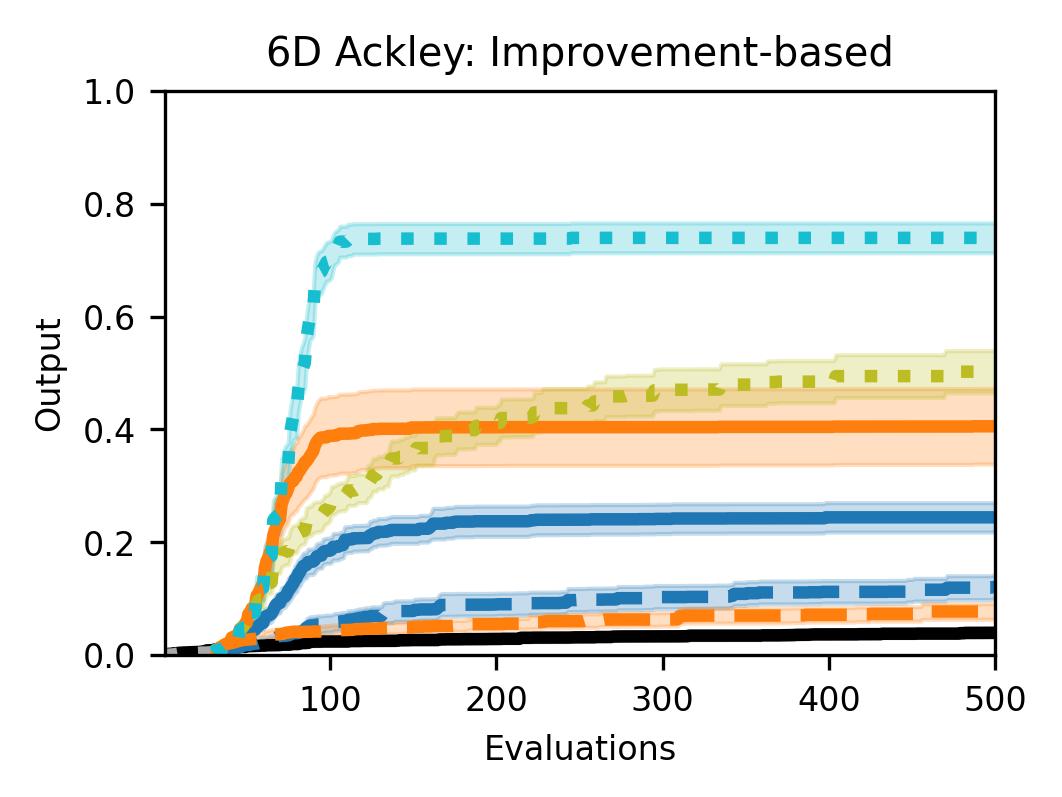}
    \end{minipage}
    \begin{minipage}[t]{0.02\textwidth}
        
        (H)
    \end{minipage}
    \begin{minipage}[t]{0.48\textwidth}
    \centering
        \includegraphics[width=0.8\linewidth]{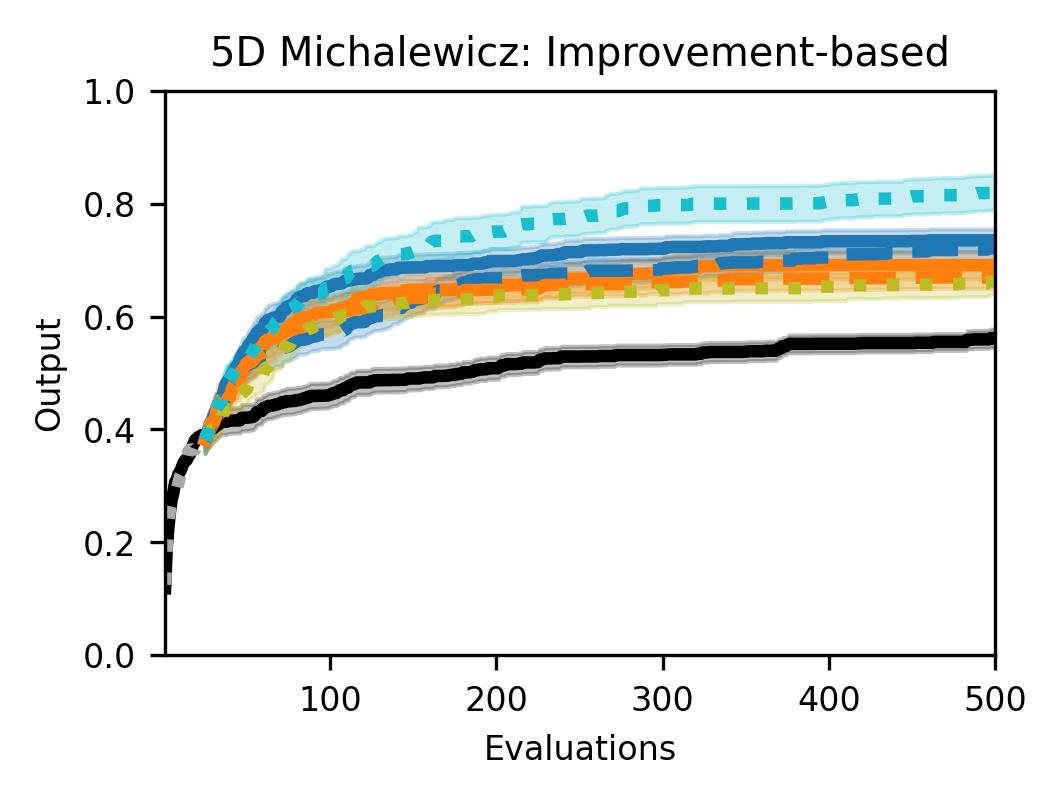}
    \end{minipage}

    \caption{Performance plots for improvement-based multi-point acquisition functions with five initial training points per input dimension. Solid lines represent the mean over the 50 runs while the shaded area represents the 95\% confidence intervals.}
\end{figure}

\FloatBarrier

\section{Supplementary Tables}

\begin{table}[h]
    \centering
    \small
    \caption{Best solutions found for analytical single-point acquisition functions with five initial training points per input dimension.}
    \begin{tabular}{ c | c c c c c c c c }
         Method & Sphere & Dixon-Price & Griewank & Hartmann & \makecell{Hartmann \\ low noise} & \makecell{Hartmann \\ high noise} & Michalewicz & Ackley \\
         \hline \hline
        & & & & & & & &\\[\dimexpr-\normalbaselineskip+5pt]
        PI & 1.00 & 1.00 & 1.00 & 0.97 & 0.99 & 1.00 & 0.78 & 0.88 \\[0.5cm]
        EI & 1.00 & 1.00 & 1.00 & 0.99 & 0.99 & 1.00 & 0.71 & 0.63 \\[0.5cm]
        \makecell{UCB \\ (variable)} & 0.98 & 1.00 & 1.00 & 0.98 & 0.99 & 1.00 & 0.80 & 0.96 \\[0.5cm]
        \makecell{UCB \\ ($\beta$=5)} & 1.00 & 1.00 & 1.00 & 0.99 & 1.00 & 1.00 & 0.85 & 0.99 \\[0.5cm]
        \makecell{UCB \\ ($\beta$=1)} & 1.00 & 1.00 & 1.00 & 0.97 & 1.00 & 1.00 & 0.89 & 0.97 \\[0.5cm]
        Hedge & 1.00 & 1.00 & 1.00 & 0.98 & 0.99 & 1.00 & 0.85 & 0.91
    \end{tabular}
    \vspace{20mm}
\end{table}

\begin{table}[h]
    \centering
    \small
    \caption{Averaged area under the curve with standard error for analytical single-point acquisition functions with five initial training points per input dimension.}
    \begin{tabular}{ c | c c c c c c c c }
         Method & Sphere & Dixon-Price & Griewank & Hartmann & \makecell{Hartmann \\ low noise} & \makecell{Hartmann \\ high noise} & Michalewicz & Ackley \\
         \hline \hline
         & & & & & & & &\\[\dimexpr-\normalbaselineskip+5pt]
        PI & \makecell{0.99 \\ ($\pm$ 0.00)} & \makecell{1.00 \\ ($\pm$ 0.00)} & \makecell{0.99 \\ ($\pm$ 0.00)} & \makecell{0.91 \\ ($\pm$ 0.04)} & \makecell{0.94 \\ ($\pm$ 0.02)} & \makecell{0.95 \\ ($\pm$ 0.02)} & \makecell{0.73 \\ ($\pm$ 0.10)} & \makecell{0.76 \\ ($\pm$ 0.08)} \\[0.5cm]
        EI & \makecell{1.00 \\ ($\pm$ 0.00)} & \makecell{1.00 \\ ($\pm$ 0.00)} & \makecell{1.00 \\ ($\pm$ 0.00)} & \makecell{0.97 \\ ($\pm$ 0.02)} & \makecell{0.97 \\ ($\pm$ 0.02)} & \makecell{0.97 \\ ($\pm$ 0.02)} & \makecell{0.68 \\ ($\pm$ 0.08)} & \makecell{0.59 \\ ($\pm$ 0.15)} \\[0.5cm]
        \makecell{UCB \\ (variable)} & \makecell{0.98 \\ ($\pm$ 0.01)} & \makecell{1.00 \\ ($\pm$ 0.00)} & \makecell{0.98 \\ ($\pm$ 0.01)} & \makecell{0.94 \\ ($\pm$ 0.03)} & \makecell{0.95 \\ ($\pm$ 0.02)} & \makecell{0.95 \\ ($\pm$ 0.03)} & \makecell{0.69 \\ ($\pm$ 0.06)} & \makecell{0.85 \\ ($\pm$ 0.01)} \\[0.5cm]
        \makecell{UCB \\ ($\beta$=5)} & \makecell{1.00 \\ ($\pm$ 0.00)} & \makecell{1.00 \\ ($\pm$ 0.00)} & \makecell{1.00 \\ ($\pm$ 0.00)} & \makecell{0.97 \\ ($\pm$ 0.02)} & \makecell{0.97 \\ ($\pm$ 0.02)} & \makecell{0.98 \\ ($\pm$ 0.02)} & \makecell{0.77 \\ ($\pm$ 0.08)} & \makecell{0.90 \\ ($\pm$ 0.02)} \\[0.5cm]
        \makecell{UCB \\ ($\beta$=1)} & \makecell{1.00 \\ ($\pm$ 0.00)} & \makecell{1.00 \\ ($\pm$ 0.00)} & \makecell{1.00 \\ ($\pm$ 0.00)} & \makecell{0.95 \\ ($\pm$ 0.04)} & \makecell{0.97 \\ ($\pm$ 0.03)} & \makecell{0.97 \\ ($\pm$ 0.02)} & \makecell{0.80 \\ ($\pm$ 0.08)} & \makecell{0.88 \\ ($\pm$ 0.05)} \\[0.5cm]
        Hedge & \makecell{1.00 \\ ($\pm$ 0.00)} & \makecell{1.00 \\ ($\pm$ 0.00)} & \makecell{1.00 \\ ($\pm$ 0.00)} & \makecell{0.95 \\ ($\pm$ 0.03)} & \makecell{0.96 \\ ($\pm$ 0.03)} & \makecell{0.97 \\ ($\pm$ 0.02)} & \makecell{0.76 \\ ($\pm$ 0.08)} & \makecell{0.77 \\ ($\pm$ 0.11)}
        \end{tabular}
        \vspace{20mm}
\end{table}

\begin{table}[h]
    \centering
    \small
    \caption{Best solutions found for analytical single-point acquisition functions with one initial training point per input dimension.}
    \begin{tabular}{ c | c c c c c c c c }
         Method & Sphere & Dixon-Price & Griewank & Hartmann & \makecell{Hartmann \\ low noise} & \makecell{Hartmann \\ high noise} & Michalewicz & Ackley \\
         \hline \hline
         & & & & & & & &\\[\dimexpr-\normalbaselineskip+5pt]
        PI & 1.00 & 1.00 & 1.00 & 0.98 & 0.99 & 1.00 & 0.78 & 0.88 \\[0.5cm]
        EI & 1.00 & 1.00 & 1.00 & 0.99 & 0.99 & 0.99 & 0.66 & 0.63 \\[0.5cm]
        \makecell{UCB \\ (variable)} & 0.98 & 1.00 & 1.00 & 0.99 & 1.00 & 1.00 & 0.79 & 0.96 \\[0.5cm]
        \makecell{UCB \\ ($\beta$=5)} & 1.00 & 1.00 & 1.00 & 0.99 & 1.00 & 1.00 & 0.81 & 0.99 \\[0.5cm]
        \makecell{UCB \\ ($\beta$=1)} & 1.00 & 1.00 & 1.00 & 0.97 & 0.98 & 0.99 & 0.89 & 0.96 \\[0.5cm]
        Hedge & 1.00 & 1.00 & 1.00 & 0.96 & 1.00 & 0.99 & 0.84 & 0.94
        \end{tabular}
        \vspace{20mm}
\end{table}

\begin{table}[h]
\centering
    \small
    \caption{Averaged area under the curve with standard error for analytical single-point acquisition functions with one initial training point per input dimension.}
    \begin{tabular}{ c | c c c c c c c c }
         Method & Sphere & Dixon-Price & Griewank & Hartmann & \makecell{Hartmann \\ low noise} & \makecell{Hartmann \\ high noise} & Michalewicz & Ackley \\
         \hline \hline
         & & & & & & & &\\[\dimexpr-\normalbaselineskip+5pt]
        PI & \makecell{0.97 \\ ($\pm$ 0.01)} & \makecell{0.99 \\ ($\pm$ 0.00)} & \makecell{0.97 \\ ($\pm$ 0.01)} & \makecell{0.82 \\ ($\pm$ 0.09)} & \makecell{0.87 \\ ($\pm$ 0.07)} & \makecell{0.90 \\ ($\pm$ 0.06)} & \makecell{0.68 \\ ($\pm$ 0.07)} & \makecell{0.73 \\ ($\pm$ 0.11)} \\[0.5cm]
        EI & \makecell{0.99 \\ ($\pm$ 0.00)} & \makecell{1.00 \\ ($\pm$ 0.00)} & \makecell{0.99 \\ ($\pm$ 0.00)} & \makecell{0.95 \\ ($\pm$ 0.03)} & \makecell{0.94 \\ ($\pm$ 0.04)} & \makecell{0.95 \\ ($\pm$ 0.03)} & \makecell{0.62 \\ ($\pm$ 0.09)} & \makecell{0.58 \\ ($\pm$ 0.13)} \\[0.5cm]
        \makecell{UCB \\ (variable)} & \makecell{0.96 \\ ($\pm$ 0.01)} & \makecell{1.00 \\ ($\pm$ 0.00)} & \makecell{0.97 \\ ($\pm$ 0.01)} & \makecell{0.91 \\ ($\pm$ 0.04)} & \makecell{0.90 \\ ($\pm$ 0.05)} & \makecell{0.91 \\ ($\pm$ 0.03)} & \makecell{0.66 \\ ($\pm$ 0.06)} & \makecell{0.82 \\ ($\pm$ 0.01)} \\[0.5cm]
        \makecell{UCB \\ ($\beta$=5)} & \makecell{0.98 \\ ($\pm$ 0.00)} & \makecell{1.00 \\ ($\pm$ 0.00)} & \makecell{0.98 \\ ($\pm$ 0.00)} & \makecell{0.94 \\ ($\pm$ 0.03)} & \makecell{0.94 \\ ($\pm$ 0.04)} & \makecell{0.95 \\ ($\pm$ 0.03)} & \makecell{0.72 \\ ($\pm$ 0.05)} & \makecell{0.87 \\ ($\pm$ 0.02)} \\[0.5cm]
        \makecell{UCB \\ ($\beta$=1)} & \makecell{0.99 \\ ($\pm$ 0.00)} & \makecell{1.00 \\ ($\pm$ 0.00)} & \makecell{0.99 \\ ($\pm$ 0.00)} & \makecell{0.93 \\ ($\pm$ 0.08)} & \makecell{0.92 \\ ($\pm$ 0.10)} & \makecell{0.93 \\ ($\pm$ 0.09)} & \makecell{0.77 \\ ($\pm$ 0.07)} & \makecell{0.85 \\ ($\pm$ 0.06)} \\[0.5cm]
        Hedge & \makecell{0.98 \\ ($\pm$ 0.01)} & \makecell{0.99 \\ ($\pm$ 0.00)} & \makecell{0.98 \\ ($\pm$ 0.01)} & \makecell{0.89 \\ ($\pm$ 0.11)} & \makecell{0.91 \\ ($\pm$ 0.05)} & \makecell{0.92 \\ ($\pm$ 0.05)} & \makecell{0.73 \\ ($\pm$ 0.08)} & \makecell{0.76 \\ ($\pm$ 0.09)}
        \end{tabular}
        \vspace{20mm}
\end{table}

\begin{table}[h]
    \centering
    \small
    \caption{Best solutions found for analytical single-point acquisition functions with ten initial training points per input dimension.}
    \begin{tabular}{ c | c c c c c c c c }
         Method & Sphere & Dixon-Price & Griewank & Hartmann & \makecell{Hartmann \\ low noise} & \makecell{Hartmann \\ high noise} & Michalewicz & Ackley \\
         \hline \hline
         & & & & & & & &\\[\dimexpr-\normalbaselineskip+5pt]
        PI & 1.00 & 1.00 & 1.00 & 0.97 & 0.99 & 0.99 & 0.73 & 0.51 \\[0.5cm]
        EI & 1.00 & 1.00 & 1.00 & 0.98 & 0.99 & 0.99 & 0.68 & 0.53 \\[0.5cm]
        \makecell{UCB \\ (variable)} & 0.99 & 1.00 & 1.00 & 0.98 & 0.98 & 0.99 & 0.77 & 0.96 \\[0.5cm]
        \makecell{UCB \\ ($\beta$=5)} & 1.00 & 1.00 & 1.00 & 0.98 & 0.99 & 1.00 & 0.81 & 0.98 \\[0.5cm]
        \makecell{UCB \\ ($\beta$=1)} & 1.00 & 1.00 & 1.00 & 0.98 & 0.99 & 1.00 & 0.88 & 0.95 \\[0.5cm]
        Hedge & 1.00 & 1.00 & 1.00 & 0.98 & 0.99 & 1.00 & 0.79 & 0.87
    \end{tabular}
    \vspace{20mm}
\end{table}

\begin{table}[h]
    \centering
    \small
    \caption{Averaged area under the curve with standard error for analytical single-point acquisition functions with ten initial training points per input dimension.}
    \begin{tabular}{ c | c c c c c c c c }
         Method & Sphere & Dixon-Price & Griewank & Hartmann & \makecell{Hartmann \\ low noise} & \makecell{Hartmann \\ high noise} & Michalewicz & Ackley \\
         \hline \hline
         & & & & & & & &\\[\dimexpr-\normalbaselineskip+5pt]
        PI & \makecell{1.00 \\ ($\pm$ 0.00)} & \makecell{1.00 \\ ($\pm$ 0.00)} & \makecell{1.00 \\ ($\pm$ 0.00)} & \makecell{0.93 \\ ($\pm$ 0.05)} & \makecell{0.96 \\ ($\pm$ 0.02)} & \makecell{0.96 \\ ($\pm$ 0.02)} & \makecell{0.69 \\ ($\pm$ 0.10)} & \makecell{0.43 \\ ($\pm$ 0.25)} \\[0.5cm]
        EI & \makecell{1.00 \\ ($\pm$ 0.00)} & \makecell{1.00 \\ ($\pm$ 0.00)} & \makecell{1.00 \\ ($\pm$ 0.00)} & \makecell{0.97 \\ ($\pm$ 0.02)} & \makecell{0.97 \\ ($\pm$ 0.02)} & \makecell{0.97 \\ ($\pm$ 0.02)} & \makecell{0.64 \\ ($\pm$ 0.09)} & \makecell{0.41 \\ ($\pm$ 0.07)} \\[0.5cm]
        \makecell{UCB \\ (variable)} & \makecell{0.99 \\ ($\pm$ 0.01)} & \makecell{1.00 \\ ($\pm$ 0.00)} & \makecell{1.00 \\ ($\pm$ 0.00)} & \makecell{0.95 \\ ($\pm$ 0.03)} & \makecell{0.96 \\ ($\pm$ 0.03)} & \makecell{0.96 \\ ($\pm$ 0.03)} & \makecell{0.66 \\ ($\pm$ 0.07)} & \makecell{0.87 \\ ($\pm$ 0.01)} \\[0.5cm]
        \makecell{UCB \\ ($\beta$=5)} & \makecell{1.00 \\ ($\pm$ 0.00)} & \makecell{1.00 \\ ($\pm$ 0.00)} & \makecell{1.00 \\ ($\pm$ 0.00)} & \makecell{0.97 \\ ($\pm$ 0.02)} & \makecell{0.98 \\ ($\pm$ 0.02)} & \makecell{0.98 \\ ($\pm$ 0.02)} & \makecell{0.74 \\ ($\pm$ 0.06)} & \makecell{0.91 \\ ($\pm$ 0.03)} \\[0.5cm]
        \makecell{UCB \\ ($\beta$=1)} & \makecell{1.00 \\ ($\pm$ 0.00)} & \makecell{1.00 \\ ($\pm$ 0.00)} & \makecell{1.00 \\ ($\pm$ 0.00)} & \makecell{0.96 \\ ($\pm$ 0.03)} & \makecell{0.97 \\ ($\pm$ 0.02)} & \makecell{0.98 \\ ($\pm$ 0.03)} & \makecell{0.77 \\ ($\pm$ 0.07)} & \makecell{0.86 \\ ($\pm$ 0.08)} \\[0.5cm]
        Hedge & \makecell{1.00 \\ ($\pm$ 0.00)} & \makecell{1.00 \\ ($\pm$ 0.00)} & \makecell{1.00 \\ ($\pm$ 0.00)} & \makecell{0.96 \\ ($\pm$ 0.03)} & \makecell{0.97 \\ ($\pm$ 0.02)} & \makecell{0.97 \\ ($\pm$ 0.02)} & \makecell{0.71 \\ ($\pm$ 0.09)} & \makecell{0.78 \\ ($\pm$ 0.13)}
        \end{tabular}
        \vspace{20mm}
\end{table}

\begin{table}[h]
    \centering
    \small
    \caption{Best solutions found for Monte Carlo single-point acquisition functions with five initial training points per input dimension.}
    \begin{tabular}{ c | c c c c c c c c }
         Method & Sphere & Dixon-Price & Griewank & Hartmann & \makecell{Hartmann \\ low noise} & \makecell{Hartmann \\ high noise} & Michalewicz & Ackley \\
         \hline \hline
         & & & & & & & &\\[\dimexpr-\normalbaselineskip+5pt]
        MC PI & 1.00 & 1.00 & 1.00 & 0.98 & 0.99 & 0.99 & 0.81 & 0.20 \\[0.5cm]
        MC EI & 1.00 & 1.00 & 1.00 & 0.99 & 0.99 & 0.99 & 0.68 & 0.07 \\[0.5cm]
        \makecell{MC UCB \\ (variable)} & 0.98 & 1.00 & 1.00 & 0.98 & 0.99 & 0.99 & 0.77 & 0.96 \\[0.5cm]
        \makecell{MC UCB \\ ($\beta$=5)} & 1.00 & 1.00 & 1.00 & 0.99 & 1.00 & 1.00 & 0.84 & 0.99 \\[0.5cm]
        \makecell{MC UCB \\ ($\beta$=1)} & 1.00 & 1.00 & 1.00 & 0.97 & 1.00 & 1.00 & 0.86 & 0.97 \\[0.5cm]
        MES & 1.00 & 1.00 & 1.00 & 0.99 & 0.99 & 1.00 & 0.83 & 0.55 
    \end{tabular}
    \vspace{20mm}
\end{table}

\begin{table}[h]
    \centering
    \small
    \caption{Averaged area under the curve with standard error for Monte Carlo single-point acquisition functions with five initial training points per input dimension.}
    \begin{tabular}{ c | c c c c c c c c }
        Method & Sphere & Dixon-Price & Griewank & Hartmann & \makecell{Hartmann \\ low noise} & \makecell{Hartmann \\ high noise} & Michalewicz & Ackley \\
        \hline \hline
        & & & & & & & &\\[\dimexpr-\normalbaselineskip+5pt]
        MC PI & \makecell{0.99 \\ ($\pm$ 0.00)} & \makecell{1.00 \\ ($\pm$ 0.00)} & \makecell{0.99 \\ ($\pm$ 0.00)} & \makecell{0.92 \\ ($\pm$ 0.06)} & \makecell{0.94 \\ ($\pm$ 0.02)} & \makecell{0.95 \\ ($\pm$ 0.04)} & \makecell{0.73 \\ ($\pm$ 0.09)} & \makecell{0.17 \\ ($\pm$ 0.12)} \\[0.5cm]
        MC EI & \makecell{1.00 \\ ($\pm$ 0.00)} & \makecell{1.00 \\ ($\pm$ 0.00)} & \makecell{1.00 \\ ($\pm$ 0.00)} & \makecell{0.97 \\ ($\pm$ 0.02)} & \makecell{0.97 \\ ($\pm$ 0.02)} & \makecell{0.97 \\ ($\pm$ 0.04)} & \makecell{0.64 \\ ($\pm$ 0.08)} & \makecell{0.05 \\ ($\pm$ 0.03)} \\[0.5cm]
        \makecell{MC UCB \\ (variable)} & \makecell{0.98 \\ ($\pm$ 0.01)} & \makecell{1.00 \\ ($\pm$ 0.00)} & \makecell{0.98 \\ ($\pm$ 0.01)} & \makecell{0.95 \\ ($\pm$ 0.02)} & \makecell{0.95 \\ ($\pm$ 0.02)} & \makecell{0.95 \\ ($\pm$ 0.04)} & \makecell{0.67 \\ ($\pm$ 0.06)} & \makecell{0.86 \\ ($\pm$ 0.01)} \\[0.5cm]
        \makecell{MC UCB \\ ($\beta$=5)} & \makecell{1.00 \\ ($\pm$ 0.00)} & \makecell{1.00 \\ ($\pm$ 0.00)} & \makecell{1.00 \\ ($\pm$ 0.00)} & \makecell{0.97 \\ ($\pm$ 0.02)} & \makecell{0.98 \\ ($\pm$ 0.02)} & \makecell{0.98 \\ ($\pm$ 0.03)} & \makecell{0.75 \\ ($\pm$ 0.06)} & \makecell{0.91 \\ ($\pm$ 0.01)} \\[0.5cm]
        \makecell{MC UCB \\ ($\beta$=1)} & \makecell{1.00 \\ ($\pm$ 0.00)} & \makecell{1.00 \\ ($\pm$ 0.00)} & \makecell{1.00 \\ ($\pm$ 0.00)} & \makecell{0.96 \\ ($\pm$ 0.07)} & \makecell{0.97 \\ ($\pm$ 0.03)} & \makecell{0.97 \\ ($\pm$ 0.07)} & \makecell{0.77 \\ ($\pm$ 0.08)} & \makecell{0.89 \\ ($\pm$ 0.03)} \\[0.5cm]
        MES & \makecell{0.99 \\ ($\pm$ 0.00)} & \makecell{0.99 \\ ($\pm$ 0.00)} & \makecell{0.99 \\ ($\pm$ 0.00)} & \makecell{0.97 \\ ($\pm$ 0.02)} & \makecell{0.97 \\ ($\pm$ 0.02)} & \makecell{0.97 \\ ($\pm$ 0.03)} & \makecell{0.74 \\ ($\pm$ 0.05)} & \makecell{0.49 \\ ($\pm$ 0.12)}
    \end{tabular}
\end{table}

\begin{table}[h]
    \centering
    \caption{Best solutions found for multi-point acquisition functions with five initial training points per input dimension.}
    \small
    \begin{tabular}{ c c | c c c c c c c c }
        Type & Method & Sphere & Dixon-Price & Griewank & Hartmann & \makecell{Hartmann \\ low noise} & \makecell{Hartmann \\ high noise} & Michalewicz & Ackley \\
        \hline \hline
        & & & & & & & &\\[\dimexpr-\normalbaselineskip+5pt]
        \parbox{0mm}{\multirow{5}{10pt}{\rotatebox{90}{Sequential Monte Carlo\kern 20pt}}} & PI & 1.00 & 1.00 & 1.00 & 0.98 & 0.98 & 0.99 & 0.74 & 0.24 \\[0.3cm]
        & EI & 1.00 & 1.00 & 1.00 & 0.98 & 0.99 & 0.99 & 0.69 & 0.41 \\[0.3cm]
        & \makecell{UCB \\ (variable)} & 0.98 & 1.00 & 0.99 & 0.98 & 0.98 & 0.99 & 0.74 & 0.97 \\[0.5cm]
        & \makecell{UCB \\ ($\beta$=5)} & 1.00 & 1.00 & 1.00 & 0.98 & 0.99 & 1.00 & 0.83 & 0.98 \\[0.5cm]
        & \makecell{UCB \\ ($\beta$=1)} & 1.00 & 1.00 & 1.00 & 0.98 & 0.99 & 1.00 & 0.87 & 0.97 \\[0.5cm]
        \hline
        & & & & & & & &\\[\dimexpr-\normalbaselineskip+5pt]
         \parbox{0mm}{\multirow{5}{10pt}{\rotatebox{90}{Joint Monte Carlo\kern 30pt}}} & PI & 1.00 & 1.00 & 1.00 & 0.97 & 0.98 & 0.98 & 0.73 & 0.12 \\[0.3cm]
        & EI & 1.00 & 1.00 & 1.00 & 0.98 & 0.98 & 0.99 & 0.67 & 0.08 \\[0.3cm]
        & \makecell{UCB \\ (variable)} & 0.96 & 0.99 & 0.99 & 0.97 & 0.97 & 0.98 & 0.65 & 0.95 \\[0.5cm]
        & \makecell{UCB \\ ($\beta$=5)} & 1.00 & 1.00 & 1.00 & 0.98 & 0.99 & 1.00 & 0.84 & 0.98 \\[0.5cm]
        & \makecell{UCB \\ ($\beta$=1)} & 1.00 & 1.00 & 1.00 & 0.99 & 0.99 & 1.00 & 0.83 & 0.97 \\[0.5cm]
        \hline
        & & & & & & & &\\[\dimexpr-\normalbaselineskip+5pt]
        \parbox{0mm}{\multirow{5}{10pt}{\rotatebox{90}{Analytical\kern 40pt}}} & CL min & 1.00 & 1.00 & 1.00 & 0.98 & 0.99 & 0.99 & 0.66 & 0.50 \\[0.3cm]
        & CL max & 1.00 & 1.00 & 1.00 & 0.98 & 0.99 & 0.99 & 0.82 & 0.74 \\[0.3cm]
        & \makecell{BUCB \\ (variable)} & 0.98 & 1.00 & 0.98 & 0.97 & 0.97 & 0.97 & 0.69 & 0.91 \\[0.5cm]
        & \makecell{BUCB \\ ($\beta$=5)} & 0.99 & 1.00 & 0.99 & 0.98 & 0.99 & 1.00 & 0.85 & 0.98 \\[0.5cm]
        & \makecell{BUCB \\ ($\beta$=1)} & 1.00 & 1.00 & 1.00 & 0.97 & 0.98 & 0.99 & 0.91 & 0.99
    \end{tabular}
\end{table}

\begin{table}[h]
    \centering
    \caption{Averaged area under the curve with standard error for multi-point acquisition functions with five initial training points per input dimension.}
    \small
    \begin{tabular}{ c c | c c c c c c c c }
    Type & Method & Sphere & Dixon-Price & Griewank & Hartmann & \makecell{Hartmann \\ low noise} & \makecell{Hartmann \\ high noise} & Michalewicz & Ackley \\
        \hline \hline
        & & & & & & & &\\[\dimexpr-\normalbaselineskip+5pt]
        \parbox{0mm}{\multirow{5}{10pt}{\rotatebox{90}{Sequential Monte Carlo\kern 20pt}}} & PI & \makecell{0.99 \\ ($\pm$ 0.01)} & \makecell{1.00 \\ ($\pm$ 0.00)} & \makecell{0.99 \\ ($\pm$ 0.00)} & \makecell{0.92 \\ ($\pm$ 0.03)} & \makecell{0.93 \\ ($\pm$ 0.02)} & \makecell{0.94 \\ ($\pm$ 0.02)} & \makecell{0.69 \\ ($\pm$ 0.06)} & \makecell{0.22 \\ ($\pm$ 0.08)} \\[0.4cm]
        & EI & \makecell{1.00 \\ ($\pm$ 0.00)} & \makecell{1.00 \\ ($\pm$ 0.00)} & \makecell{1.00 \\ ($\pm$ 0.00)} & \makecell{0.95 \\ ($\pm$ 0.03)} & \makecell{0.96 \\ ($\pm$ 0.02)} & \makecell{0.97 \\ ($\pm$ 0.02)} & \makecell{0.64 \\ ($\pm$ 0.08)} & \makecell{0.37 \\ ($\pm$ 0.22)} \\[0.4cm]
        & \makecell{UCB \\ (variable)} & \makecell{0.98 \\ ($\pm$ 0.01)} & \makecell{1.00 \\ ($\pm$ 0.00)} & \makecell{0.98 \\ ($\pm$ 0.01)} & \makecell{0.94 \\ ($\pm$ 0.03)} & \makecell{0.94 \\ ($\pm$ 0.02)} & \makecell{0.94 \\ ($\pm$ 0.03)} & \makecell{0.64 \\ ($\pm$ 0.08)} & \makecell{0.84 \\ ($\pm$ 0.01)} \\[0.5cm]
        & \makecell{UCB \\ ($\beta$=5)} & \makecell{0.99 \\ ($\pm$ 0.00)} & \makecell{1.00 \\ ($\pm$ 0.00)} & \makecell{0.99 \\ ($\pm$ 0.00)} & \makecell{0.95 \\ ($\pm$ 0.02)} & \makecell{0.96 \\ ($\pm$ 0.02)} & \makecell{0.96 \\ ($\pm$ 0.02)} & \makecell{0.72 \\ ($\pm$ 0.06)} & \makecell{0.88 \\ ($\pm$ 0.02)} \\[0.5cm]
        & \makecell{UCB \\ ($\beta$=1)} & \makecell{1.00 \\ ($\pm$ 0.00)} & \makecell{1.00 \\ ($\pm$ 0.00)} & \makecell{1.00 \\ ($\pm$ 0.00)} & \makecell{0.96 \\ ($\pm$ 0.02)} & \makecell{0.97 \\ ($\pm$ 0.02)} & \makecell{0.97 \\ ($\pm$ 0.02)} & \makecell{0.77 \\ ($\pm$ 0.07)} & \makecell{0.86 \\ ($\pm$ 0.02)} \\[0.5cm]
        \hline
        & & & & & & & &\\[\dimexpr-\normalbaselineskip+5pt]
        \parbox{0mm}{\multirow{5}{10pt}{\rotatebox{90}{Joint Monte Carlo\kern 30pt}}} & PI & \makecell{0.98 \\ ($\pm$ 0.01)} & \makecell{1.00 \\ ($\pm$ 0.00)} & \makecell{0.99 \\ ($\pm$ 0.00)} & \makecell{0.86 \\ ($\pm$ 0.05)} & \makecell{0.89 \\ ($\pm$ 0.03)} & \makecell{0.89 \\ ($\pm$ 0.04)} & \makecell{0.65 \\ ($\pm$ 0.06)} & \makecell{0.09 \\ ($\pm$ 0.06)} \\[0.4cm]
        & EI & \makecell{1.00 \\ ($\pm$ 0.00)} & \makecell{1.00 \\ ($\pm$ 0.00)} & \makecell{1.00 \\ ($\pm$ 0.00)} & \makecell{0.95 \\ ($\pm$ 0.03)} & \makecell{0.95 \\ ($\pm$ 0.02)} & \makecell{0.96 \\ ($\pm$ 0.02)} & \makecell{0.63 \\ ($\pm$ 0.08)} & \makecell{0.06 \\ ($\pm$ 0.03)} \\[0.4cm]
        & \makecell{UCB \\ (variable)} & \makecell{0.96 \\ ($\pm$ 0.02)} & \makecell{0.99 \\ ($\pm$ 0.01)} & \makecell{0.97 \\ ($\pm$ 0.02)} & \makecell{0.89 \\ ($\pm$ 0.04)} & \makecell{0.90 \\ ($\pm$ 0.05)} & \makecell{0.90 \\ ($\pm$ 0.04)} & \makecell{0.56 \\ ($\pm$ 0.07)} & \makecell{0.81 \\ ($\pm$ 0.02)} \\[0.5cm]
        & \makecell{UCB \\ ($\beta$=5)} & \makecell{0.99 \\ ($\pm$ 0.00)} & \makecell{1.00 \\ ($\pm$ 0.00)} & \makecell{0.99 \\ ($\pm$ 0.00)} & \makecell{0.94 \\ ($\pm$ 0.03)} & \makecell{0.96 \\ ($\pm$ 0.02)} & \makecell{0.95 \\ ($\pm$ 0.03)} & \makecell{0.72 \\ ($\pm$ 0.05)} & \makecell{0.87 \\ ($\pm$ 0.02)} \\[0.5cm]
        & \makecell{UCB \\ ($\beta$=1)} & \makecell{1.00 \\ ($\pm$ 0.00)} & \makecell{1.00 \\ ($\pm$ 0.00)} & \makecell{1.00 \\ ($\pm$ 0.00)} & \makecell{0.96 \\ ($\pm$ 0.02)} & \makecell{0.97 \\ ($\pm$ 0.02)} & \makecell{0.97 \\ ($\pm$ 0.02)} & \makecell{0.75 \\ ($\pm$ 0.07)} & \makecell{0.78 \\ ($\pm$ 0.06)} \\[0.5cm]
        \hline
        & & & & & & & &\\[\dimexpr-\normalbaselineskip+5pt]
        \parbox{0mm}{\multirow{5}{10pt}{\rotatebox{90}{Analytical\kern 43pt}}} & CL min & \makecell{0.99 \\ ($\pm$ 0.00)} & \makecell{1.00 \\ ($\pm$ 0.00)} & \makecell{0.99 \\ ($\pm$ 0.00)} & \makecell{0.94 \\ ($\pm$ 0.02)} & \makecell{0.95 \\ ($\pm$ 0.02)} & \makecell{0.95 \\ ($\pm$ 0.02)} & \makecell{0.62 \\ ($\pm$ 0.08)} & \makecell{0.40 \\ ($\pm$ 0.08)} \\[0.4cm]
        & CL max & \makecell{1.00 \\ ($\pm$ 0.00)} & \makecell{1.00 \\ ($\pm$ 0.00)} & \makecell{1.00 \\ ($\pm$ 0.00)} & \makecell{0.95 \\ ($\pm$ 0.03)} & \makecell{0.96 \\ ($\pm$ 0.02)} & \makecell{0.96 \\ ($\pm$ 0.03)} & \makecell{0.74 \\ ($\pm$ 0.10)} & \makecell{0.67 \\ ($\pm$ 0.09)} \\[0.4cm]
        & \makecell{BUCB \\ (variable)} & \makecell{0.98 \\ ($\pm$ 0.01)} & \makecell{1.00 \\ ($\pm$ 0.00)} & \makecell{0.98 \\ ($\pm$ 0.02)} & \makecell{0.85 \\ ($\pm$ 0.05)} & \makecell{0.84 \\ ($\pm$ 0.06)} & \makecell{0.85 \\ ($\pm$ 0.05)} & \makecell{0.61 \\ ($\pm$ 0.05)} & \makecell{0.65 \\ ($\pm$ 0.03)} \\[0.5cm]
        & \makecell{BUCB \\ ($\beta$=5)} & \makecell{0.98 \\ ($\pm$ 0.01)} & \makecell{1.00 \\ ($\pm$ 0.00)} & \makecell{0.98 \\ ($\pm$ 0.00)} & \makecell{0.88 \\ ($\pm$ 0.05)} & \makecell{0.89 \\ ($\pm$ 0.04)} & \makecell{0.90 \\ ($\pm$ 0.05)} & \makecell{0.69 \\ ($\pm$ 0.05)} & \makecell{0.66 \\ ($\pm$ 0.05)} \\[0.5cm]
        & \makecell{BUCB \\ ($\beta$=1)} & \makecell{0.99 \\ ($\pm$ 0.01)} & \makecell{1.00 \\ ($\pm$ 0.00)} & \makecell{0.99 \\ ($\pm$ 0.00)} & \makecell{0.88 \\ ($\pm$ 0.06)} & \makecell{0.89 \\ ($\pm$ 0.04)} & \makecell{0.90 \\ ($\pm$ 0.06)} & \makecell{0.74 \\ ($\pm$ 0.08)} & \makecell{0.65 \\ ($\pm$ 0.05)}
    \end{tabular}
\end{table}

\end{document}